%% file: main.tex
\documentclass{article}

    \usepackage[nonatbib, final]{neurips_2025}

\usepackage[numbers,sort&compress]{natbib}

\usepackage[utf8]{inputenc} 
\usepackage[T1]{fontenc}    
\usepackage{hyperref}       
\usepackage{url}            
\usepackage{booktabs}       
\usepackage{amsfonts}       
\usepackage{nicefrac}       
\usepackage{microtype}      

\hypersetup{
    colorlinks=true,
    linkcolor=RoyalBlue,
    filecolor=RoyalBlue,
    urlcolor=RoyalBlue,
    citecolor=RoyalBlue,
}

\input{format}

\title{RoboRefer: Towards Spatial Referring with Reasoning in Vision-Language Models for Robotics}

\author{%
\textbf{Enshen Zhou}\textsuperscript{1,3}\footnotemark[1], ~
\textbf{Jingkun An}\textsuperscript{1}\footnotemark[1], ~
\textbf{Cheng Chi}\textsuperscript{3}\footnotemark[1]~~\footnotemark[3]~~\footnotemark[2]~~, ~
\textbf{Yi Han}\textsuperscript{1,3}, ~
\textbf{Shanyu Rong}\textsuperscript{2,3}, ~
\textbf{Chi Zhang}\textsuperscript{2}, \\
\textbf{Pengwei Wang}\textsuperscript{3}, ~
\textbf{Zhongyuan Wang}\textsuperscript{3},~
\textbf{Tiejun Huang}\textsuperscript{2,3}, ~
\textbf{Lu Sheng}\textsuperscript{1}\footnotemark[2]~, ~
\textbf{Shanghang Zhang}\textsuperscript{2,3}\footnotemark[2]\\ 
$^{1}$School of Software, Beihang University\\ 
$^{2}$ State Key Laboratory of Multimedia Information Processing, \\ School of Computer Science, Peking University~~
$^{3}$Beijing Academy of Artificial Intelligence\\
\tt\footnotesize zhouenshen@buaa.edu.cn~~anjingkun02@gmail.com~~chicheng@baai.ac.cn\\
\tt\footnotesize lsheng@buaa.edu.cn~~shanghang@pku.edu.cn \\
}

\begin{document}

\maketitle

\vspace{-9mm}
\begin{figure}[ht]
\centering
\includegraphics[width=\linewidth]{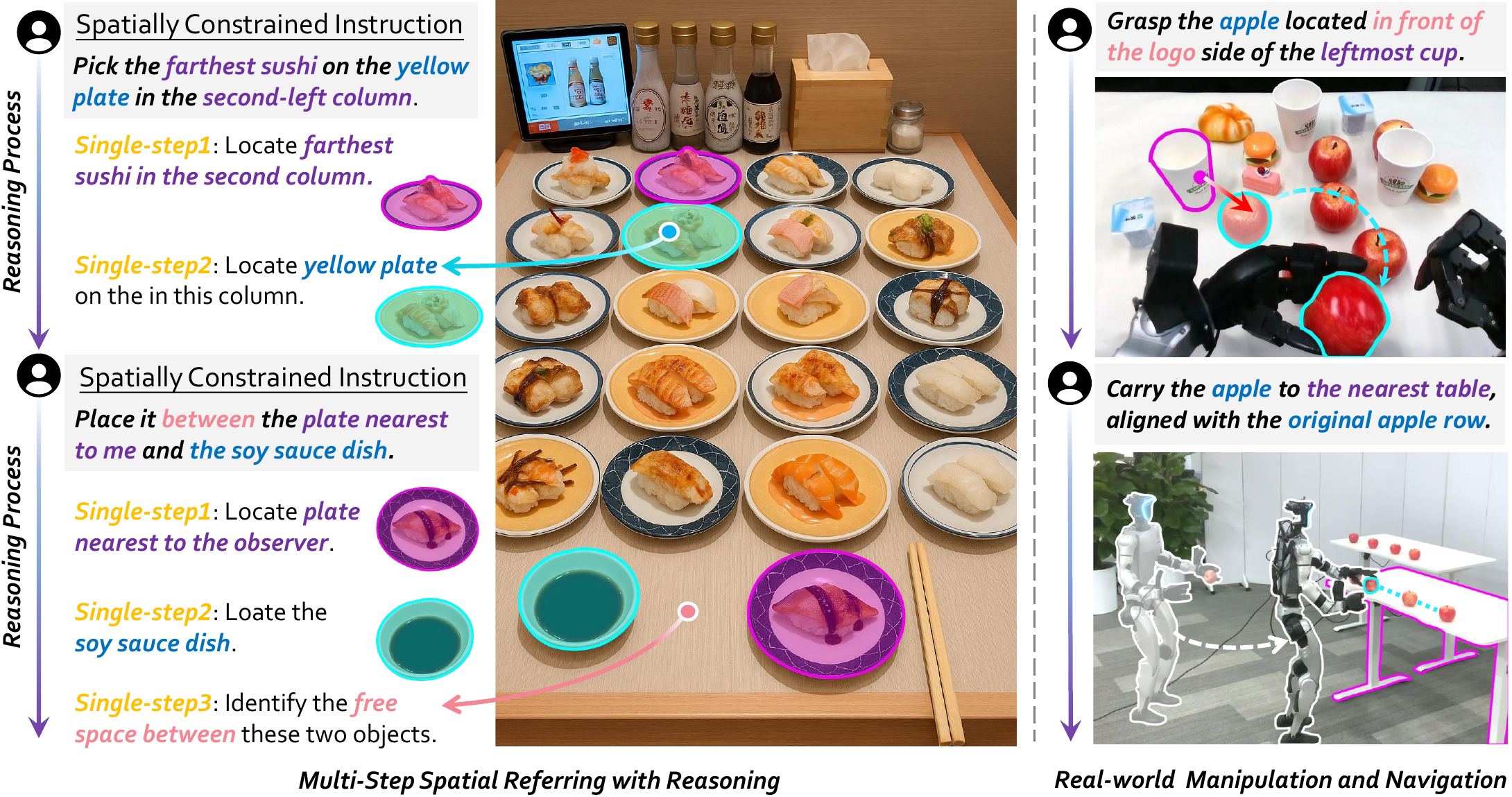}
\vspace{-6mm}
   \caption{Spatial referring in complex 3D environments demands not only precise \textit{single-step spatial understanding} but also \textit{multi-step spatial reasoning} to resolve intricate references step-by-step, thereby enabling efficient control of diverse robots across tasks~(\eg, manipulation, navigation).}
   \label{fig: motivation}
\end{figure}

\let\thefootnote\relax\footnotetext{$^*$ Equal contribution\hspace{3pt} \hspace{5pt}$^\dagger$ Corresponding author\hspace{5pt} $^\ddagger$ Project leader
}


\newtoggle{inappendix}
\togglefalse{inappendix}

\let\oldaddcontentsline\addcontentsline
\renewcommand{\addcontentsline}[3]{%
  \iftoggle{inappendix}{\oldaddcontentsline{#1}{#2}{#3}}{}%
}

\input{sec/0_abstract_v3}
\input{sec/1_intro_v4}
\input{sec/2_related_work}
\input{sec/3_method}
\input{sec/4_exp}

\input{sec/5_conclusion}

\bibliography{main}
\bibliographystyle{unsrt}

\input{sec/Appendix}


\end{document}

%% file: format.tex
\usepackage{wrapfig}
\usepackage{pifont}
\usepackage{graphicx}
\usepackage{tabularx}
\usepackage{multirow}
\usepackage{booktabs}
\usepackage{amsmath}  
\usepackage{amssymb}  
\usepackage{mathrsfs} 
\usepackage{amsthm,amsfonts}
\usepackage[table,dvipsnames]{xcolor}
\usepackage{array}
\usepackage{float}
\usepackage{pifont} 
\usepackage{bbding} 
\usepackage{fontawesome} 
\usepackage[most]{tcolorbox}
\usepackage{listings}
\usepackage{graphicx}
\usepackage{etoolbox}

\def\ie{\emph{i.e.}}
\def\eg{\emph{e.g.}}

\newcommand{\mname}{\textsl{RoboRefer}}
\newcommand{\dname}{\textsl{RefSpatial}}
\newcommand{\bname}{\textsl{RefSpatial-Bench}}
\definecolor{ourpurple}{HTML}{7030A0} 
\definecolor{ouryellow}{HTML}{F4B402}

\newcommand{\highlight}[1]{\textbf{\color{RoyalBlue}#1}}

\definecolor{codeblue}{rgb}{0.25,0.5,0.5}
\definecolor{myblue}{rgb}{0.88,0.98,1}
\definecolor{mygreen}{rgb}{0.92, 1.0, 0.92}
\definecolor{myred}{rgb}{1, 0.9, 0.9}
\definecolor{mygray}{gray}{0.95}
\definecolor{mydarkblue}{rgb}{0,0.08,1}
\definecolor{mydarkred}{rgb}{0.8,0.02,0.02}
\definecolor{mydarkorange}{rgb}{0.40,0.2,0.02}
\definecolor{mypurple}{RGB}{239,229,253}
\definecolor{mygold}{rgb}{0.75,0.6,0.12}
\definecolor{mydarkgray}{rgb}{0.66, 0.66, 0.66}
\definecolor{mydarkgreen}{rgb}{0.02,0.6,0.02}
\definecolor{mygray}{gray}{0.9}
\definecolor{tablegray}{gray}{0.95}

%% file: sec/0_abstract_v3.tex
\begin{abstract}
Spatial referring is a fundamental capability of embodied robots to interact with the 3D physical world.
However, even with the powerful pretrained vision language models (VLMs), recent approaches are still not qualified to accurately understand the complex 3D scenes and dynamically reason about the instruction-indicated locations for interaction.
%
%
%
%
To this end, we propose {\mname}, a 3D-aware VLM that can first achieve precise spatial understanding by integrating a disentangled but dedicated depth encoder via supervised fine-tuning (SFT).
%
Moreover, {\mname} advances generalized multi-step spatial reasoning via reinforcement fine-tuning (RFT), with metric-sensitive process reward functions tailored for spatial referring tasks.
%
%
%
To support SFT and RFT training, we introduce {\dname}, a large-scale dataset of $20$M QA pairs~($2\times$ prior), covering $31$ spatial relations (\textit{vs.} $15$ prior) and supporting complex reasoning processes (up to $5$ steps).
In addition, we present {\bname}, a challenging benchmark filling the gap in evaluating spatial referring with multi-step reasoning.
Experiments show that SFT-trained {\mname} achieves state-of-the-art spatial understanding, with an average success rate of $89.6$\%. 
%
RFT-trained {\mname} further outperforms all other baselines by a large margin, even surpassing Gemini-2.5-Pro by 17.4\% in average accuracy on {\bname}.
%
%
%
Notably, {\mname} can be integrated with various control policies to execute long-horizon, dynamic tasks across diverse robots~(\eg, UR5, G1 humanoid) in cluttered real-world scenes.
See the project page at \href{https://zhoues.github.io/RoboRefer/}{https://zhoues.github.io/RoboRefer}.

\end{abstract}

%% file: sec/1_intro_v4.tex
\section{Introduction}
\label{sec: intro}

Open-world spatial intelligence is crucial for embodied AI, as robots must understand and reason about 3D scenes to interact effectively in complex environments.
As one vital topic in this field, \textit{spatial referring}, which bridges spatial intelligence and embodied AI by formalizing how agents interpret and act upon spatially constrained instructions, has received increasing attention.
%
Specifically, given sensor observations (\eg, RGB or RGB-D) and a spatially constrained instruction, the spatial referring task aims to predict a precise point that satisfies complex spatial relations within the instruction.
%
%
This predicted point can serve various downstream embodied functions as navigation waypoints, manipulation targets, or placement locations, enabling wide robotic applications, as shown in Fig~\ref{fig: motivation}.

Spatial referring task comprises two distinct levels of complexity:
%
\textbf{(1)} \textit{Single-step spatial understanding}, which forms the foundation of spatial perception by accurately recognizing objects' spatial properties (\eg, position, orientation) and their spatial relations (\eg, distance, direction). This level, where most current research~\cite{cheng2024spatialrgpt, song2024robospatial, daxberger2025mm, ray2024sat, yuan2024robopoint, chen2024spatialvlm, qi2025sofar} concentrates, provides the essential perceptual basis for complex spatial referring.
\textbf{(2)} \textit{Multi-step spatial reasoning}, which transcends basic understanding through compositional reasoning to resolve complex spatial references sequentially. Despite its importance for sophisticated spatial intelligence, this capability remains underexplored. 
Thus, this work attempts to address this gap by integrating both levels for comprehensive spatial referring.
In Fig~\ref{fig: motivation}, one must first identify the plate closest to the observer and locate the desired soy sauce dish, then determine the free space between them, which is increasingly challenging as more spatial constraints are introduced.


Specifically, existing vision-language models (VLMs)~\cite{achiam2023gpt, team2023gemini, deitke2024molmo, bai2025qwen2}-based methods mainly attempt to enhance the first level, \ie, \textit{single-step spatial understanding} by integrating 3D inputs.
However, they either demand costly 3D reconstruction of multi-view images~\cite{hong20233dllm, hong20233d}, causing modality gaps 
%
, or treat depth as RGB-like inputs~\cite{cheng2024spatialrgpt,cai2024spatialbot, daxberger2025mm} via a shared image encoder, risking modality interference and degrading pretrained image encoders, requiring additional co-training data for compensation.
In contrast, the second level, \ie, \textit{multi-step spatial referring with reasoning}, remains underexplored due to the scarcity of suitable datasets, limiting current models' capability and preventing exploration of how single-step understanding might support it.
Moreover, current VLMs depend heavily on supervised fine-tuning (SFT) for implicit reasoning, risking memorizing answers over explicit reasoning and thereby hindering generalization and accuracy in open-world spatial referring.


In this work, we propose {\mname}, a 3D-aware VLM that not only acquires precise spatial understanding via SFT but also exhibits generalized strong reasoning capabilities for spatial referring via reinforcement fine-tuning~(RFT).
%
%
Specifically, for single-step spatial understanding, {\mname} employs a dedicated depth encoder to enhance precise spatial perception 
without interfering RGB branch.
%
%
%
%
To enable multi-step spatial reasoning, we design an RFT stage after SFT 
with explicitly annotated reasoning processes.
%
%
%
%
This stage allows {\mname} to break down complex spatial referring tasks into sequential analytical steps.
In each step, {\mname} can leverage the spatial understanding gained in SFT and refine the intermediate reasoning precision with our proposed metric-sensitive process reward functions, thus making more accurate point predictions.
%
%
To our best knowledge, 
{\mname} is the first 3D-aware reasoning VLM for multi-step spatial referring with explicit reasoning.

To advance spatial referring, we introduce {\dname}, a large-scale dataset of $2.5$M high-quality examples with $20$M QA pairs~($2\times$ prior~\cite{daxberger2025mm}).
Leveraging diverse data sources from 2D/3D/Simulation, this dataset can teach a general VLM to achieve spatial referring in a bottom-up manner.
Specifically, 2D web images provide fundamental spatial concepts and broad depth perception (indoor and outdoor), 3D embodied videos refine fine-grained spatial understanding of indoor scenes for robotics, and simulated data with ground-truth reasoning processes encourage multi-step spatial referring~(up to $5$ steps). 
Notably, {\dname} includes $31$ spatial relations, far exceeding $15$ found in previous datasets~\cite{song2024robospatial, daxberger2025mm}, and each sample contains RGB-D data to support depth alignment in SFT stage. 
%
%
%
%

%

%

We evaluate our SFT-trained model on existing single-step spatial reasoning benchmarks (\eg, CV-Bench~\cite{tong2024cambrian}, BLINK~\cite{fu2024blink}), achieving SOTA performance with an average success rate of $89.6$\%.
%
%
To address the lack of multi-step spatial referring benchmarks, we introduce {\bname}, comprising $200$ real-world images with manually annotated tasks for object location and placement. 
Over $70$\% of the samples require multi-step reasoning (up to $5$ steps) and are annotated with precise masks.
Our model consistently outperforms all baselines on this benchmark, even surpassing Gemini-2.5-Pro by an average of $17.4$\%.
Moreover, in Fig.~\ref{fig: motivation} and Sec.~\ref{subsec: deployment}, {\mname} can execute long-horizon, dynamic tasks in cluttered real-world scenes with various control policies, exhibiting strong generalization across robots (\eg, UR5, G1 humanoid) and tasks (\eg, manipulation, navigation).

Our contributions are summarized as follows:
\textbf{(1)} We propose {\mname}, a 3D-aware reasoning VLM trained using a sequential SFT-RFT strategy with metric-sensitive process reward functions to achieve spatial referring.
%
\textbf{(2)} We construct {\dname}, a well-annotated dataset tailored for spatial referring, facilitating both SFT and RFT training, and introduce {\bname}, a benchmark that fills the gap in evaluating spatial referring with multi-step reasoning.
%
%
\textbf{(3)} Extensive experiments show that {\mname} generalizes well, surpasses baselines in spatial understanding and referring with reasoning, and efficiently controls diverse robots across tasks in the real world.

    



%% file: sec/2_related_work.tex
\section{Related work}
\label{sec: related work}

\noindent\textbf{Spatial Understanding with VLMs.}
Spatial understanding~\cite{azuma2022scanqa, fu2024blink, hudson2019gqa, jia2024sceneverse, johnson2017clevr, szymanska2024space3d, du2024embspatial, ramakrishnan2024does, zhou2025robotracer} focuses on object-centric properties (\eg, position, orientation) and inter-object relations (\eg, distance, direction), while spatial reasoning~\cite{kamath2023s, liu2023visual, rajabi2023towards, ranasinghe2024learning, shiri2024empirical, lee2025perspective, wang2024picture, tang2024sparkle, liu2025spatialcot, liu2025ssr, ma2025spatialllm, ma2025spatialreasoner} draws higher-level inferences over such information.
Recent advances in VLMs~\cite{achiam2023gpt, team2023gemini, anthropic2024claude, liu2024nvila, bai2025qwen2, deitke2024molmo, yin2023lamm, qin2024mp5, qin2025navigatediff, qin2024worldsimbench, chen2024rh20t, zhou2024minedreamer, ji2025robobrain, huang2024story3d, li2025t2isafety, zhu2025internvl3, zhang2024navid, zhang2024uninavid, liu2024robomamba,  team2025robobrain, ji2025mathsticks,zhang2025beyond,li2025labutopia,wang2025trackvla,liu2023visual_instruction_tuning} enhance these two abilities via two paradigms: 
\textbf{(1)} tool-based approaches~\cite{maspatialpin, qi2025sofar, cai2024spatialbot, ding2024open6dor, qin2025robofactory, zhu2025dexflywheel,tan2025roboos, zhou2024code, han2025tiger} that integrate VLMs with vision foundation models~\cite{jin2023perspective, zhang2024recognize, liu2024segment, piccinelli2025unidepthv2,bochkovskii2024depth, yuan2025motiontrans,li2024lamp,li2024mulsmo,li2025languagelocomotionretargetingfreehumanoid,tschannen2025siglip, zhai2023sigmoid, zhu2023tame, kirillov2023segment, ravi2024sam} to extract and reason spatial cues
and \textbf{(2)} data-driven methods, which fine-tune VLMs using pseudo-3D annotations~\cite{chen2024spatialvlm, cheng2024spatialrgpt}, real-world 3D datasets~\cite{song2024robospatial, daxberger2025mm}, or simulated data~\cite{ray2024sat, wang2025pulsecheck457}.
However, existing datasets lack multi-step reasoning annotations critical for spatial referring tasks, and a benchmark for evaluating such abilities remains unavailable.
We thus introduce a new dataset and benchmark specifically tailored for spatial referring.




\noindent\textbf{Referring with VLMs for Robotics.}
Referring, also known as Referring Expression Comprehension (REC)~\cite{mao2016generation, van2006building, golland2010game, mitchell2010natural, mitchell2013generating, fitzgerald2013learning, kazemzadeh2014referitgame, yu2016modeling}, leverages unambiguous descriptions to localize a unique region/point in an image, and has seen great progress via VLMs~\cite{you2023ferret, peng2023kosmos, chen2023shikra, zhao2023bubogpt, wang2023visionllm, xiao2024florence}. 
Unlike Phrase Localization~\cite{plummer2015flickr30k, wang2016structured, plummer2017phrase} and Generalized Visual Grounding~\cite{xie2023described, he2023grec, liu2023gres, hu2023beyond, liu2024grounding}, which address ambiguous or multiple referents, REC focuses on one single target—an emphasis crucial for robotics, especially in pick-and-place tasks requiring precise object identification and destination~\cite{zeng2021transporter, liu2022structformer, liu2023structdiffusion, yuan2023m2t2, shridhar2022cliport}. 
While 2D REC relies on object attributes~(\eg, color) and image-plane localization~(\eg, top right of the image), real-world scenarios for robotics require 3D spatial reasoning to localize~(\eg, ``near'' \textit{vs.} ``far'').
Although efforts~\cite{team2025gemini, nasiriany2024pivot, yang2025magma} like RoboPoint~\cite{yuan2024robopoint} incorporate basic spatial cues via images to meet such expectations, they often struggle with complex environments and instructions required for spatial referring. 
Thus, we propose {\mname}, a 3D-aware framework that employs multi-step reasoning to ensure precise spatial referring for robotics.

\noindent\textbf{Reinforcement Fine-tuning for VLMs.}
Reinforcement Fine-tuning (RFT)~\cite{bai2022training, bai2022constitutional, rafailov2023direct, an2025agfsync, shao2024deepseekmath} is a post-training strategy that aligns models with human preferences or specific goals via feedback, complementing SFT~\cite{wei2021finetuned, zhou2023lima}, which adapts pre-trained models using task-oriented data.
Recent advances in LLM-based reasoning~\cite{jaech2024openai, guo2025deepseek, shao2024deepseekmath, zhang2025landscape,zhou2025reso} have shifted RL in VLMs toward visual reasoning~\cite{ yang2025r1, zhang2025r1, zhang2025mm, tan2025reason, kang2025viki, tan2025robo, zhang2025prune4web}, grounding~\cite{liu2025visual, zhan2025vision, shen2025vlm}, segmentation~\cite{liu2025seg} and trajectory prediction~\cite{song2025maniplvm}.
However, most methods rely solely on 2D perception, limiting their ability to handle spatial referring tasks that require 3D spatial reasoning.
To address this, we propose a two-stage training strategy: (1) incorporate depth information during SFT to strengthen spatial understanding; (2) RFT stage then leverages intermediate perception outputs powered by SFT to enable multi-step spatial referring with reasoning.


%% file: sec/3_method.tex
\section{Method}
\label{sec: method}

We first formulate the spatial referring task (Sec.~\ref{subsec: problem definition}). Then, we elaborate on {\mname}, including its architecture and training strategies (Sec.~\ref{subsec: mname}). Finally, we describe the construction of the {\dname} dataset (Sec.~\ref{subsec: dataset}) and necessary training details about {\mname} (Sec.~\ref{subsec: training details}).

\subsection{Problem Formulation}
\label{subsec: problem definition}

\begin{figure*}[t]
\centering
\includegraphics[width=\linewidth]{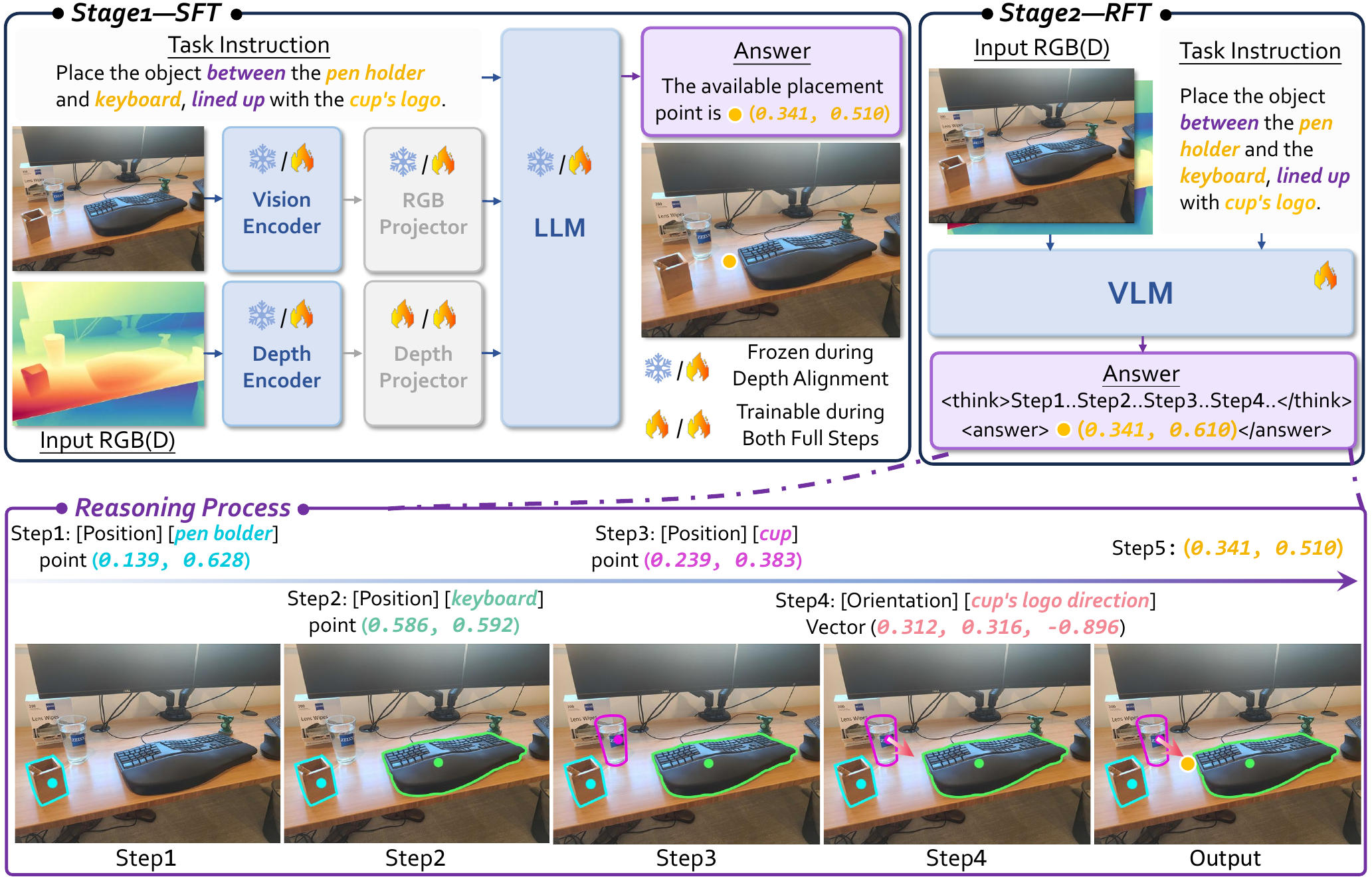}
\vspace{-5mm}
   \caption{Overview of {\mname}. {\mname} can perform \textit{single-step precise spatial understanding} from RGB(D) inputs with spatially constrained instructions (enabled by the SFT stage introducing depth modality), and \textit{multi-step spatial referring} with explicit reasoning (powered by the reinforcement fine-tuning stage and leveraging spatial understanding within each step learned in SFT).}
\label{fig: pipeline}
\vspace{-3mm}
\end{figure*}

We formulate spatial referring as predicting a single 2D point \((x, y)\) in image space to specify a target location or destination, given visual inputs \(\mathcal{O}\) (\eg, RGB or RGB-D) from the sensor and a textual instruction \(\mathcal{L}\).
This instruction encodes not only single-step spatial properties such as size (\eg, large, small), position (\eg, relative or ordinal location),  orientation (\eg, front-facing), and spatial relations (\eg, distance, direction), but also requires multi-step spatial reasoning.
For example, ``Place the object between the pen holder and keyboard, lined up with the cup's logo.'' (See Fig.~\ref{fig: pipeline}) becomes more complex as multiple spatial constraints are combined.
%


%
Unlike region-based 2D referring methods~\cite{liu2024grounding, you2023ferret, xiao2024florence}, \textit{this point-based formulation is more suitable and generalizable for robotics.}
Compared to the 2D bbox, point can naturally map to 3D coordinates via depth, providing accurate spatial anchors.
%
By leveraging predicted points for navigation, grasping, or placement, this formulation enables multi-task learning and execution.
Moreover, it can accurately localize a visible object part in occlusion scenarios, while 2D bbox often includes irrelevant objects.
%
%
%

\subsection{{\mname}: A 3D-aware reasoning VLM for spatial referring}
\label{subsec: mname}


\textbf{VLM Architecture.}
In Fig.~\ref{fig: pipeline}, {\mname} employs separate RGB and depth encoders to extract features, which are then aligned via projectors with the LLM for VQA or point prediction.
As 3D cues are vital for spatial understanding, 2D VLMs pretrained solely on RGB lack accurate 3D perception.
Recent methods~\cite{cheng2024spatialrgpt, daxberger2025mm, cai2024spatialbot} avoid explicit 3D representations by treating depth as an image-like modality and sharing the RGB encoder, but this causes modality interference, degrading the pretrained encoder and requiring additional RGB co-training to compensate.
To address this, we propose a simple yet effective approach: a dedicated depth encoder and projector, initialized from their RGB counterparts.
Notably, during joint RGB and RGB-D training, the image encoder remains unaffected by depth input, while the depth encoder is updated independently.
This design not only avoids modality interference and preserves general VQA performance without extensive RGB-only co-training, but it also improves spatial understanding through enhanced perception of depth cues (\eg, distance, near–far relations, and perspective-based size variations).
See Appx.~\ref{suppsubsec: architecture} for details.









\textbf{Supervised Fine-tuning.}
We adopt NVILA~\cite{liu2024nvila} as our base VLM; however, its 2D-only pretraining limits spatial understanding.
To address this, we propose a two-step SFT.
\textbf{(1)} Depth alignment. In Fig.~\ref{fig: pipeline}, we first train a depth projector to align the newly introduced depth space with the textual space, leveraging RGB-D annotations of the {\dname} (see Sec.~\ref{subsec: dataset}).
In this step, only the depth projector is updated.
\textbf{(2)} Spatial understanding enhancement.
We fine-tune all parameters on the {\dname}, including single-step fine-grained annotations and multi-step reasoning data with explicit reasoning processes, and additional instruction-following datasets~\cite{liu2024mitigating, liu2024improved, yu2016modeling}.
Therefore, the model is jointly optimized on RGB and RGB-D inputs, with separate updates for the image and depth encoders.
This process not only enhances single-step spatial understanding via the new depth modality but also bolsters implicit multi-step reasoning through data with explicit reasoning processes, providing a ``cold start'' for the subsequent RFT stage.
As a result, this SFT-trained model exhibits improved capability for multi-step spatial referring tasks.
Please check Appx.~\ref{suppsubsec: SFT training details} for details.

\textbf{Reinforcement Fine-tuning.}
Though SFT employs data with precise reasoning, it tends to memorize answers rather than generalize to novel spatial constraints.
We thus design a subsequent RFT stage using Group Relative Policy Optimization (GRPO~\cite{shao2024deepseekmath}) with multi-step reasoning data from {\dname}.
To guide RFT for more accurate point predictions, we first define two outcome reward~(\ie, only care about whether the output answer is correct) functions: \textbf{(1)} Outcome Format Reward~($R_{OF}$) for structured reasoning and clarity; 
and \textbf{(2)} Point L1 Reward~($R_{P}$) granting a score of $1$ if the final prediction falls within a specific range near the ground-truth point, and 0 otherwise.
%
To enhance intermediate reasoning precision, we exploit key-step perception annotations from {\dname} and design specialized metric-sensitive process reward functions:  
\textbf{(1)} Process Format Reward ($R_{PF}$), enforcing the format ``[Perception Type] [Target Object]:'';
\textbf{(2)} Accuracy Reward ($R_{Acc}$), which applies to steps included in the key-step perception annotations.
For each relevant step, we measure the prediction error using a specific metric, according to the perception type (\eg, L1 distance for positions between ground-truth points and predicted points).
Notably, this design is order-invariant and does not constrain the reasoning trajectory to a fixed sequence.
%
%
%
We sample $N$ responses $\{a_1, \ldots, a_N\}$ from the current policy (initialized from the SFT model) to encourage exploration. Each response receives a combined reward ($r_i = R_{OF}(a_i) + R_{P}(a_i) + \alpha R_{PF}(a_i) + \alpha R_{Acc}(a_i)$).
where $\alpha$ is set to 0.25.
Rewards are normalized within each group to compute relative advantages ($A_i = \frac{r_i - \text{mean}(\{r_j\})}{\text{std}(\{r_j\})}$), which are then used to update the policy, reinforcing high-quality responses and suppressing suboptimal ones.
%
%
A KL-divergence regularization term stabilizes updates by constraining them near the reference policy.  
Notably, the SFT initialization provides a strong prior, enabling rapid adaptation to output formats and supporting accurate, step-wise spatial reasoning by using the spatial understanding learned from SFT.
Fig~\ref{fig: pipeline} shows that the RFT-trained model generalizes well to tasks like $4$-step spatial referring, progressively handling intricate spatial relations, and yielding precise point predictions.
For more details about the RFT training and reward design, please see Appx.~\ref{suppsubsec: RFT training details}.

\subsection{{\dname} dataset}
\label{subsec: dataset}


\subsubsection{Overview}
\begin{figure*}[t]
\centering
\includegraphics[width=\linewidth]{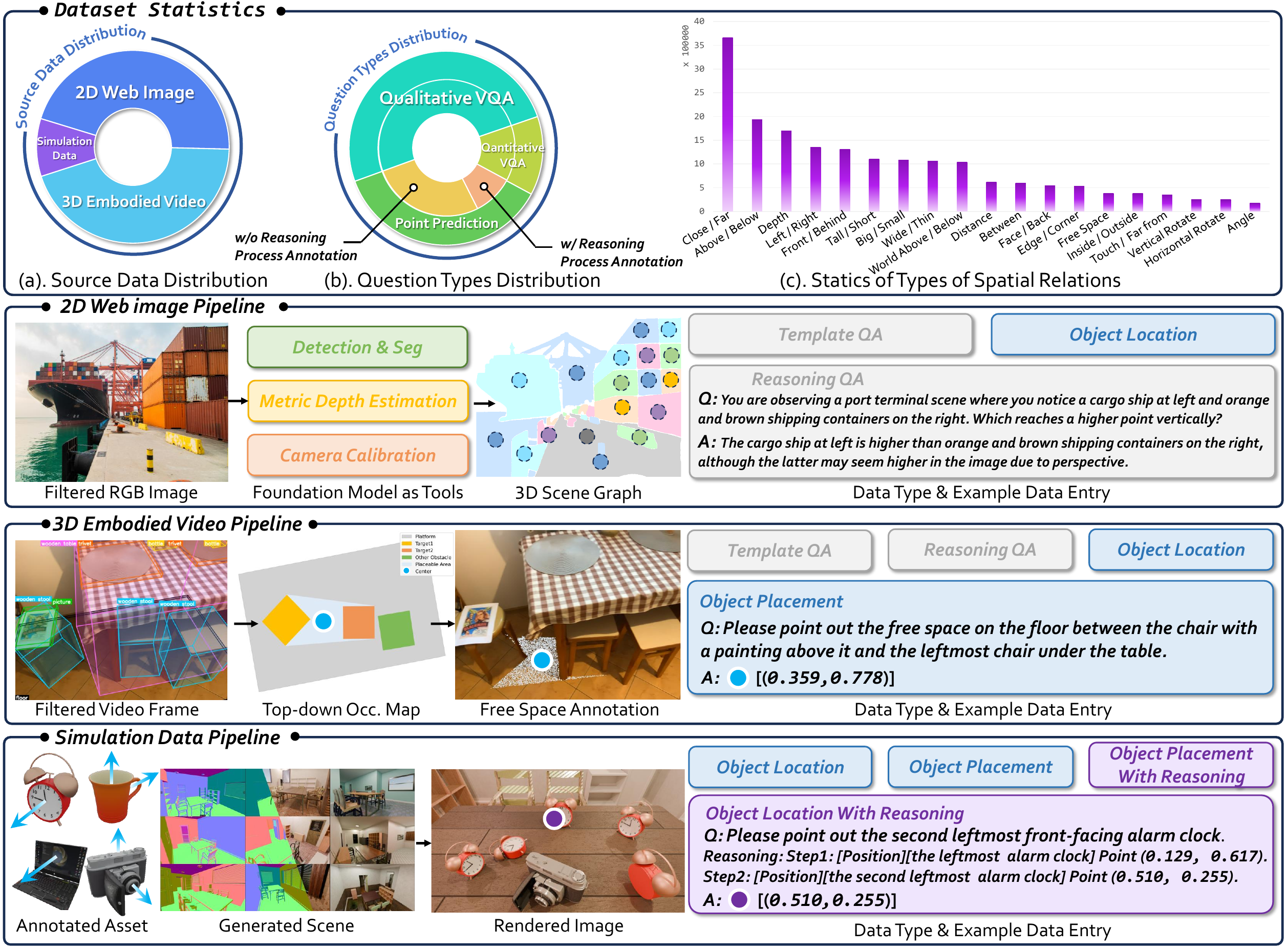}
\vspace{-4mm}
   \caption{{\dname}: $2.5$M data samples from 2D/3D/Simulated sources, with $31$ spatial relations.}
\label{fig: dataset}
\vspace{-5mm}
\end{figure*}

%
{\dname} is a comprehensive dataset integrating 2D images from OpenImages~\cite{kuznetsova2020open}, 3D embodied videos from CA-1M~\cite{lazarow2024cubify}, and simulated scenes from Infinigen~\cite{raistrick2024infinigen} using Objaverse~\cite{deitke2023objaverse} assets (See Fig.~\ref{fig: dataset}~(a)). 
{\dname}'s key features are:
\textbf{(1) Fine-Grained Annotations.} 
While prior spatial datasets~\cite{daxberger2025mm,song2024robospatial} simplify object reference by limiting each category to a single instance per scene, {\dname} includes multiple objects of the same category.
Moreover, each object is annotated with hierarchical captions—from broad categories (\eg, ``cup'') to precise spatial referents (\eg, ``the third cup from the left'', ``the cup closest to the camera'')—enabling unambiguous spatial referring in cluttered environments.
%
%
%
%
%
\textbf{(2) Multi-Dimensionality.} Beyond basic spatial concepts, relations, point coordinates, and point depth predictions, the dataset supports multi-step spatial reasoning by annotating detailed reasoning processes (all simulated data), addressing limitations in existing datasets.
\textbf{(3) High Quality.} We rigorously filter data to maintain quality. Retain 466k OpenImages containing text-referable, spatially relevant objects (down from 1.7M); sample 100k frames from CA-1M with text-identifiable 3D bounding boxes (down from 2M); and manually check and annotate 3k Objaverse-LVIS assets with semantic orientation labels (down from 46k).
\textbf{(4) Large Scale.} Comprising 2.5M samples and 20M QA pairs, our dataset spans qualitative VQA, quantitative queries on object attributes/relations, and point coordinate prediction (Fig.~\ref{fig: dataset}(b)).
\textbf{(5) Rich Diversity.} {\dname} spans indoor and outdoor scenes, covers common embodied scenes and integrates $31$ distinct spatial relations (See Fig.~\ref{fig: dataset}~(c)), fostering precise spatial understanding during SFT.
\textbf{(6) Easy Scalability.} Our pipeline seamlessly scales spatial referring data using diverse sources, including 2D images, 3D videos with bounding boxes, and simulation assets.
%
See Appx.~\ref{suppsec: dataset} for more dataset details.

%
%

\subsubsection{Data Recipe}

In Fig.~\ref{fig: dataset}, we present the dataset recipe that progressively integrates 2D, 3D, and simulated data to enable general VLMs to adapt to spatial referring tasks, thereby enhancing bottom-up spatial understanding and reasoning.
%
%
%
%
%
\textbf{(1) 2D Web Images} aim to endow the model with core spatial concepts and comprehensive depth perception across both indoor and outdoor scenes.
To mitigate depth scale and category discrepancies between indoor and outdoor scenes, we leverage the large-scale, diverse 2D web image dataset, OpenImage~\cite{kuznetsova2020open}.
However, directly extracting 3D-aware spatial information is challenging.
Inspired by prior work~\cite{cheng2024spatialrgpt, ma2025spatialreasoner}, we transform 2D images into pseudo-3D scene graphs.
In detail, after high-quality filtering (from 1.7M to 466K images), we further enhance the data using Qwen2.5-VL~\cite{bai2025qwen2} and a heuristic for generating hierarchical region captions, capturing both coarse labels and fine-grained spatial references, differentiating our approach from previous methods. 
We then construct scene graphs via object detection/segmentation, depth estimation, and camera intrinsic estimation, using object captions as nodes and spatial relations as edges.
Finally, we generate QA pairs via template-based or LLM-based approaches, augmented by object-location QA derived from the annotated captions.
\textbf{(2) 3D Embodied Videos} want to provide the model with a focused spatial understanding of indoor scenes, with a finer-grained perception of spatial relations and concepts.
We therefore leverage the richly annotated CA-1M~\cite{lazarow2024cubify}. 
After fine-grained filtering (from 2M to 100K frames), we construct 3D scene graphs with more diverse spatial relations, enabled by precise 3D bounding boxes compared to 2D approaches. 
Moreover, we generate top-down occupancy maps that encode the object positions, orientations, and metric distances (\eg, ``10cm right of the chair''), enabling accurate spatial referring for placement.
\textbf{(3) Simulation Data} arms the model with multi-step referring capabilities with spatial reasoning.
While 2D and 3D data enable single-step spatial understanding, they are less scalable for multi-step spatial referring with reasoning. 
Therefore, we leverage procedurally generate scene layouts~\cite{raistrick2024infinigen}, using manually verified assets~\cite{deitke2023objaverse} (from 46k to 3k) with semantic orientation annotations~\cite{qi2025sofar}. 
Tasks are purposefully designed to foster multi-step spatial referring and generate corresponding data. 
We assume that the generated code reflects optimal reasoning, with each line translated into textual form and intermediate results filled into structured formats (\eg, coordinates, distances), as shown in Fig.~\ref{fig: pipeline}, Fig.~\ref{fig: dataset}, and Appx.~\ref{suppsubsubsection: reward design and policy update}, yielding QA pairs with reasoning annotations.
For more demonstrations about {\dname}, please refer to Appx.~\ref{suppsec: more demonstrations}.

\subsection{Training Details}
\label{subsec: training details}

We adopt NVILA~\cite{liu2024nvila} (2B/8B) as the base model and apply SFT to obtain {\mname}-SFT.
Due to computational limits, RFT is applied only to the 2B model, yielding {\mname}-RFT.
SFT has two steps: the first uses only the {\dname}; the second trains on a mixture of {\dname}, instruction tuning~(1/20 the size of {\dname} QA)\cite{liu2024mitigating,liu2024improved}, and referring datasets\cite{yu2016modeling}.
Notably, {\dname} is reused with both RGB and RGB-D inputs in the second step to enforce the image encoder to learn spatial understanding beyond depth cues.
Thus, the model supports both RGB-only and RGB-D inference, with depth optionally inferred via a relative depth estimation model~\cite{yang2024depth}.
Finally, RFT stage uses the multi-step reasoning data from {\dname} to train.
See Appx.~\ref{suppsec: implementation details} for details.

%% file: sec/4_exp.tex
\section{Experiments}
\label{sec: experiments}


\begin{table}[t]
\caption{Performance on the \textit{single-step spatial understanding} benchmarks across different model types.
Top-1 \& Top-2 accuracies are represented using \textbf{bold text}, and \underline{underlines}.}
\centering
\scriptsize
\setlength{\tabcolsep}{1.5pt}
\begin{tabular}{ll|ccc|cc|ccc}
\toprule
\multirow{2}{*}{Method}  & \multirow{2}{*}{Input}             & \multicolumn{3}{c|}{CV-Bench~\cite{tong2024cambrian}}       & \multicolumn{2}{c|}{$\text{BLINK}_{val}$~\cite{fu2024blink}} & \multirow{2}{*}{RoboSpatial~\cite{song2024robospatial}} & \multirow{2}{*}{SAT~\cite{ray2024sat}} & \multirow{2}{*}{EmbSpatial~\cite{du2024embspatial}} \\
                        &                                    & 2D-Relation & 3D-Depth & 3D-Distance & 2D-Relation & 3D-Depth &   &  & \\
\midrule
\multicolumn{10}{c}{\cellcolor{mygray}\textit{Proprietary Models}} \\
\midrule
GPT-4o~\cite{achiam2023gpt}              & RGB   & 84.62 & 86.50 & 83.33 & 82.52 & 78.23 & 77.20 & 68.67 & 63.38  \\
Gemini-2.5-Pro~\cite{team2023gemini}  & RGB   & 93.54 & 91.00 & 90.67 & \textbf{91.61} & 87.90 & 77.24 & 70.59 & \textbf{76.67}\\
Claude-3.7-Sonnet~\cite{anthropic2024claude}   & RGB   & 74.15 & 85.83 & 84.17 & 74.83 & 67.74 & 60.73 & 40.67 & 33.33\\
\midrule
\multicolumn{10}{c}{\cellcolor{myred}\textit{Open-Source Vison-Language Models}} \\
\midrule
NVILA-2B~\cite{liu2024nvila}            & RGB   & 70.15 & 79.67 & 60.00 & 67.83 & 62.10 & 51.79 & 31.33 & 47.34\\
NVILA-8B~\cite{liu2024nvila}            & RGB   & 91.54 & 91.83 & 90.67 & 76.92 & 76.61 & 59.35 & 63.33 & 67.72\\
Qwen-2.5-VL-7B~\cite{bai2025qwen2}      & RGB   & 82.15 & 60.17 & 69.00 & 64.34 & 60.98 & 49.59 & 30.00 & 40.20\\
Qwen-2.5-VL-72B~\cite{bai2025qwen2}     & RGB   & 84.15 & 86.17 & 84.15 & 78.32 & 73.55 & 70.73 & 65.33 & 57.69\\
\midrule
\multicolumn{10}{c}{\cellcolor{mygreen}\textit{Spatial Specialist Models}} \\
\midrule
SpatialBot-3B~\cite{cai2024spatialbot}       & RGB-D  & 69.38 & 77.33 & 60.83 & 67.83 & 67.74 & 72.36 & 63.33 & 50.66\\
%
SpatialRGPT-8B~\cite{cheng2024spatialrgpt}	 & RGB-D  & 91.00 &	89.8 &88.50	&81.12	&89.51	&66.67	&64.00	&59.62 \\
SpaceLLaVA-13B~\cite{chen2024spatialvlm}     & RGB   & 63.69 & 66.83 & 70.17 & 72.73 & 62.90 & 61.00 & 62.67 & 49.40\\
RoboPoint-13B~\cite{yuan2024robopoint}       & RGB   & 75.85 & 77.83 & 44.50 & 60.84 & 61.29 & 69.90 & 46.60 & 49.31\\
\midrule
\multicolumn{10}{c}{\cellcolor{myblue}\textit{{\mname} Variants}} \\
\midrule
{\mname}-2B-SFT     & RGB    & 96.15 & 95.83 & 90.67 & 83.92 & 88.71 & \underline{82.93} & 71.33 & 70.66\\
{\mname}-2B-SFT     & RGB-D  & \underline{96.31} & \underline{97.17} & \underline{90.83} & \underline{87.41} & \underline{91.13} & \underline{82.93} & \underline{82.00} & 71.10  \\
{\mname}-8B-SFT     & RGB-D  & \textbf{96.90} & \textbf{98.33} & \textbf{93.50} & \textbf{91.61} & \textbf{92.74} & \textbf{84.55} & \textbf{86.67} & \underline{72.53}  \\
\bottomrule[1pt]
\end{tabular}
\label{tab: understanding}
\end{table}

\begin{table}[t]
\caption{
Performance on current referring and \textit{multi-step spatial referring} benchmarks.
L. and P. denote our benchmark's Location and Placement parts; U. indicates unseen compositional spatial relations during SFT/RFT.
Top-1 \& Top-2 accuracies are represented using \textbf{bold text}, and \underline{underlines}.
}
\centering
\scriptsize
\setlength{\tabcolsep}{2pt}
\begin{tabular}{l|ccccccccc}
\toprule
\multirow{3}{*}{Benchmark}
& \cellcolor{mygray} \textit{Proprietary Models}
& \multicolumn{4}{c}{\cellcolor{mygreen}\textit{Referring Specialist Models}}   
& \multicolumn{3}{c}{\cellcolor{myblue}{\mname} \textit{Variants}} \\

& \cellcolor{mygray}Gemini-2.5-Pro~\cite{team2023gemini}
& \cellcolor{mygreen}SpaceLLaVA~\cite{chen2024spatialvlm}
& \cellcolor{mygreen}RoboPoint~\cite{yuan2024robopoint}
& \cellcolor{mygreen}Molmo-7B~\cite{tong2024cambrian}
& \cellcolor{mygreen}Molmo-72B~\cite{tong2024cambrian}
& \cellcolor{myblue}2B-SFT  
& \cellcolor{myblue}8B-SFT
& \cellcolor{myblue}2B-RFT\\

\midrule
RoboRefIt   & -    & 21.3 & 49.8 & -    & -    & 72.8 & \textbf{75.9} & \underline{74.2}\\
Where2Place & 61.9 & 11.8 & 46.8 & 45.0 & 63.8 & 66.0 & \underline{70.0} & \textbf{71.0}\\
RoboSpatial & 40.2 & 16.0 & 41.3 & 38.0 & 40.9 & 66.4 & \underline{70.8} & \textbf{71.3}\\


\midrule

{\bname}-L.  & 46.96 & 5.82 & 22.87 & 21.91 & 45.77 & \underline{47.00} & \textbf{52.00} & \textbf{52.00} \\
{\bname}-P.  & 24.21 & 4.31 & 9.27 & 12.85 & 14.74 & 48.00 & \underline{53.00} & \textbf{54.00} \\
{\bname}-U.  & 27.14 & 4.02 & 8.40 & 12.23 & 21.24 & 33.77 & \underline{37.66} & \textbf{41.56} \\


\bottomrule[1pt]
\end{tabular}
\label{tab: referring}
\vspace{-5mm}
\end{table}

\subsection{Single-step Spatial Understanding}
\label{subsec: single-step spatial understanding}

We evaluate on public \textit{single-step spatial understanding benchmarks}, including CV-Bench~\cite{tong2024cambrian}, the BLINK~\cite{fu2024blink} validation split, RoboSpatial~\cite{song2024robospatial} configuration part, SAT~\cite{ray2024sat}, and EmbSpatial~\cite{du2024embspatial}. 
Check Appx.~\ref{suppsubsec: spatial understanding benchmarks} for more evaluation details.
%
The following parts present our analyses.

\textbf{SFT stage enables strong spatial understanding.}
In Tab~\ref{tab: understanding}, trained solely on {\dname}, {\mname}-SFT surpasses all spatial specialist models on these benchmarks, even surpassing Gemini-2.5-Pro by 5\% (absolute) on average.
%
Moreover, our 2B variant outperforms NVILA-2B by 21.7\% (absolute).

\textbf{Depth input improve 3D spatial understanding during inference.}
In Tab.~\ref{tab: understanding}, we find that incorporating depth information during inference leads to relative improvements in 3D benchmarks compared to 2D ones by 1.5\%, although our model exhibits strong spatial understanding with RGB input alone by reusing the {\dname} dataset with both RGB and RGB-D inputs during SFT's second step.

\subsection{Multi-step Spatial Referring}
\label{subsec: multi-step spatial referring}
\vspace{-2mm}

We first evaluate current robotic referring benchmarks, namely RoboRefIt~\cite{lu2023vl} (location) and Where2Place~\cite{yuan2024robopoint}/RoboSpatial~\cite{song2024robospatial} (placement), all limited to $2$ reasoning steps. 
To evaluate more complex multi-step spatial referring, we propose {\bname}, a challenging benchmark based on real-world cluttered scenes. 
It contains two subsets, Location and Placement, each with $100$ images.
%
Notably, $77$ images involve spatial relation combinations unseen in {\dname}.
Over $70\%$ requires multi-step reasoning (up to $5$ steps), including precise ground-truth masks.
More details about {\dname} can be found in Appx.~\ref{suppsec: implementation_of_benchmark}.
%
For metrics, we report the average success rate of predicted points within the mask.
We evaluate {\mname} using RGB-D inputs by default, with depth maps generated from RGB images via DepthAnything V2~\cite{yang2024depth}.
See Appx.~\ref{suppsubsec: spatial referring benchmarks} for more details.

\begin{table}[t]
\centering
\begin{minipage}[t]{0.55\textwidth}
\centering
\caption{Performance on general referring benchmarks.
B. and P. denote Bounding Box and Point.
Top-1/2 accuracies are indicated by \textbf{bold}/\underline{underlined} text.
%
}
\scriptsize
\setlength{\tabcolsep}{0.8pt}
\begin{tabular}{lc|ccc}
\toprule
\multirow{2}{*}{Method} & \multirow{2}{*}{Output} & RefCOCO & RefCOCO+ & RefCOCOg \\
       &        & val / testA / testB  & val / testA / testB & val / test \\
\midrule
\multicolumn{5}{c}{\cellcolor{mygray}\textit{Grounding Specialist Models}} \\
\midrule
GroundingDINO & BBox & 90.6 / 93.2 / 88.2 & 88.2 / 89.0 / 75.9 & 86.1 / 87.0\\
\midrule
\multicolumn{5}{c}{\cellcolor{myred}\textit{Open-Source Vision-Language Models}} \\
\midrule
Qwen2.5-VL-72B & BBox & 92.7 / 94.6 / 89.7 & 88.9 / 92.2 / 83.7 & 89.9 / 90.3\\
Qwen2.5-VL-72B & Point & 95.2 / 96.5 / 93.8 & 90.4 / 93.5 / 86.7 & 91.5 / 92.0\\
Qwen2.5-VL-72B & B. $\rightarrow$ P. & 95.4 / \underline{97.0} / \underline{94.2} & 91.5 / \underline{94.9}/ \textbf{88.2} & 92.5 / 92.5\\
\midrule
\multicolumn{5}{c}{\cellcolor{myblue}\textit{{\mname} Variants}} \\
\midrule
{\mname}-2B-SFT& Point & 95.5 / 96.7 / 93.8 & 91.8 / 94.3 / 86.3 & \underline{93.3} / \underline{93.6} \\
{\mname}-8B-SFT& Point & \textbf{96.6} / \textbf{97.7} / \textbf{94.7} & \underline{91.9} / \textbf{95.2} / \underline{87.5} & \textbf{94.3} / \textbf{94.1}\\
{\mname}-2B-RFT& Point & \underline{95.6} / 96.6 / 93.9 & \textbf{92.0} / 94.2 / 86.3 & 93.2 / \underline{93.63} \\
\bottomrule[1pt]
\end{tabular}
\label{tab: refcoco}
\end{minipage}
\hfill
\begin{minipage}[t]{0.435\textwidth}
\setlength{\tabcolsep}{0.4pt}
\centering
\caption{
Performance on general VLM benchmarks. We also show the advantage of dedicated depth encoder (E. = Encoder).
%
%
We use the same evaluation protocol of CV-Bench/BLINK in Sec.~\ref{subsec: single-step spatial understanding}.
}
\vspace{+2.5mm}
\scriptsize
\begin{tabular}{l|c|c|c}
\toprule
\multirow{2}{*}{Benchmark} & \multirow{2}{*}{NVILA-2B~\cite{liu2024nvila}} & \multicolumn{2}{c}{{\mname}-2B-SFT}\\
                           &                                               & Shared E. & Dedicated E.~(Ours)  \\
\midrule
\multicolumn{4}{c}{\cellcolor{myred}\textit{Public Vision-Language Benchmarks}} \\
\midrule
$\text{MME}_{test}$ & \underline{1547} & 1541 & \textbf{1553} \\
$\text{MMBench}_{dev}$ & \textbf{78.63} & 76.23 & \underline{77.73}\\
OK-VQA   & 61.93 & \underline{64.9} & \textbf{65.81}\\
POPE   & \underline{81.96} & 81.93 & \textbf{83.44}\\
\midrule
\multicolumn{4}{c}{\cellcolor{mygreen}\textit{Single-step Spatial Understanding Benchmarks}} \\
\midrule
CV-Bench  & 69.94 & \underline{93.99} & \textbf{94.22} \\
$\text{BLINK}_{val}$ & 64.97  & \underline{80.02} & \textbf{85.26} \\
\bottomrule[1pt]
\end{tabular}
\label{tab: general}
\end{minipage}
\vspace{-1.5mm}
\end{table}

\begin{figure*}[t]
\centering
\includegraphics[width=\linewidth]{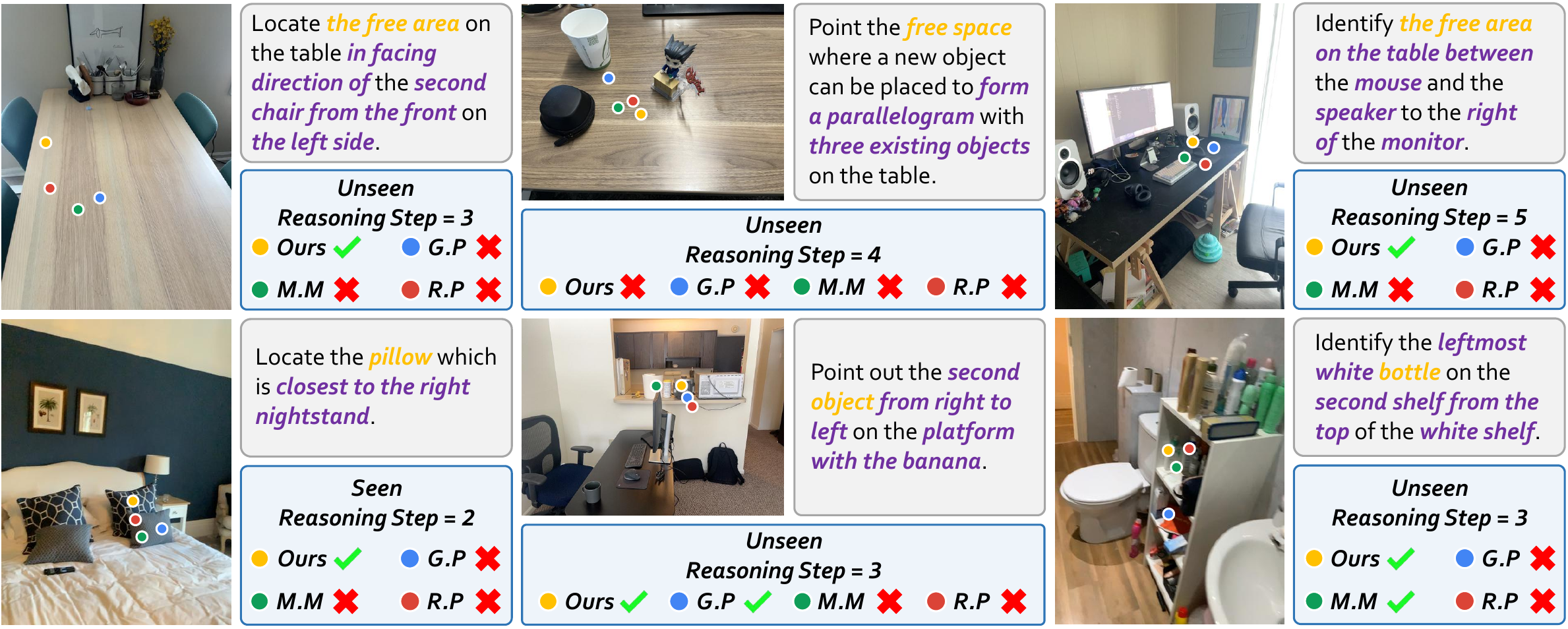}
\vspace{-3mm}
   \caption{{\bname} results. G.P., M.M., and R.P. donate Gemini-2.5-Pro~\cite{team2023gemini}, Molmo-72B~\cite{tong2024cambrian}, and RoboPoint~\cite{yuan2024robopoint}. {\mname}-RFT excels in unseen and multi-step cases.}
\label{fig: benchmark}
\vspace{-4mm}
\end{figure*}

\textbf{RFT stage fosters better reasoning ability.}
As shown in Tab.~\ref{tab: referring}, the 2B-RFT variant outperforms all baselines, exceeding the prior SOTA (Gemini-2.5-Pro~\cite{team2023gemini}) by 17.4\% (absolute) on {\bname}. 
We find that although Gemini-2.5-Pro excels in 2D referring (\eg, color, image-space localization), it struggles with 3D spatial relations involving distance (\eg, identifying the second-farthest object), reducing overall performance when multiple spatial constraints are combined. 
Fig.~\ref{fig: benchmark} shows complex multi-step spatial reasoning cases from {\bname} with model comparisons.

\textbf{RFT stage provides powerful generalization ability.}
In the {\bname}-Unseen row of Tab.~\ref{tab: referring}, we evaluate novel spatial relation combinations omitted during our SFT/RFT training. 
The 2B-RFT model exceeds 2B-SFT by 9.1\% (absolute), indicating that SFT overfits the training distribution, whereas the RFT model better generalizes by leveraging learned spatial knowledge, consistent with prior findings~\cite{chu2025sft}. Fig.~\ref{fig: benchmark} shows results on these unseen combinations across different models.

\subsection{Public vision-language benchmarks}
\label{subsec: public vision-language benchmarks}

\textbf{{\dname} enhance 2D general referring ability.}
We also evaluate 2D referring capability on the ReCOCO/+/g~\cite{yu2016modeling}. 
Since our model predicts a single point, we deem a prediction correct if the point lies within the ground-truth bounding box. 
As this evaluation differs from standard visual grounding protocols, we additionally assess Qwen-2.5VL-72B~\cite{bai2025qwen2} by using either its predicted point or the center of its predicted box as baselines. 
In Tab.~\ref{tab: refcoco}, our method surpasses baselines, indicating that our dataset not only supports 3D spatial referring but also enhances 2D referring performance.

\textbf{Joint RGB and RGB-D training preserves commonsense knowledge.}
In Tab.~\ref{tab: general}, we assess how spatial and depth information influences overall VQA performance by comparing {\mname}-2B-SFT with the baseline NVILA-2B~\cite{liu2024nvila}, trained on standard VQA datasets. 
Our model achieves comparable or slightly superior results, corroborating insights from SpatialVLM~\cite{chen2024spatialvlm} and SpatialRGPT~\cite{cheng2024spatialrgpt}. 
These findings indicate that although VLMs often struggle with spatial reasoning, targeted spatial VQA training, especially with combined RGB and RGB-D data enriched by general visual instruction datasets, can enhance spatial understanding without compromising overall VQA performance.

\subsection{Simulator and Real-world Evaluation for Robotics}
\label{subsec: deployment}

\textbf{{\mname} can be integrated into the system as a useful tool.}
We evaluate our model on the Open6DOR~\cite{ding2024open6dor} V2 position track, comparing against VLA-based baselines (pretrained Octo~\cite{team2024octo}, LIBERO-finetuned OpenVLA~\cite{kimopenvla}) and SoFar~\cite{qi2025sofar}, which integrates Florence-2~\cite{xiao2024florence}, SAM~\cite{ravi2024sam}, GPT-4o, and GSNet. 
{\mname} serves as a lightweight alternative to Florence-2 and GPT-4o for object localization and placement.
%
%
%
%
By using a single target point predicted by {\mname}, the system can generate more accurate masks and corresponding grasp poses than those from 2D boxes under occlusion in cluttered scenes, yielding a 6.8\% absolute improvement in success rate (Tab.~\ref{tab: simulation}).
%
Its compact size also reduces execution time by 27.5\% relative to GPT-4o.
See Appx.~\ref{suppsubsec: simulation evaluation} for details.

\begin{table}[t]
\centering
\begin{minipage}[t]{0.30\textwidth}
\centering
\scriptsize
\setlength{\tabcolsep}{1pt}
\caption{Simulation Results}
\begin{tabular}{l|cccc|c}
\toprule                          
\multirow{2}{*}{Method} & \multicolumn{4}{c}{Success Rate(\%) $\uparrow$} & Execution\\
                        & L.1 & L.2 & L.3 &Avg. & Time(s) $\downarrow$  \\
\midrule
Octo & 51.2 & 12.7 & 0.0 & 43.2 & -\\
OpenVLA & 51.6&  13.1 &  0.0&  43.6 & -\\
\midrule
SoFar & 75.3 & 65.6 & 50.0 & 72.4 & 40\\
\rowcolor{myblue}
\textbf{Ours} & \textbf{81.4} & \textbf{73.1} & \textbf{80.0} & \textbf{79.2} & \textbf{29}\\
\bottomrule[1pt]
\end{tabular}
\label{tab: simulation}
\end{minipage}
\hfill
\begin{minipage}[t]{0.69\textwidth}
\centering
\scriptsize
\setlength{\tabcolsep}{2pt}
\caption{Real-world robot evaluation requiring spatial referring.
%
}
\vspace{-0.8mm}
\begin{tabular}{l|ccc}
\toprule                          
\multirow{2}{*}{Manipulation or Navigation tasks with spatial referring} & \multicolumn{3}{c}{Success Rate(\%) $\uparrow$} \\
                      & OpenVLA & RoboPoint & \textbf{Ours} \\
\midrule
Pick the specific hamburger closest to the mug nearest & \multirow{2}{*}{0.00} & \multirow{2}{*}{0.00} & \multirow{2}{*}{\textbf{80.00}} \\
the camera and place it in front of the teddy bear. & & &\\
\midrule
Pick the apple in front of the leftmost cup's logo side, navigate & \multirow{2}{*}{0.00} & \multirow{2}{*}{0.00} & \multirow{2}{*}{\textbf{60.00}} \\
to the nearest table, and place it aligned with the apple row. & & & \\
\bottomrule[1pt]
\end{tabular}
\label{tab: real world}
\end{minipage}
\end{table}

\begin{figure*}[t]
\centering
\includegraphics[width=\linewidth]{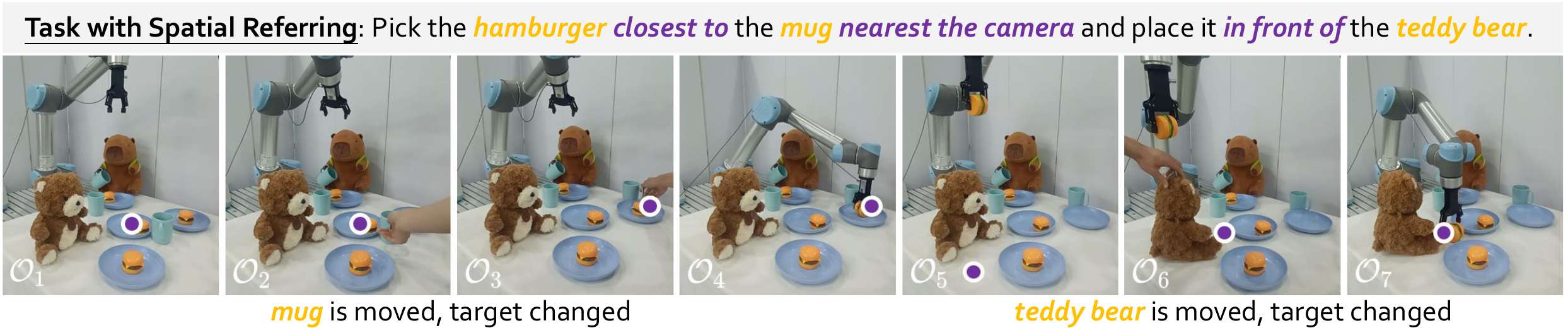}
\vspace{-6mm}
   \caption{Real-World Evaluation. The purple point denotes the current target predicted by our model.}
\label{fig: real_exp}
\vspace{-3mm}
\end{figure*}

\textbf{Spatial referring from {\mname} is crucial for real-world robots.}
In Tab.~\ref{tab: real world}, only our method can handle long-horizon tasks requiring complex multi-step spatial referring in cluttered and dynamic environments. 
These tasks are challenging, as the robot must precisely identify objects and their placement to satisfy spatial constraints that may change over time. 
In Fig.~\ref{fig: real_exp}, integrating {\mname} with an open-loop policy enables rapid updates at 2.5 Hz.
Thus, when the mug nearest the camera is moved, the robot adapts by grasping the hamburger closest to the mug’s new position and also readjusts placement after the teddy bear’s $90^\circ$ rotation, preserving correct spatial alignment.
Notably, spatial referring unifies both manipulation and navigation under a single formulation.
This allows the G1 humanoid to navigate while performing spatially constrained pick-and-place actions (Fig.~\ref{fig: motivation}), thereby enabling more complex, long-horizon tasks.
Check Appx.~\ref{suppsubsec: real-world evaluation} for more details.

\subsection{Ablation Study}
\label{subsec: ablation}

\begin{wraptable}{r}{0.32\textwidth}
\vspace{-7mm}
\caption{
Ablation Studies. S.D. means simulated data. P.R. denotes process reward.
We use the same evaluation protocol in Sec.~\ref{subsec: single-step spatial understanding} and Sec.~\ref{subsec: multi-step spatial referring}.
}
\vspace{1mm}
\scriptsize
\centering
\setlength{\tabcolsep}{1pt}
\begin{tabular}{cccc|cc}
\toprule                          
\multicolumn{3}{c}{Data Recipe} & Depth & \multicolumn{2}{c}{Spatial Understanding} \\
\cmidrule(lr){1-3} \cmidrule(lr){5-6}
2D & 3D & S.D. & Encoder & CV-Bench & $\text{BLINK}_{val}$ \\
\midrule
\multicolumn{6}{c}{\cellcolor{mygreen}\textit{SFT Variants (2B)}} \\
\midrule
\ding{55} & \checkmark & \checkmark  & \checkmark  & 84.17 & 74.48 \\
\checkmark & \ding{55} & \checkmark  & \checkmark  & 81.83 & 74.61 \\
\checkmark & \checkmark & \ding{55}  & \checkmark  & 83.96 & 75.10 \\
\checkmark & \checkmark & \checkmark & \ding{55}   & 91.24 & 85.27 \\
\checkmark & \checkmark & \checkmark & \checkmark & \textbf{94.77} & \textbf{89.27} \\
\midrule
\multicolumn{2}{c}{Reward} & \multicolumn{2}{c|}{Depth} & \multicolumn{2}{c}{Spatial Referring} \\
\cmidrule(lr){1-2}  \cmidrule(lr){5-6}
\multicolumn{2}{c}{P.R.} & \multicolumn{2}{c|}{ Encoder} & \multicolumn{2}{c}{{\bname}} \\
\midrule
\multicolumn{6}{c}{\cellcolor{myblue}\textit{RFT Variants (2B)}}\\
\midrule
\multicolumn{2}{c}{\checkmark} & \multicolumn{2}{c|}{\ding{55}} & \multicolumn{2}{c}{40.00} \\
\multicolumn{2}{c}{\ding{55}} & \multicolumn{2}{c|}{\checkmark} & \multicolumn{2}{c}{48.00} \\
\multicolumn{2}{c}{\checkmark} & \multicolumn{2}{c|}{\checkmark} & \multicolumn{2}{c}{\textbf{53.00}} \\
\bottomrule[1pt]
\end{tabular}
\label{tab: ablation}
\vspace{-5mm}
\end{wraptable}

\textbf{Data recipe is critical for SFT training.}
Ablation results in Tab.~\ref{tab: ablation} reveal that combining 2D, 3D, and simulated data yields optimal performance.
As noted in Sec.~\ref{subsec: dataset}, 2D data spans indoor/outdoor scenes, enabling depth learning across scales; its removal severely degrades performance on outdoor-centric BLINK~\cite{fu2024blink}.
Meanwhile, 3D data captures embodied indoor environments and mitigates the Sim2Real gap, benefiting indoor-focused CV-Bench~\cite{tong2024cambrian}.
Finally, simulated data broadens spatial diversity. 
This tripartite data composition is thus key to effective SFT training.

\textbf{Dedicated depth encoder preserves image understanding.}
We compare dedicated and shared image-depth encoders during SFT.
In Tab.~\ref{tab: general}, the dedicated encoder better maintains image understanding under limited RGB-only data (1/20 {\dname} QA), while the shared encoder harms general performance.
Though prior work\cite{cheng2024spatialrgpt} adopts a shared encoder, it 
\textbf{(1)} requires over twice as much RGB-only data compared to spatial-related data for co-training; 
\textbf{(2)} \textit{targets region-level depths, differing from our full-image approach}.



\textbf{Depth encoder improves both spatial understanding and reasoning.}
%
%
Recent VLMs~\cite{team2025gemini, daxberger2025mm} show that large-scale spatial training enables implicit 3D understanding (\eg, depth, distance, 3D boxes) from images alone.
To assess this, we fine-tune NVILA-2B~\cite{liu2024nvila} on {\dname} without the depth encoder, followed by continued RFT.
Results indicate that depth improves single-step spatial understanding, consistent with MM-Spatial~\cite{daxberger2025mm}, and yields greater gains in multi-step spatial referring.
We attribute this to: 
\textbf{(1)} the need for precise coordinate prediction in spatial referring, unlike VQA’s multiple-choice;
\textbf{(2)} cumulative reasoning across steps, amplifying the utility of depth cues.

\textbf{Process reward advances the accuracy of intermediate perception.}
Tab~\ref{tab: ablation} shows a 5-point improvement with process reward, which leverages key step annotations from {\dname} to refine step-wise perception, thereby predicting more accurate points with complex spatial relations.
%

%% file: sec/5_conclusion.tex
\section{Conclusion and Future work}
\label{sec: conclusion}

In this paper, we introduce {\mname}, a novel 3D-aware VLM that addresses spatial referring through the combination of both single-step accurate understanding and multi-step spatial reasoning. 
In detail, we enhance 3D perception with a separate depth encoder via SFT, and enable generalized multi-step spatial referring via RFT with our proposed metric-sensitive process reward functions.
We also present {\dname}, a large-scale, well-designed dataset for SFT and RFT training, with {\bname}, a benchmark tailored to evaluate spatial referring.
Extensive experiments show the effectiveness of {\mname} and highlight its potential for a broad range of robotic applications.

\noindent Our future work will focus on two main directions.
\textbf{(1)} Enhancing the model's understanding of human priors and intent. As discussed in Appx.~\ref{suppsec: limitation}, human instructions are often brief and ambiguous, even when the correct location is unique.
Potential solutions include exploring procedural synthesis of intent-aware data or improving model performance through co-training with intent-rich datasets.
\textbf{(2)} Improving the model’s 3D perception capabilities. Our current models predominantly rely on qualitative spatial relations (\eg, left, right) and predict 2D image-plane coordinates, necessitating depth-based conversion to 3D, as discussed in Appx.~\ref{suppsec: discussion}. 
Future directions include directly modeling quantitative geometry to enable precise 3D reasoning or direct prediction of 3D points and visual traces, which are more challenging if combined with spatially constrained instructions.



\section*{Acknowledgement}
This work was supported by the National Natural Science Foundation of China (62132001, 62476011), Capital's Funds for Health Improvement and Research (CFH 2024-2-40611), and the Fundamental Research Funds for the Central Universities.
We sincerely thank Jiayuan Zhang, Jiawei He for their valuable discussions and insightful feedback about the method.
%
%
We also sincerely appreciate Yusu Deng’s excellent figure design (\eg, teaser figure, pipeline overview) and demo reviewing work.

%% file: sec/Appendix.tex
\clearpage
\appendix

\toggletrue{inappendix} 

\renewcommand{\contentsname}{Appendix Table of {\mname}}
\tableofcontents

\section{Discussion}
\label{suppsec: discussion}

\noindent\textbf{Distinction from SpatialRGPT.}
Our approach differs from SpatialRGPT in 4 key aspects:
\textbf{(1) Task Setting}: Our model addresses a more challenging spatial referring task, where takes a spatially constrained textual instruction as input. It requires multi-step spatial reasoning with learned spatial knowledge to precisely localize the referred object as a 2D point step-by-step. In contrast, SpatialRGPT addresses a simpler VQA task and relies on externally provided region information as input for specific object referring.
\textbf{(2) Model Usage}: Unlike SpatialRGPT, which needs additional masks or detection tools to generate masks or 2D boxes as inputs for object reference and simplify referring tasks, our model can use textual descriptions for object referencing (see L283), which better aligns with real-world robotic applications.
\textbf{(3) Data Pipeline}: Our data pipeline adopts a more structured, progressive design than SpatialRGPT. It first uses 2D image data to teach core spatial concepts and general depth perception across diverse indoor and outdoor scenes. Next, accurate 3D data enhances fine-grained spatial understanding in indoor settings for robotics. Finally, simulation data introduces multi-step spatial referring with reasoning. This staged approach yields stronger spatial understanding and reasoning than SpatialRGPT, which relies solely on data generated from 2D images and lacks precise spatial perception for more complex spatial referring tasks.
\textbf{(4) Training Pipeline}: Our training pipeline includes process-based RFT after SFT, further to improve multi-step reasoning and generalization for spatial referring tasks, whereas SpatialRGPT is trained with SFT only.

\noindent\textbf{Justification for RFT.}
RFT brings two main benefits:
\textbf{(1) Generalization to Unseen Cases}: In Tab.~\ref{tab: referring} in the main paper (RefSpatial-Bench Unseen raw), which features novel combinations of spatial relations absent from {\dname} dataset, our 2B-RFT model surpasses 2B-SFT by 9.1\% in accuracy, showing the strong generalization enabled by the RFT stage.
\textbf{(2) Enhance Multi-Step Reasoning Ability}: In the Tab.~\ref{supptab: reasoning step}, the RFT-based model consistently outperforms the SFT-based model across varying reasoning steps, especially at larger steps, showing the RFT stage's effectiveness in enhancing multi-step reasoning.

\noindent\textbf{Why is NVILA chosen as the backbone?}
In Tab~\ref{tab: understanding} in the main paper, NVILA outperforms other open-source VLMs under comparable model scales, such as Qwen 2.5-VL (even 72B), in spatial understanding. 
Enhancing a strong baseline with our dataset and training strategy further validates their effectiveness. 
Notably, our dataset is model-agnostic and transferable to other backbones. Despite partial training on the {\dname} dataset, Qwen2.5-VL-7B still shows notable improvements on spatial understanding benchmarks in the Tab.~\ref{supptab: qwen} below.

\noindent\textbf{Depth-to-3D mapping assumption.}
The depth-to-3D mapping assumption is essential in our real-world evaluation, as our model predicts only 2D image-plane points, while real-world tasks typically require 3D coordinates for grasping, placement, or navigation.
While depth noise and partial observations are important real-world challenges, our setting follows prior work~\cite{ji2025robobrain, yuan2024robopoint}, which assumes that accurate depth-to-3D mapping is feasible given known camera intrinsics and extrinsics—sufficient for common manipulation and navigation tasks. 
Moreover, these challenges can be effectively mitigated via the following strategies:
(1) Depth noise can be mitigated by recent advances in monocular depth estimation~\cite{piccinelli2025unidepthv2}, monocular geometry prediction~\cite{wang2025moge}, and stereo methods~\cite{wen2025foundationstereo}. In cases of severe noise, we employ FoundationStereo~\cite{wen2025foundationstereo} in real-world settings to mitigate this issue.
(2) Partial views are mitigated in our method by leveraging pixel-level target points. Further improvement is possible by incorporating RoboRefer as a spatially-aware planner for active perception.

\noindent\textbf{The effect of depth noise on the model's accuracy and robustness.}
In real-world experiments, we utilize a relative depth estimation model, DepthAnything v2, to obtain relative depth as the model's depth input, thereby effectively reducing depth noise from a real camera.
We also evaluate success rates under depth noise in real-world settings (see Tab below). 
Depth maps generated from the strong monocular relative depth estimation model (\ie, DepthAnything V2) offer the highest robustness and success. 
Despite depth noise from a real camera, RoboRefer maintains great performance by leveraging RGB priors due to mixed RGB and RGB-D training during the SFT stage.

\begin{table}[htbp]
\centering
\footnotesize
\caption{We report the success rates (\%) of 2B-SFT and 2B-RFT model at each reasoning step on {\bname}.}
\begin{tabular}{lcccc}
\toprule
\textbf{Benchmark} & \textbf{Reasoning Step Num.} & \textbf{2B-SFT} & \textbf{2B-RFT} & \textbf{Gain} \\
\midrule

\multirow{4}{*}{RefSpatial-Bench-Location}
  & Step 1 & 63.33 & 66.67 & +3.34 \\
  & Step 2 & 39.58 & 43.75 & +4.17 \\
  & Step 3 & 27.27 & 36.36 & +9.09 \\
  & \textbf{Total} & \textbf{47.00} & \textbf{52.00} & \textbf{+5.00} \\
\midrule

\multirow{5}{*}{RefSpatial-Bench-Placement}
  & Step 2 & 55.56 & 55.56 & +0.00 \\
  & Step 3 & 41.67 & 41.67 & +0.00 \\
  & Step 4 & 41.67 & 45.83 & +4.16 \\
  & Step 5 & 0.00  & 25.00 & +25.00 \\
  & \textbf{Total} & \textbf{48.00} & \textbf{54.00} & \textbf{+6.00} \\
\bottomrule
\end{tabular}
\label{supptab: reasoning step}
\end{table}

\begin{table}[t]
\caption{Performance on the \textit{single-step spatial understanding} benchmarks across different model types.
Top-1 \& Top-2 accuracies are represented using \textbf{bold text}, and \underline{underlines}.}
\centering
\scriptsize
\setlength{\tabcolsep}{1.5pt}
\begin{tabular}{l|ccc|cc|ccc}
\toprule
\multirow{2}{*}{Method}              & \multicolumn{3}{c|}{CV-Bench~\cite{tong2024cambrian}}       & \multicolumn{2}{c|}{$\text{BLINK}_{val}$~\cite{fu2024blink}} & \multirow{2}{*}{RoboSpatial~\cite{song2024robospatial}} & \multirow{2}{*}{SAT~\cite{ray2024sat}} & \multirow{2}{*}{EmbSpatial~\cite{du2024embspatial}} \\
                                   & 2D-Relation & 3D-Depth & 3D-Distance & 2D-Relation & 3D-Depth &   &  & \\
\midrule

Qwen-2.5-VL-7B (base)        & 82.15 & 60.17 & 69.00 & 64.34 & 60.98 & 49.59 & 30.00 & 40.20\\
Qwen-2.5-VL-7B (finetuned)        & 95.85&	95.00&	90.83&	83.22&	84.68&	69.92&	85.75&	76.32\\
\midrule
NVILA-8B (base)              & 91.54 & 91.83 & 90.67 & 76.92 & 76.61 & 59.35 & 63.33 & 67.72\\
RoboRefer-8B-SFT (finetuned)              & 96.90&	98.33&	93.50&	91.61&	92.74&	84.55&	86.67&	72.53\\

\bottomrule[1pt]
\end{tabular}
\label{supptab: qwen}
\end{table}

\begin{table}[t]
\centering
\caption{We report the success rates (\%) of real-world evaluation performance when using depth from DepthAnything V2 and Real Camera.}
\scriptsize
\setlength{\tabcolsep}{1.5pt}
\begin{tabular}{lcc}
\toprule
\textbf{Real-world Task} & \textbf{Depth from DepthAnything V2} & \textbf{Depth from a Real Camera} \\
\midrule
Pick the specific hamburger closest to the mug nearest to the camera. & 80 & 70 \\
Place the hamburger in front of the teddy bear. & 90 & 90 \\
Pick the apple in front of the leftmost cup’s logo side. & 80 & 80 \\
Place the apple aligned with the existing apple row. & 60 & 40 \\
\bottomrule
\end{tabular}
\label{supptab: depth}
\end{table}

\section{Implementation Details and Samples of {\dname} Dataset}
\label{suppsec: dataset}

In this section, we present a comprehensive exposition of the implementation details and representative data samples underpinning the construction of the \dname{} dataset. 
As this dataset is intended to equip general VLMs with the ability to adapt to spatial referring tasks, thereby enhancing spatial understanding and reasoning in a bottom-up manner, we meticulously design a multi-data-source generation pipeline. 
We elaborate on the three core components of this pipeline as follows:

    \textbullet\ \textbf{2D Web Image~(Appx.~\ref{suppsubsec: 2D Web Image})}: We present a 2D data pipeline comprising image filtering, pseudo-3D scene graph construction, hierarchical referential description generation—from coarse categories to fine-grained spatial referents—and diverse QA pair creation.
    
    \textbullet\ \textbf{3D Embodied Video~(Appx.~\ref{suppsubsec: 3D Embodied Video})}: This section outlines the 3D data selection process from CA-1M~\cite{lazarow2024cubify}, discusses its limitations and mitigation strategies, and presents methods for enriched scene graph construction compared to the 2D data source. We further describe a QA generation framework that leverages detailed 3D annotations (\eg, depth maps, oriented 3D bounding boxes) to capture richer spatial relations. Finally, we detail how to generate QA pairs for the problem of ``feasibility assessment for object placement in free space''.
    
    \textbullet\ \textbf{Synthetic Data from Simulator~(Appx.~\ref{suppsubsec: Synthetic Data Generation})}: We describe how to synthesize 3D scenes, select and annotate digital assets, efficient scene assembly and rendering, and the generation of QA pairs grounded in these simulated scenes.

In the following subsections of each section, we detail the employed models, prompt design rationale, data processing steps, filtering criteria, and illustrative examples, providing a clear and thorough overview of the construction pipeline and core technical details of the \dname{} dataset.


\subsection{2D Web Image}
\label{suppsubsec: 2D Web Image}
\subsubsection{Multi-Stage Image Filtering}
\label{subsubsec:multi_stage_filtering}

2D Web Images aim to endow the model with basic spatial concepts and comprehensive depth perception across both indoor and outdoor scenes. Here we use OpenImage\footnote{\href{https://storage.googleapis.com/openimages/web/index.html}{https://storage.googleapis.com/openimages/web/index.html}}~\cite{kuznetsova2020open} as 2D data source.

\textbf{Overall Motivation and Goals for Filtering.}
%
%
%
The OpenImages dataset offers a vast collection of 2D internet images~(1.7 M images in training split) with extensive visual diversity, but many images, such as text-only graphics, QR codes, medical scans, or abstract art, lack relevance for spatial understanding and referring with reasoning. 
To curate a dataset tailored for these tasks, we employ a two-stage filtering pipeline: a coarse pre-filtering using SigLIP2~\cite{tschannen2025siglip}, followed by fine-grained selection via Qwen2.5-VL~\cite{bai2025qwen2}, to retain images rich in spatial semantics.
%


\textbf{Stage 1: Initial Coarse Filtering.}
%
%
\highlight{We employ the \texttt{siglip2-giant-opt-patch16-384} model for initial filtering to efficiently discard low-quality or off-theme images (\eg, irrelevant scenes or content lacking multiple everyday objects).}
This step greatly reduces data volume, streamlining subsequent processing.
Specifically, the SigLIP2 model is guided by predefined positive and negative textual labels.
Positive labels represent desired image characteristics, while negative labels describe undesired content. 
For each image, the model computes cosine similarity between its embedding and all label embeddings. The label with the highest similarity is selected; if it belongs to the positive set, the image is retained, otherwise discarded.
Label sets are manually refined iteratively to balance recall and precision, ensuring relevance while excluding noise. 
These labels act as semantic anchors for visual-text alignment. 
In this stage, only $934$k images are qualified to be retained.
Positive and negative label lists are provided in Listings~\ref{lst:siglip_positive_labels} and~\ref{lst:siglip_negative_labels}.
For more details about the compute resources needed in this stage, please see Appx.~\ref{suppsubsec: compute resources}.

\begin{lstlisting}[basicstyle=\ttfamily\footnotesize, backgroundcolor=\color{myblue!50}, caption={Positive Labels used during SigLIP2 filtering.}, captionpos=t, breaklines=false, label={lst:siglip_positive_labels}]
Positive Labels = [
    "Mid-distance observation of some objects on a table",
    "Some objects on the desktop",
    "Distant view of some animals",
    "Mid-distance observation of some animals",
    "Distant view of one object",
    "Mid-distance observation of one object",
    "Distant view of some objects",
    "Mid-distance observation of some objects",
    "Distant view of a person",
    "Mid-distance observation of a person",
    "Distant view of some people",
    "Mid-distance observation of some people",
    "Distant view of indoor scene",
    "Distant view of outdoor scene",
    "Distant view of traffic",
    "Distant view of Urban architecture"
]
\end{lstlisting}

\begin{lstlisting}[basicstyle=\ttfamily\footnotesize, backgroundcolor=\color{myblue!50}, caption={Negative Labels used during SigLIP2 filtering.}, captionpos=t, breaklines=false, label={lst:siglip_negative_labels}]
Negative Labels = [
    "Macro shot of an animal",
    "Macro shot of one object",
    "Macro shot of a person",
    "Macro shot of flowers",
    "A piece of text",
    "A person displayed in front of a white background",
    "A product displayed in front of a white background",
    "A screenshot of the graphics user interface",
    "A dimly lit environment"
]
\end{lstlisting}

SigLIP2 preserves images rich in object diversity, depth cues, and scene context (indoor/outdoor) through the above labeling process, but its performance declines on certain image types, including:

\begin{enumerate}
\item \textbf{Paintings/Artworks}: Especially those with visible brushstrokes or canvas textures.
\item \textbf{Low-Light Scenes}: Images with minimal illumination and strong shadows.
\item \textbf{Grayscale Photographs}: Black-and-white imagery lacking color cues.
\item \textbf{Distorted Images}: Those exhibiting warping, mirroring, or other geometric anomalies.
\item \textbf{Multi-Scene Collages}: Images containing three or more distinct scenes with hard borders.
\end{enumerate}

SigLIP2 struggles to detect and interpret these categories reliably, highlighting the need for a secondary, fine-grained filtering stage.


\textbf{Stage 2: Fine-grained Filtering}
\label{par:qwen_vl_filtering_stage}
Due to SigLIP2's limitations in handling certain visual content mentioned above, 
\highlight{we introduce a fine-grained filtering stage using the Qwen2.5-VL-7B model to further improve dataset quality. 
This step ensured the final images are clear, authentic, and well-suited} for spatial understanding and reasoning required for spatial referring tasks.
Qwen2.5-VL processed $934$k images filtered by SigLIP2, retaining $846$k. 
Although Qwen2.5-VL offers higher filtering precision, its slower speed necessitated the use of SigLIP2 for initial fast filtering, significantly improving overall efficiency.
%
%
%
%
To ensure accurate and consistent fine-grained filtering, we adopt a structured prompt engineering strategy for the Qwen2.5-VL model. 
The process begins with a system prompt (See Listing~\ref{lst:qwen_system_prompt}) that defines the model’s role as an image analysis expert, specifying key visual attributes to assess and negative categories to detect, and enforcing a strict workflow.
For each image, a corresponding user prompt (See Listing~\ref{lst:qwen_user_prompt}) instructs the model to determine whether the image belongs to any predefined negative categories. 
The model’s response follows a structured format: if the segment after the pipe symbol (|) is “Yes”, the image is classified as negative and discarded; otherwise, it is retained.
This prompting scheme ensures that the model adheres to a consistent output, enhancing the reliability of filtering outcomes.
Please refer to Appx.~\ref{suppsubsec: compute resources} for details on the computational requirements of this stage.

\begin{lstlisting}[basicstyle=\ttfamily\footnotesize, backgroundcolor=\color{myblue!50}, caption={System Prompt for Qwen2.5-VL-7B filtering.}, captionpos=t, breaklines=true, label={lst:qwen_system_prompt}]
system_prompt = """
You are an image analysis expert. Follow this workflow rigidly:

1. **Content Analysis**:
   - Inspect: Main subjects, artistic style, visual characteristics
   - Check: Lighting intensity, color channels, geometric integrity, composition structure

2. **Category Verification** (YES if matches ANY):
   a) Painting/Artwork - Visible brushstrokes/canvas texture
   b) Dim Lighting - Very low brightness, heavy shadows
   c) B&W Photo - Grayscale only (0 color channels)
   d) Distorted Image - Warping/mirroring anomalies
   e) Multi-image Collage - >=3 distinct scenes with hard borders

3. **Structured Response**:
   Output EXACTLY in this format:
   "[Analysis sentence]. | Yes/No"
   - Analysis must contain observable evidence
   - Final answer MUST use pipe separator

Examples of VALID responses:
    "This image is a composite created by stitching together multiple smaller images, with distinct white borders visible between the individual components. | Yes"
    "This image features vibrant colors, is neither an artistic painting nor a composite of multiple images, and does not conform to any of the specified categories. | No"
"""
\end{lstlisting}

\begin{lstlisting}[basicstyle=\ttfamily\footnotesize, backgroundcolor=\color{myblue!50}, caption={User Prompt for Qwen2.5-VL-7B filtering.}, captionpos=t, breaklines=true, label={lst:qwen_user_prompt}]
user_prompt = """
Analyze if this image belongs to ANY of these categories:
1. Painting/artwork
2. Dim lighting
3. Black-and-white
4. Geometric distortion
5. Multi-image collage

Respond EXACTLY FORMATTED as:
"[Your evidence-based analysis]. | Yes/No"
"""
\end{lstlisting}

\begin{figure}[h!]
  \centering
  \includegraphics[width=\linewidth]{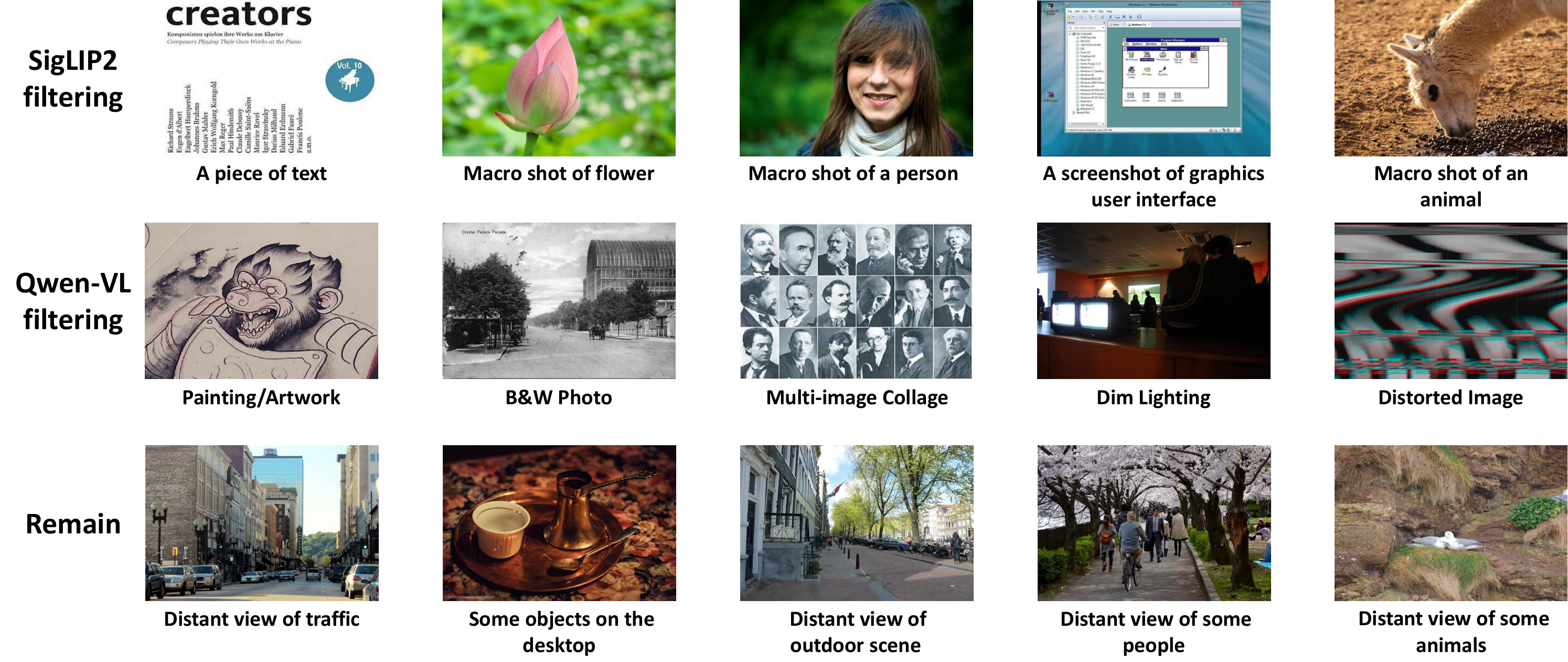} 
  \caption{
  Visual overview of the multi-stage filtering results. Row 1: Images discarded by SigLIP2 due to insufficient spatial context (\eg, close-ups, text). Row 2: Additional filtering by Qwen2.5-VL removes non-natural content (\eg, artwork, collages). Row 3: Remaining high-quality images suitable for spatial understanding and referring.
  }
  \label{fig:filtering_stages_comparison}
\end{figure}

\textbf{Visualizing Overall Filtering Results.}
\label{par:filtering_visualization}
Fig.~\ref{fig:filtering_stages_comparison} presents a visual overview, showing the effectiveness of our multi-stage filtering pipeline.
The first row shows the output of SigLIP2, which effectively removes images lacking spatial semantics, such as macro shots of animals, people, or flowers, textual content, and GUI screenshots. 
The second row demonstrates how Qwen2.5-VL-7B further eliminates unsuitable categories, including artworks, dimly lit scenes, black-and-white images, geometrically distorted content, and image collages. 
The third row displays the retained images after both stages, which exhibit rich spatial relationships, confirming their suitability for spatial understanding and reasoning and the overall quality of our dataset.


\subsubsection{Pseudo-3D Scene Graphs Construction}
\label{supsubsubsec: 3D Scene Graph Construction}

Although filtered 2D images contain some spatial cues, QA pairs containing sufficient 3D spatial information~(\eg, ``near''\textit{ vs.} ``far'', distances) derived directly from these 2D images are challenging.
Inspired by prior work~\cite{cheng2024spatialrgpt, chen2024spatialvlm}, we construct pseudo-3D scene graphs from 2D images to enhance the generation of QA pairs with rich 3D spatial semantics.
In these graphs, nodes represent object attributes, while edges encode inter-object spatial relations.
We detail the process of converting 2D images into pseudo-3D scene graphs below.

%
%
%
\textbf{Object Detection and Annotation.}
\label{par:detection_annotation_groundingdino_ram} 
Although the OpenImages dataset provides annotations, its limited vocabulary and coarse labeling are insufficient for open-world scenarios. 
To address this, we leverage state-of-the-art foundation models for enhanced object detection and annotation. 
Specifically, our scene graph construction pipeline integrates the Recognize Anything Model (RAM)~\cite{zhang2024recognize} and GroundingDINO~\cite{liu2024grounding} to assign semantic labels and bounding boxes to key objects in filtered raw 2D images.
The workflow is broadly as follows:
\begin{enumerate}
    \item \textbf{Semantic Labeling via RAM}: RAM analyzes each image to generate category labels for all recognized objects. Its broad recognition capability ensures comprehensive semantic coverage, guiding subsequent localization.
    
    \item \textbf{Bounding Box Localization via GroundingDINO}: The labels from RAM are used as text prompts for GroundingDINO, an open-vocabulary detector that localizes the corresponding objects and outputs precise bounding boxes.
\end{enumerate}


%
%
%
%
%

\begin{figure}[h!]
  \centering
  \includegraphics[width=\linewidth]{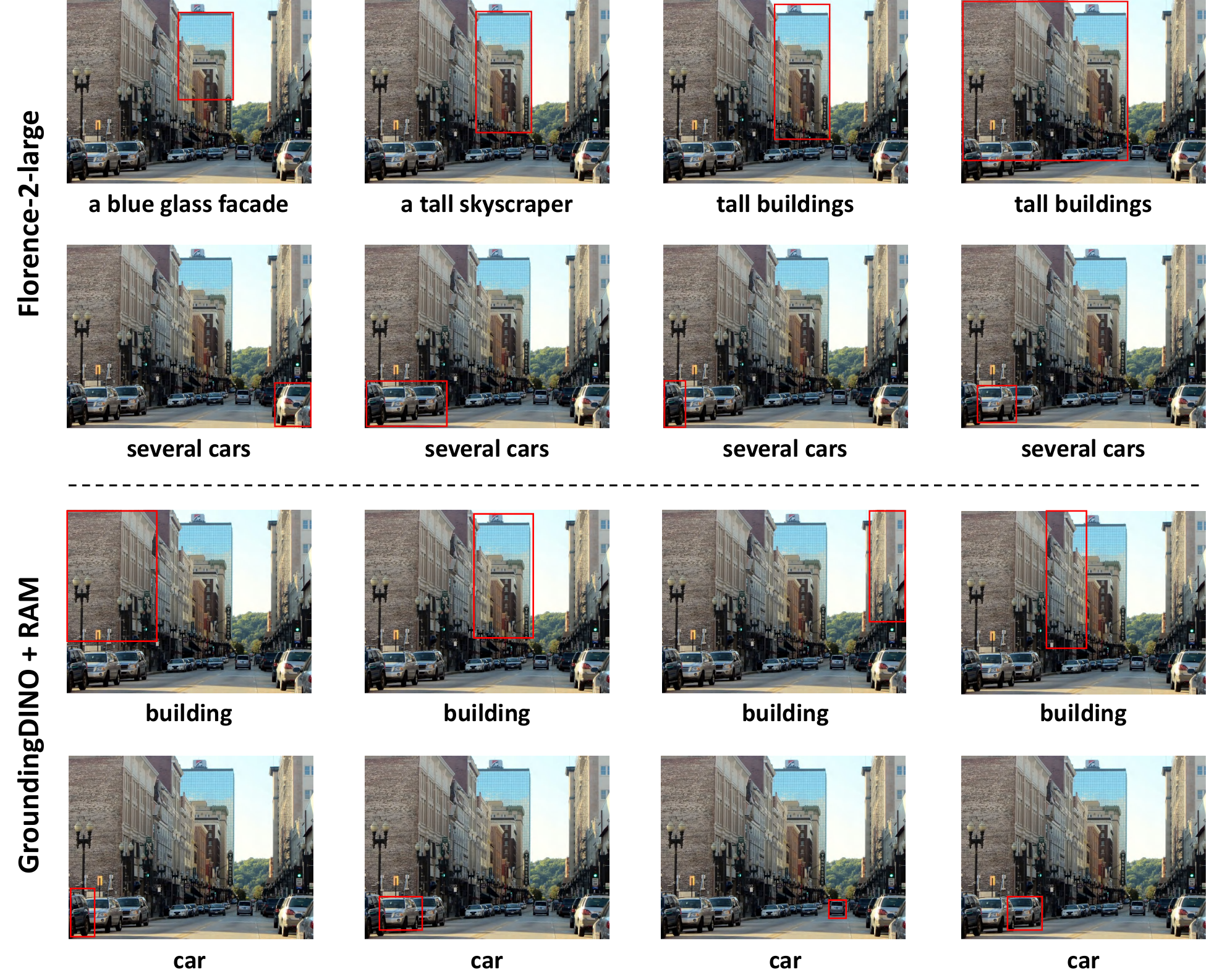} 
  \caption{Object detection comparison: Florence-2 (above), GroundingDINO+RAM (below)}
  \label{suppfig: detection_comparison}
\end{figure}

\highlight{Although recent VLM, \ie, Florence-2-Large supports instruction-based recognition and detection simultaneously, we find that combining RAM with GroundingDINO yields superior results.}
In Fig.~\ref{suppfig: detection_comparison}, Florence-2 (top) often produces ambiguous or redundant detections (\eg, vague labels, multiple boxes for a single object, or single boxes covering multiple objects), which are unsuitable for precise object referring. 
In contrast, GroundingDINO+RAM (bottom) generates concise labels and one-to-one object-bounding-box mappings, better satisfying the requirements of referring tasks.


\begin{figure}[h!]
  \centering
  \includegraphics[width=1\textwidth]{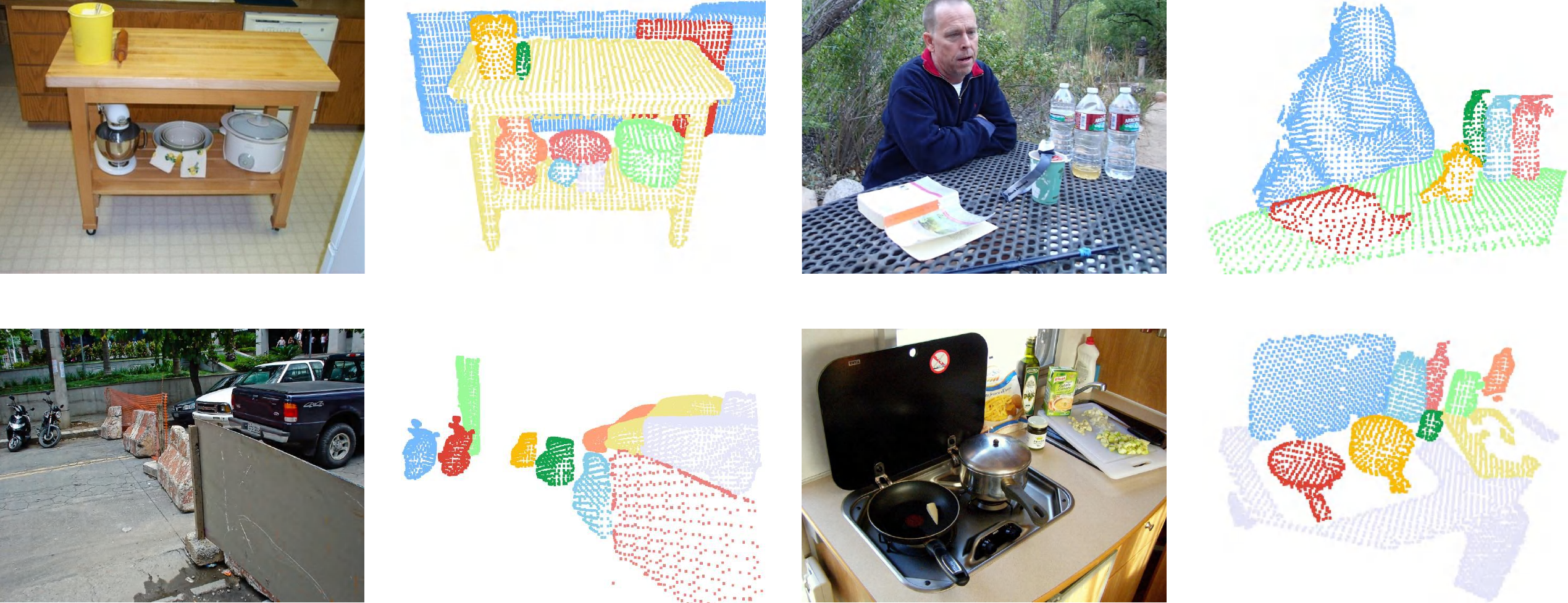}
  \caption{Scene graph visualization with image, detected objects, and corresponding point clouds.}
  \label{suppfig: scene_graph_visualization}
\end{figure}

\textbf{3D-aware information Extraction.}
To further extract 3D-aware information from 2D images, we adopt UniDepth V2~\cite{piccinelli2025unidepthv2} for metric depth estimation due to its recent state-of-the-art performance, surpassing models such as DepthPro and Metric3Dv2 across multiple benchmarks. 
For camera intrinsic prediction, we employ WildeCamera~\cite{zhu2023tame}. 
Together, these models enable robust 3D point cloud reconstruction of the scene. 
Based on previously annotated object bounding boxes, we apply SAM 2.1~\cite{ravi2024sam} to generate instance masks. 
Each resulting Pseudo-3D scene graph comprises object labels (via RAM), 2D bounding boxes (via GroundingDINO), instance masks (via SAM 2.1), and object-level point clouds (via UniDepth v2 with WildeCamera), resulting in \highlight{axis-aligned 3D bounding boxes}.
A visualization is provided in Fig.~\ref{suppfig: scene_graph_visualization}.


\subsubsection{Hierarchical Object Description Generation}
\label{suppsubsubsec: object_description_generation}
While 3D scene graphs encode basic object categories, real-world scenes often contain multiple instances per category. 
Prior datasets~\cite{daxberger2025mm,song2024robospatial} simplify reference by assuming a single object per category, limiting their utility in spatial referring tasks. 
To overcome this, \highlight{we augment object descriptions with attributes and spatial relations, enabling finer-grained disambiguation among similar instances of the same category.}
We present our two-stage generation pipeline below.

\textbf{Stage 1: Generating Object and Image Dense Descriptions in image space.}
We first generate detailed descriptions for each detected object and the entire image, employing the Qwen2.5-VL-7B model. 
This stage involves both object-level captioning and comprehensive image-level captioning.
The global captions are essential for providing contextual grounding to downstream large language models (QwQ-32B) during LLM QA generation (detailed in Appx.~\ref{supppar:qwq_diversification_2d}), enhancing the relevance and accuracy of the outputs.
Prompt templates are detailed in Listings~\ref{lst:qwen_image_capt_prompts} and~\ref{lst:qwen_object_capt_prompts}. 
In particular, the \texttt{object\_caption\_user\_text\_prompt} uses a dynamic placeholder \texttt{[class\_name]}, which is filled with the object category predicted by the RAM model (See Appx.~\ref{supsubsubsec: 3D Scene Graph Construction}).



\begin{lstlisting}[basicstyle=\ttfamily\footnotesize, backgroundcolor=\color{myblue!50}, caption={Prompts for Image Caption Generation with Qwen-VL.}, captionpos=t, breaklines=true, label={lst:qwen_image_capt_prompts}, escapeinside={(*@}{@*)}]
image_caption_system_text_prompt =  """
    You are an expert image analysis assistant. Your task is to generate a detailed and comprehensive description of the image.
    Please focus on accurately capturing all visual elements present in the image, including objects, scenery, colors, shapes, textures, and lighting.
    Your description should be clear, precise, and professional. Additionally, ensure that your description begins with either `this image' or `the image'.
"""

image_caption_user_text_prompt = """
    Please carefully examine the provided image and generate a detailed description.
    Include all visible elements such as objects, scenery, colors, shapes, textures, and lighting.
    Ensure that your description is thorough, accurate, and complete, and that it starts with either `this image' or `the image'.
"""
\end{lstlisting}

\begin{lstlisting}[basicstyle=\ttfamily\footnotesize, backgroundcolor=\color{myblue!50}, caption={Prompts for Object Caption Generation with Qwen-VL.}, captionpos=t, breaklines=true, label={lst:qwen_object_capt_prompts}]
object_caption_system_text_prompt = """
    You are a visual localization analyzer working with TWO distinct images:
    1. [POSITION-REFERENCE] (First Image):
    - Contains ONLY location clues with background
    - Strictly use ONLY for determining spatial position (left/right/upper/lower/center)
    - Ignore all visual features except object placement

    2. [DETAIL-SOURCE] (Second Image):
    - Shows the object's TRUE APPEARANCE without background
    - Extract EXCLUSIVELY from this: color, texture, material, shape
    - Never infer details from the first image

    Generate phrase in pattern: [Color][Material][Object] at [Position]
    Example: "Matte black laptop on the left" NOT "Red-boxed laptop"
"""

object_caption_user_text_prompt = """
    For the [class_name] marked by red box in FIRST image and fully shown in SECOND image:
    -> COLOR/MATERIAL: Must come from SECOND image
    -> POSITION: Only from FIRST image's placement
    Forbidden actions:
    x Mention 'red box' or background elements
    x Use location terms in second image
    x Combine features across images

    Describe the [class_name] marked by red box in FIRST image and fully shown in SECOND image with this format:
    [Color][Material/Texture][Object] at [Position]
    Samples:
    - "Brushed metal water at bottle left"
    - "Glossy ceramic mug at upper center"
    - "Faded denim jacket at lower right"
"""
\end{lstlisting}

\textbf{Stage 2: Generating Object Description with Spatial Cues.}
\label{par:heuristic_obj_description}
To enhance referential specificity in object captions, particularly when multiple instances of the same category coexist, we adopt a heuristic strategy that appends spatially indicated relative information, such as ``the third chair from the front''.
This method leverages 3D object positions from the scene graph (See Appx.~\ref{supsubsubsec: 3D Scene Graph Construction}). 
By comparing same-category objects along the three principal axes (front–back, left–right, top–bottom), we identify the axis with the largest spatial variation as the primary arrangement direction to guide relative spatial reference generation.
%
%
%
%
Once the main sorting axis is identified, we retrieve appropriate templates from a predefined library (see Listing~\ref{lst:spatial_order_templates}) to augment the initial object descriptions. 
These templates are designed to capture diverse natural language patterns. 
For instance, for a row of chairs arranged left to right, templates may include: ``\{dense\_caption\}, which is the \{ordinal\} \{class\_name\} from left to right,'' or ``\{dense\_caption\}, the \{ordinal\} \{class\_name\} in the left-to-right sequence''.
Here, \texttt{{dense\_caption}} denotes the initial description generated by the Qwen2.5-VL model, \texttt{{ordinal}} indicates the object’s position in the sorted sequence, and \texttt{{class\_name}} is the category label predicted by RAM.
This spatially-aware enhancement is applied only when multiple instances of the same category are detected to avoid redundancy. 
If an object appears only once, its original dense caption is used directly. 
To ensure spatial diversity, we set a variance threshold across the three principal axes; images with multiple same-category objects but low variance on all axes are discarded, resulting in a final set of $466$k images.
%
%
By integrating spatial ordering with visual descriptions, this heuristic enables the generation of precise and discriminative referential expressions, essential for producing high-quality, unambiguous question-answer pairs.

\begin{lstlisting}[basicstyle=\ttfamily\footnotesize, backgroundcolor=\color{myblue!50}, caption={Templates for Spatial Order Description Enhancement.}, captionpos=t, breaklines=true, label={lst:spatial_order_templates}]
TEMPLATES = {
    "left_to_right": [
        "{dense_caption}, which is the {ordinal} {class_name} from left to right",
        "{dense_caption}, marked as the {ordinal} {class_name} in a left-to-right arrangement",
    ],
    "right_to_left": [
        "{dense_caption}, the {ordinal} {class_name} viewed from the right",
        "{dense_caption}, the {ordinal} {class_name} from the right",
    ],
    "front_to_back": [
        "{dense_caption}, which appears as the {ordinal} {class_name} when viewed from the front",
        "{dense_caption}, positioned as the {ordinal} {class_name} in front-to-back order",
    ],
    "back_to_front": [
        "{dense_caption}, which is counted as the {ordinal} {class_name}, starting from the back",
        "{dense_caption}, the {ordinal} {class_name} in the back-to-front sequence",
    ],
    "top_to_bottom": [
        "{dense_caption}, the {ordinal} {class_name} viewed from the top",
        "{dense_caption}, placed as the {ordinal} {class_name} when sorted from top to bottom",
    ],
    "bottom_to_top": [
        "{dense_caption}, which ranks as the {ordinal} {class_name} in bottom-to-top order",
        "{dense_caption}, arranged as the {ordinal} {class_name} when ordered from the bottom",
    ]
}
\end{lstlisting}

\begin{figure}[h!]
  \centering
  \includegraphics[width=\linewidth]{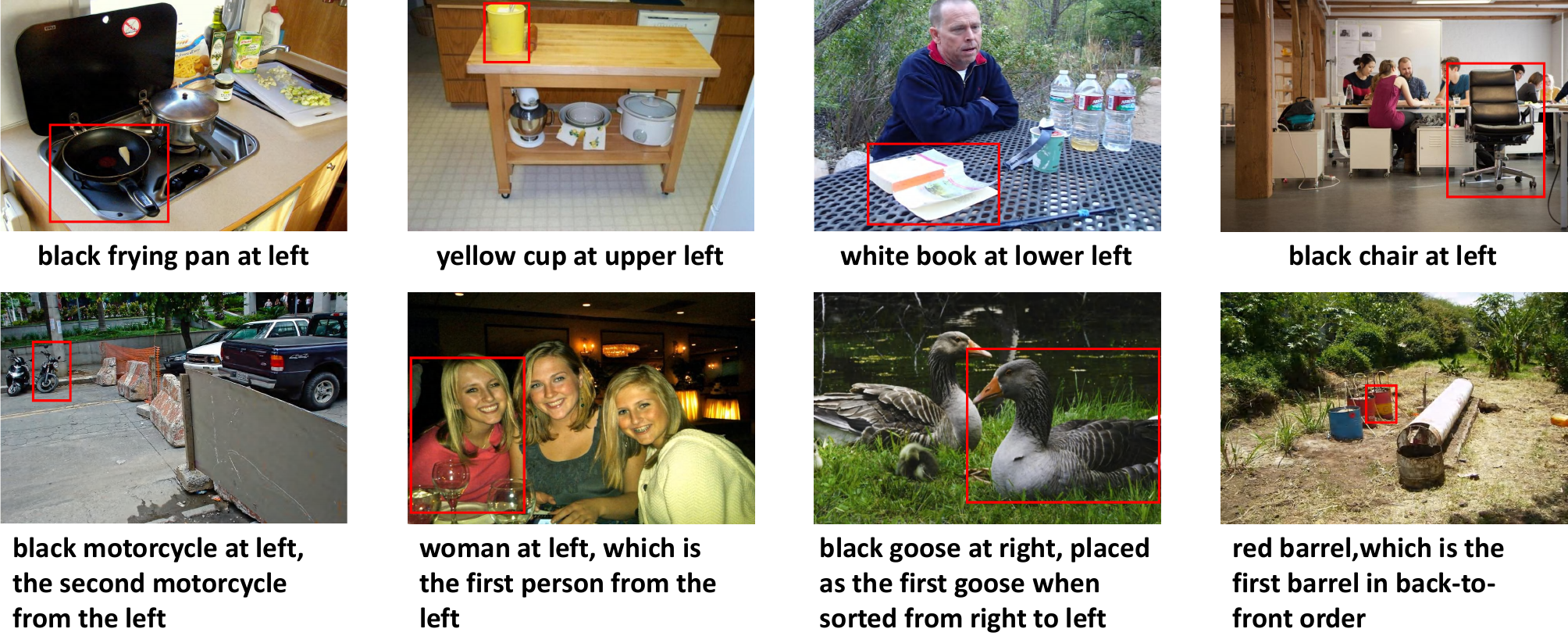} 
  \caption{Generated object descriptions. Top: Unique-category captions. Bottom: Spatially-aware captions for the same categories. Red boxes indicate referenced objects.}
  \label{fig:caption_visualization_example}
\end{figure}

\begin{figure}[h!]
  \centering
  \includegraphics[width=\linewidth]{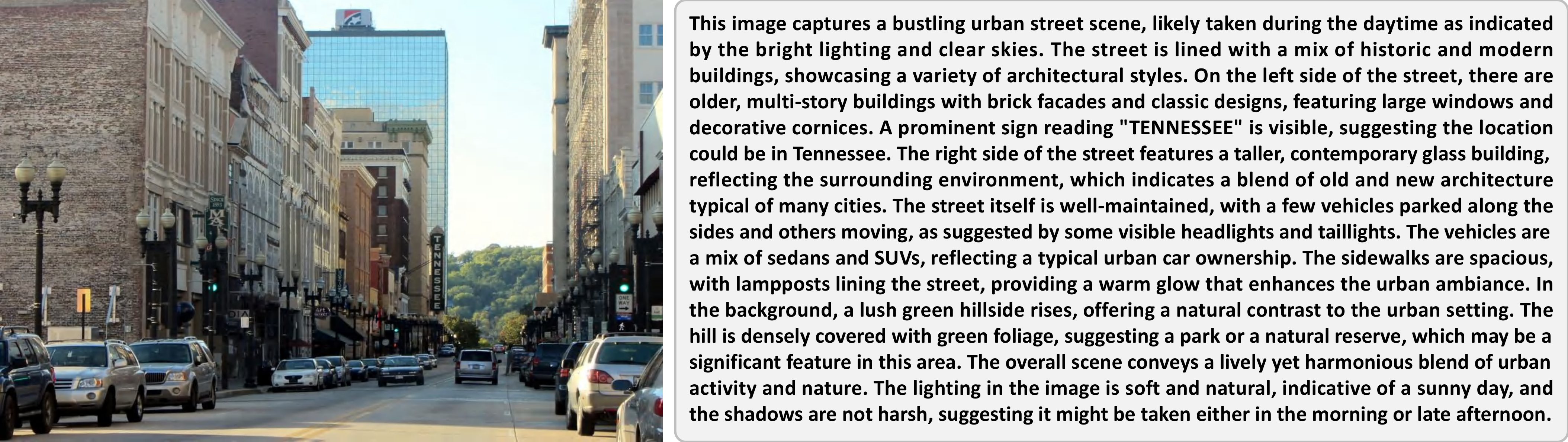} 
  \caption{Visualization of generated image detailed descriptions.}
  \label{fig:image_caption_visualization_example}
\end{figure}

\textbf{Examples of Object and Image Descriptions}
\label{par:2d_description_visable}
This part qualitatively shows the representative examples with the generated descriptions.
As shown in Fig.~\ref{fig:caption_visualization_example}, we present two types of object captions. 
The top row shows simple captions produced by Qwen2.5-VL for single-instance object categories, where spatial ordering is unnecessary. 
The bottom row includes captions augmented with spatial order information to distinguish multiple instances of the same category. 
Additionally, Fig.~\ref{fig:image_caption_visualization_example} demonstrates Qwen2.5-VL’s ability to generate detailed global descriptions of entire images used in the following.


\subsubsection{Generating Diverse QA Pairs via Pseudo-3D Scene Graphs}
\label{subsubsec:qa_generation_2d}

After constructing scene graphs and generating hierarchical object descriptions, we can leverage this information to generate diverse QA pairs from pseudo-3D scene graphs to support SFT training for improved single-step spatial understanding.

\textbf{Template, Choice and Fact QA Generation.}
\label{par:template_qa_fact_2d}
We first adopt a template-based method to generate structured preliminary QA pairs, multiple-choice questions, and factual statements. 
The templates are derived from scene graph information (\eg, object attributes, positions) and refined hierarchical object descriptions.
When designing QA templates using pseudo-3D scene graphs from 2D images, we explicitly account for the spatial ambiguity inherent in single-view pseudo-3D representations (\eg, inaccuracies in monocular depth, lack of object orientation, absence of 3D oriented bounding boxes). 
Consequently, \highlight{our QA templates from 2D data source focus mainly on qualitative spatial understanding and reasoning}, while incorporating quantitative cues only when they can be inferred reliably from 2D or pseudo-3D signals.
The spatial concepts covered in the QA templates fall into the following categories:




\begin{enumerate}
\item \textbf{Relative position relations}: capture spatial layouts (left/right, above/below, front/behind).
\item \textbf{Relative size comparisons}: describe object attributes (\eg, bigger/smaller, taller/shorter, wider/thinner) often inferred from image-plane projections.
\item \textbf{Quantitative information from 2D or pseudo-3D label}: include spatial reasoning based on estimated depth maps, 2D object coordinates, and coarse monocular depth approximations.
\end{enumerate}

Accordingly, we design diverse QA templates in these types for spatial understanding:



\begin{enumerate}
    \item Templates assess spatial and size relations:
    \begin{itemize}
        \item Position relation: \texttt{``Is [A] to the left of [B]?''}
        \item Size comparison: \texttt{``Which object is larger, [A] or [B]?''}
    \end{itemize}
    
    \item Templates query 2D point of a uniquely identified object~`[A]' (defined in Appx.~\ref{suppsubsubsec: object_description_generation}): 
    \begin{itemize}
        \item \texttt{``Where is [A] located? Provide its 2D coordinates.''} (\eg, ``\textit{Where is the red apple at left, which is the second apple from left to right, located? Provide its 2D coordinates.}'')
    \end{itemize}

    \item Templates query attributes at a specific 2D point `[X]' (formatted as `(x, y)'):
    \begin{itemize}
        \item Depth retrieval: \texttt{``What is the depth at point [X]?''} (\eg, ``\textit{What is the depth at (0.528, 0.317)}?'')
        \item Object identification: \texttt{``Which object is at point [X]?''} (\eg, \textit{``Which object is at (0.753, 0.839)?''})
    \end{itemize}
\end{enumerate}


We further design fact templates to generate declarative statements, forming a structured basis for prompting Reasoning LLM to produce richer and more natural QA pairs. 
Example templates include:
\begin{enumerate}
    \item Approximate depth: \texttt{``Point [A] and the camera are [X] apart.''} (based on depth estimation).
    \item Precise 2D object location: \texttt{``[A] is located at point [X].''}, \texttt{``[A] is to the right of [B].''}
\end{enumerate}


\textbf{Reasoning QA Generation.}
\label{supppar:qwq_diversification_2d}
\highlight{To generate more natural, complex, and diverse QA pairs beyond templated formats, we leverage a powerful reasoning LLM, QwQ-32B~\cite{qwq32b}}. 
It takes the factual statements, initial QA pairs, and multiple-choice questions (if available) as input, along with global image captions and precise object descriptions. 
QwQ-32B then produces more challenging and conversational spatial reasoning QA. 
The prompt design is shown in Listing~\ref{lst:qwq_prompt_2d}.


\begin{lstlisting}[basicstyle=\ttfamily\footnotesize, backgroundcolor=\color{myblue!50}, caption={Prompt for QwQ-32B QA Diversification and reasoning QA Generation.}, captionpos=t, breaklines=true, label={lst:qwq_prompt_2d}, escapeinside={(*@}{@*)}]
"""
You are a helpful assistant tasked with generating spatial reasoning-based questions and answers from provided descriptions of scenes.

Rules:
1. **We have three types of input information**:
- **[Scene]**: A general description of the entire image, which provides context for the objects and their surroundings.  
    **Example:** 
    [Scene]: The image shows a tranquil lakeside with a small wooden dock on the right and calm, reflective water in the center. The sky is overcast.
    
- **[Objects]**: A list containing one or more object labels separated by "|".  
    **Example:** 
    [Objects]: teal glossy water at lower center | green bamboo dock at lower right.
    
- **[Objects Description]**: Provides spatial or comparative details between those objects.  
    **Example:** 
    [Objects Description]: teal glossy water at lower center is taller than green bamboo dock at lower right.

2. **When crafting a Question**:
- **Always use the provided [Scene] description as context** to ensure the question aligns with the overall image.  
- **Mention all object labels from [Objects]** in the question.  
- **Do not modify or paraphrase the object labels**; they must appear **exactly** as given in `[Objects]'.  
- **Do not assume or invent additional scene details** beyond what is provided in `[Scene]'.  
- **Do not reveal the specific details in [Objects Description]** (like which object is taller, shorter, wider, etc.).
- Always generate questions related to the description using the object labels from [Objects].  
- Each object label in `[Objects]' **must appear exactly once** in the Question.  
- The question should read from **an observer's perspective**.
- The description should always be used to answer and not leak into the question. 

3. **When crafting an Answer**:
- **Mention at least one object label from [Objects]** in the answer.  
- **Use the `[Objects Description]' to provide a correct answer**.  
- **Ensure the answer is concise, factual, and directly related to the provided `[Scene]' and `[Objects]'**.  
- **You may restate or summarize the relevant details from `[Objects Description]', but do not introduce new assumptions**.

Here's several examples:

[Scene]: The image depicts a modern living room with a large window allowing warm sunlight to enter. The room has a wooden floor, a patterned rug in the center, and a coffee table with a few magazines neatly stacked on it. A yellow leather sofa is positioned centrally, facing the television mounted on the opposite wall. To the left of the sofa, a black metal chair with a cushioned seat is placed beside a tall bookshelf filled with an assortment of books and decorative items. The furniture arrangement leaves an open pathway between the sofa and the chair.
[Objects]: yellow leather sofa at lower center, black metal chair on the left. 
[Objects Description]: The path between yellow leather sofa at lower center and black metal chair on the left is 1.5 meters.
"Question": You are a cleaning robot that is 1 meter wide. Now you are standing in a living room and see the image; you want to move from here to the door that leads to the backyard. Do you think you can go through the path between the yellow leather sofa at lower center and the black metal chair on the left? 
"Answer": The path between the yellow leather sofa at lower center and the black metal chair on the left is 1.5 meters, so yes, the robot can go through the path between the yellow leather sofa at lower center and the black metal chair on the left since it is wider than the robot's width.

[Scene]: The image showcases a modern kitchen with a wooden countertop that extends across the space, separating the cooking area from the dining area. On the left side of the countertop, a fruit bowl holds a variety of fresh produce. A red fresh apple is placed on the left side of the bowl, while a bright fresh orange sits neatly on the right side. Behind the fruit bowl, a glass pitcher filled with orange juice and a stack of white ceramic plates are visible. Natural light streams in from a large window above the sink, reflecting off the stainless steel appliances and giving the space a bright, clean feel.
[Objects]: red fresh apple on the left, fresh orange on the right. 
[Objects Description]: red fresh apple on the left is positioned on the left side of fresh orange on the right.
"Question": You see two fruits, a red fresh apple on the left and a fresh orange on the right. Which one is more on the left side? 
"Answer": The red fresh apple on the left is more on the left.

Now its your turn!
"""
\end{lstlisting}

\textbf{Training data visualization.}
For specific examples of training data generated from 2D web images and their visualizations, please refer to Appx.~\ref{suppsec: more demonstrations}, which contains detailed sample presentations.

\subsection{3D Embodied Video}
\label{suppsubsec: 3D Embodied Video}

\subsubsection{Why Use CA-1M and How to Pre-process It}
\label{subsubsec:ca1m_rationale_new}

\textbf{Rationale for Selecting CA-1M as the 3D Data Source.}
To enable fine-grained spatial reasoning in indoor environments, we adopt Apple’s open-source CA-1M\footnote{\href{https://github.com/apple/ml-cubifyanything}{https://github.com/apple/ml-cubifyanything}}~\cite{lazarow2024cubify} dataset as our primary 3D data source.
CA-1M aligns closely with our objectives due to the following key attributes:

\begin{enumerate}
    \item Dense 2D/3D Annotations: CA-1M provides per-frame 2D/3D oriented bounding boxes, enabling spatial localization (\eg, 3D spatial occupancy) and accurate interaction modeling.

    \item Comprehensive Camera and Depth Data: The inclusion of camera intrinsics, extrinsics, and depth maps supports accurate 3D reconstruction and geometric reasoning.

    \item Large-Scale Coverage: Its extensive volume enables training/evaluation of VLMs at scale.
\end{enumerate}

\highlight{These features offer a strong foundation for constructing 3D scene graphs and generating spatial reasoning data involving 3D geometry, object interactions, and egocentric understanding.}




\textbf{Comparative Analysis with Alternative 3D Datasets.}
While several high-quality 3D datasets exist, they fall short of meeting the specific requirements of our project, motivating our selection of CA-1M:

\begin{enumerate}
    \item \textbf{ARKitScenes}~\cite{baruch2021arkitscenes}: As a predecessor to CA-1M, it provides only global 3D bounding boxes without per-frame 3D or 2D annotations. Projecting these 3D boxes to 2D yields oversized and contain irrelevant objects. Additionally, its annotations are less comprehensive, and image resolution is also much lower compared to CA-1M.
    
    \item \textbf{ScanNet V2}~\cite{dai2017scannet}: Lacks 3D bounding boxes, making object orientation estimation infeasible. Although EmbodiedScan introduces per-frame 3D bounding boxes, it still lacks 2D annotations, and projected 2D boxes remain imprecise and contain irrelevant objects.
    
    \item \textbf{3RScan}~\cite{wald2020beyond}: Suffers from low image quality, hindering spatial information extraction and limiting its usability.
\end{enumerate}



In summary, despite some limitations (See Appx.~\ref{suppsubsubsec: ca1m_challenges}), CA-1M provides large-scale egocentric video, per-frame 2D/3D annotations, depth, and camera parameters, \highlight{making CA-1M the most suitable choice for generating rich and complex 3D spatial data from an embodied perspective}.



\begin{figure}[h!]
  \centering
  \includegraphics[width=\linewidth]{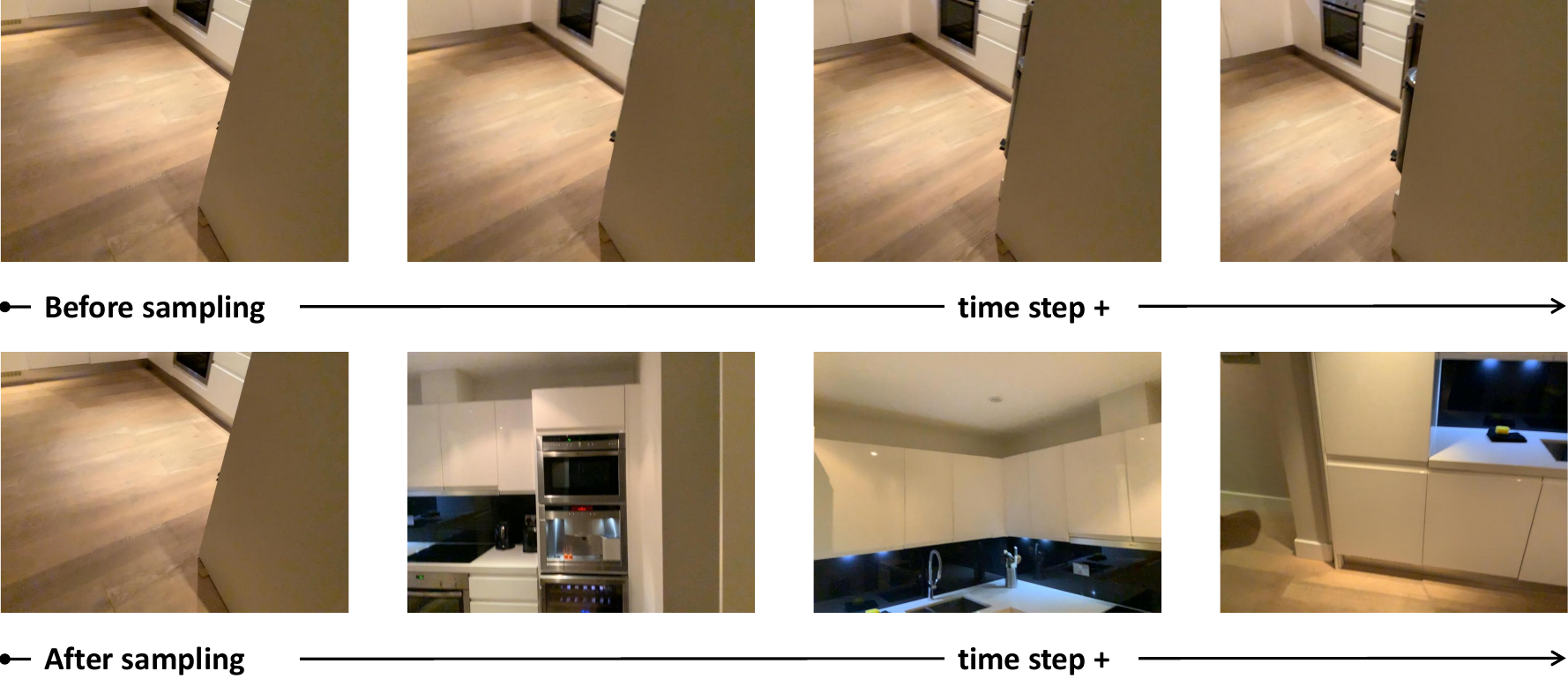}
  \caption{Comparison of frame sequences before and after sampling. Top: Four consecutive frames from the original video. Bottom: Corresponding frames sampled every 20 frames, exhibiting increased temporal variation between adjacent frames.}
  \label{fig:3d_sampling}
\end{figure}

\textbf{Pre-processing.}
Video datasets capturing continuous activities, such as CA-1M, exhibit high temporal redundancy, as consecutive frames often contain near-identical visual content with minor variations. 
Processing all frames is computationally intensive and yields redundant samples with limited informational gain for model training.
To mitigate this, we adopt a frame sampling strategy, selecting one frame every 20 frames. This reduces redundancy while preserving meaningful scene and viewpoint transitions.
The resulting subset maintains scene diversity and supports efficient downstream processing, including 3D scene graph construction and question-answer generation.
In Fig.~\ref{fig:3d_sampling}, the top row shows four consecutive frames before sampling, revealing minimal variation; the bottom row, sampled at 20-frame intervals, demonstrates significantly greater scene variation.



\subsubsection{Inherent Challenges and Limitations in CA-1M}
\label{suppsubsubsec: ca1m_challenges}

Although the CA-1M dataset is chosen for its large scale, egocentric perspective, and rich annotations, it presents inherent limitations that undermine scene graph construction if unaddressed.

\begin{figure}[h!]
  \centering
    \includegraphics[width=\textwidth]{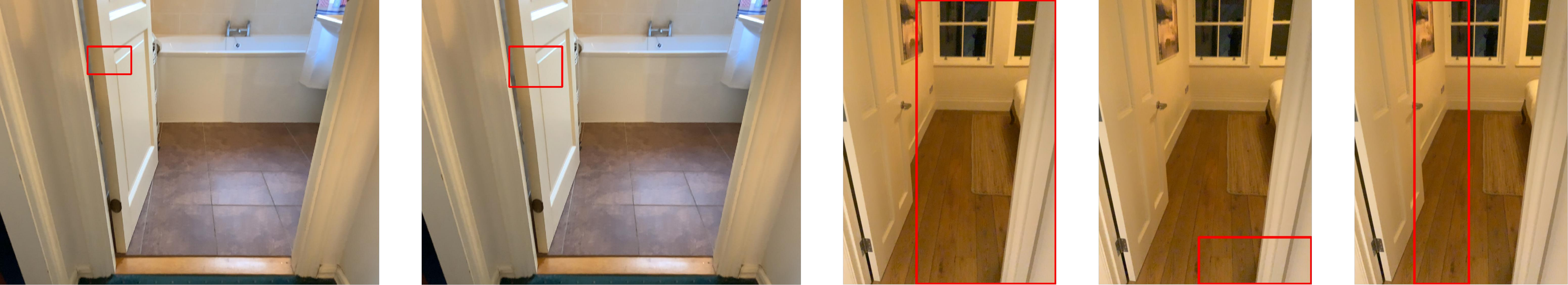}
    \caption{Example of some meaningless annotations in CA-1M.}
    \label{suppfig: ca1m_meaningless_example_c}

\end{figure}

\textbf{Ambiguous or Meaningless Object Annotations.}
A major issue is the prevalence of ambiguous or semantically insignificant object annotations. 
Many instances are difficult to interpret, even for humans, due to unclear boundaries or a lack of identifiable semantics. 
For example, some bounding boxes enclose small, indiscernible regions such as a patch of wall or vague background elements (see Fig.~\ref{suppfig: ca1m_meaningless_example_c}). 
Incorporating such annotations into model training introduces noise, hampers spatial understanding, and may mislead the model's perception of object relationships.




\textbf{Widespread Absence of Semantic Labels.}
Another major limitation of CA-1M is the lack of semantic labels for most annotated objects. 
Unlike datasets (\eg, ARKitScenes and ScanNet V2), which provide object category annotations, CA-1M includes only a few structural categories (\eg, floors, walls, doors), leaving the majority of object instances unlabeled.
Although bounding boxes indicate the presence of objects (\ie, labeled as ``object''), their categories (\eg, ``chair'', ``table'') remain unknown. 
This absence of semantic information hinders spatial understanding, making it impossible to generate category-dependent queries such as \textit{``Is the red chair to the left of the table?''}



\subsubsection{Addressing Limitations: Object Annotation and Bounding Box Filtering}
\label{subsubsec:ca1m_solutions}

To address the limitations of the CA-1M dataset, particularly the lack of semantic labels and the prevalence of noisy bounding boxes (See Appx.~\ref{suppsubsubsec: ca1m_challenges} above), we develop a dedicated annotation and filtering pipeline. 
This enhances the dataset’s usability for downstream tasks such as 3D scene graph construction and spatial reasoning.
We detail the two-stage pipeline below.

\textbf{Stage 1: Initial Annotation and 2D Bounding Box Prediction.}
We employ a multi-model combination strategy to annotate unlabelled objects and refine bounding boxes in CA-1M video frames.
Specifically, we integrated GroundingDINO, RAM, and Florence-2 to perform object semantic labeling and 2D bounding box prediction.
\highlight{Our approach yields semantically meaningful and visually coherent object bounding boxes compared to the often difficult-to-discern or ambiguous bounding boxes from the original CA-1M annotations.}
To maximize recall, we intentionally lower the confidence thresholds of GroundingDINO and RAM, ensuring the inclusion of low-confidence but potentially relevant objects. 
These candidates are retained for further validation in subsequent matching and filtering stages.

\textbf{Stage 2: Bidirectional 2D Bounding Box Matching.}
To associate model-predicted 2D bounding boxes (with semantic labels) with unlabeled boxes in CA-1M, \highlight{we propose a bidirectional matching and refinement strategy.
This not only enables semantic annotation of meaningful objects but also filters out noisy or ambiguous CA-1M bounding boxes}, thus enhancing the utility of its 3D annotations (\eg, 3D oriented bounding boxes).
The strategy consists of two steps:

\begin{enumerate}
    \item \textbf{Matching CA-1M Bounding Boxes to Model Predictions.} We first match original CA-1M 2D {bounding} boxes to model-predicted {bounding} boxes based on the IoU metric. Due to the sparsity, occlusions, and fragmentation of annotation, multiple CA-1M boxes may correspond to a single prediction, resulting in many-to-one matches.
    
    \item \textbf{Refining Model Predictions via One-to-One Mapping.} To resolve many-to-one matches, we retain only predicted boxes that match at least one CA-1M {bounding} box. For each, we assign the CA-1M {bounding} box with the highest IoU as its unique match. This enforces a one-to-one correspondence, eliminating redundant or weakly aligned CA-1M boxes. The result is a refined set of original {bounding} boxes with strong semantic alignment.
\end{enumerate}

\begin{figure}[h!]
  \centering
    \includegraphics[width=\textwidth]{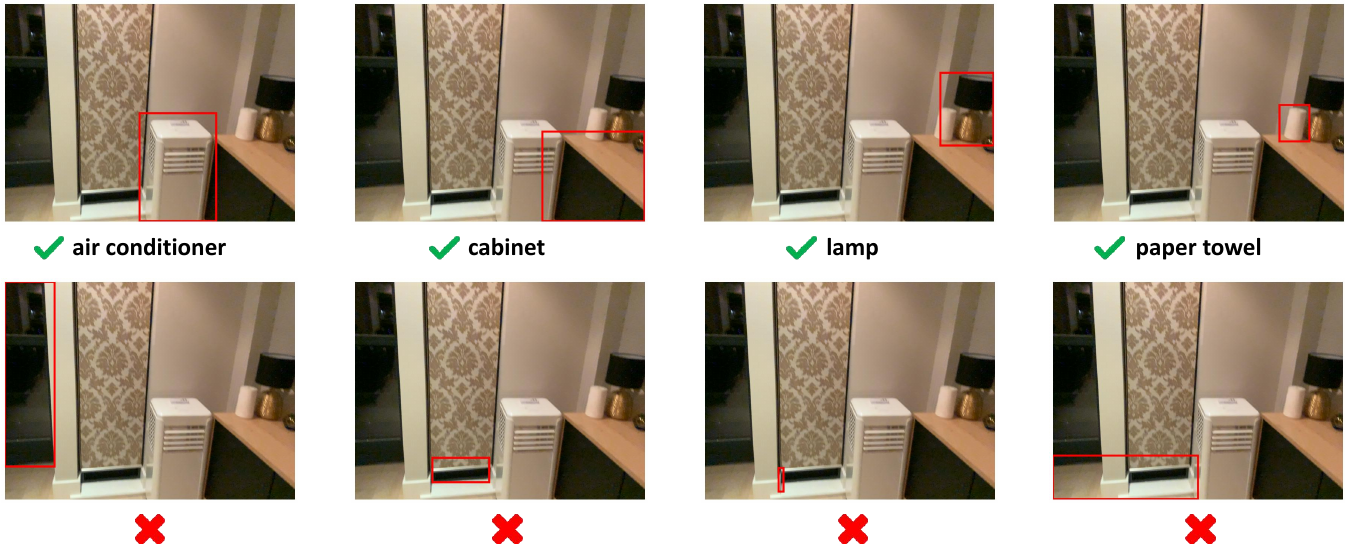}

    \caption{Visual comparison of 2D bounding box matching. The top row shows matched CA-1M {bounding} boxes with object labels; the bottom row displays unmatched or ambiguous cases.}
  \label{fig:bbox_matching_comparison}
\end{figure}

\textbf{Visualizing Matching Results.}
Fig.~\ref{fig:bbox_matching_comparison} shows the effectiveness of our {bounding} box matching procedure. 
The first row shows successfully matched CA-1M {bounding} boxes annotated with RAM-predicted object labels, while the second row highlights unmatched {bounding} boxes, typically corresponding to ambiguously annotated objects. 
This demonstrates that our method reliably filters out uncertain annotations and assigns semantic labels to clearly identifiable objects.
For subsequent scene graph generation, we adopt the model-predicted {bounding} boxes instead of the original CA-1M ones, as they better align with the visible object extents, facilitating more accurate instance mask extraction via models like SAM2.


\subsubsection{3D Object Description Generation and Scene Graph Construction.}
%
%
The process of 3D scene graphs construction largely follows the 2D scene graph pipeline in Appx.~\ref{supsubsubsec: 3D Scene Graph Construction} and the description generation in Appx.~\ref{suppsubsubsec: object_description_generation}. 
The resulting graphs are structurally similar to those from OpenImages in Fig.~\ref{suppfig: scene_graph_visualization}. 
The key distinction is CA-1M’s focus on indoor environments and its provision of high-precision geometric data (\eg, ground-truth depth, camera intrinsics/extrinsics, 3D oriented bounding boxes). 
%
%
\highlight{The enhanced 3D scene graph outperforms 2D-based counterparts in object localization and spatial relation accuracy, enabling structurally rich and quantitatively grounded 3D QA data across $28$ spatial relation types.}.
%
%


\subsubsection{Free Space QA Generation for Object Placement }
\label{subsubsec:free_space_identification}

A key challenge in spatial referring is identifying unoccupied regions suitable for object placement. 
We propose a multi-step pipeline to address this.

\begin{figure}[h!]
  \centering
  \includegraphics[width=\linewidth]{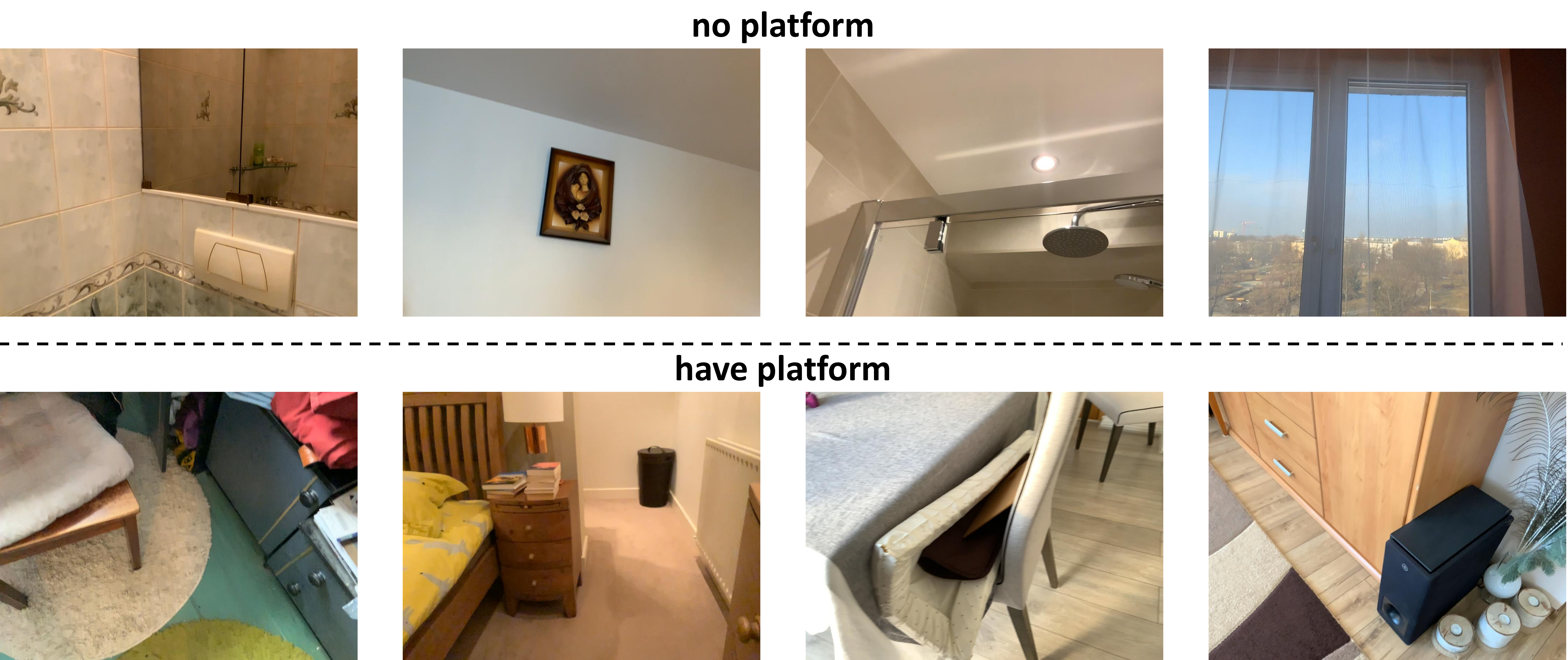} 
  \caption{Visualization of platform filtering results. Row 1: Examples of images filtered out (lacking platforms). Row 2: Examples of images retained (containing platforms).}
  \label{fig:platform_filtering_results}
\end{figure}

\textbf{Step 1: Detecting Viable Platforms via Qwen2.5-VL.}
We first filter out scenes lacking plausible placement surfaces (\eg, {only }walls or ceilings). 
To this end, we employ the Qwen2.5-VL-7B, initialized with a system prompt (Listing~\ref{lst:qwen_platform_system_prompt}) that specifies its role and procedure. 
For each image, a user prompt (Listing~\ref{lst:qwen_platform_user_prompt}) directs the model to detect candidate {images with} surfaces such as tables, floors, or shelves.
Fig.~\ref{fig:platform_filtering_results} shows the filtering results. 
The first row shows scenes without platforms, while the second row shows scenes with platforms. 
This filtering enables subsequent computationally intensive analysis to focus on semantically relevant scenes for placement queries.
%




\begin{lstlisting}[basicstyle=\ttfamily\footnotesize, backgroundcolor=\color{myblue!50}, caption={System Prompt for Qwen2.5-VL to identify images with suitable platforms.}, captionpos=t, breaklines=true, label={lst:qwen_platform_system_prompt}]
image_have_platform_system_text_prompt = """
You are an expert visual scene understanding assistant.

Your task is to analyze an image and determine whether it contains **any obvious flat horizontal surfaces** where physical objects can be placed. These include **floors, tabletops, bed surfaces, or other flat and stable areas**.

IMPORTANT:
- If you can see any part of the **floor**, **tabletop**, **bed**, or **similar flat surfaces**, you MUST assume it can support physical objects (\eg, books, boxes, pillows).
- Do NOT consider whether the surface is cluttered, partially visible, or obstructed. If the platform exists and is horizontal, assume it can hold objects.
- Your answer must be based strictly on visible surfaces.

You must provide a short reasoning based on visual evidence in the image, followed by a final conclusion.

Your response MUST strictly follow this format:
"[Your analysis]. | Yes/No"

Examples of valid responses:

Example 1:  
The image shows a wooden floor that is flat and unobstructed. And it could potentially support physical objects. | Yes

Example 2:  
There is a bed clearly visible in the scene with a flat top surface where items like pillows or books can be placed. | Yes

Example 3:  
A rectangular table is visible in the center of the image, providing a flat surface suitable for placing objects. | Yes

Example 4:  
The image contains mostly a wall with a window and no visible floor, table, or other flat surfaces. | No

Do NOT provide any extra commentary or formatting outside this exact format.
"""
\end{lstlisting}

\begin{lstlisting}[basicstyle=\ttfamily\footnotesize, backgroundcolor=\color{myblue!50}, caption={User Prompt for Qwen2.5-VL to identify images with suitable platforms.}, captionpos=t, breaklines=true, label={lst:qwen_platform_user_prompt}]
image_have_platform_user_text_prompt = """
Please examine the image and determine whether it contains any **clear horizontal platforms** where physical objects can be placed. These platforms include: **floors, tabletops, bed surfaces, or other flat and stable horizontal areas**.

Important notes:
- As long as **any part** of a **floor**, **tabletop**, **bed surface**, or similar platform is visible in the image, you must assume it is capable of supporting physical objects (such as books, boxes, pillows, etc.).
- Do NOT consider whether the surface is messy, partially blocked, or whether there's enough space. If the platform exists and is horizontal, you must assume it can hold objects.
- Your answer must be based entirely on visible visual evidence in the image.

You must respond in the following exact format:
"[Your analysis]. | Yes/No"

Refer to the following examples to guide your response:

Example 1:  
The image shows a wooden floor that is flat and unobstructed. And it could potentially support physical objects. | Yes

Example 2:  
There is a bed clearly visible in the scene with a flat top surface where items like pillows or books can be placed. | Yes

Example 3:  
A rectangular table is visible in the center of the image, providing a flat surface suitable for placing objects. | Yes

Example 4:  
The image contains mostly a wall with a window and no visible floor, table, or other flat surfaces. | No

Please strictly follow the format and do not add any extra commentary.
"""
\end{lstlisting}

\begin{figure}[h!]
  \centering
  \begin{minipage}[b]{0.48\textwidth}
    \centering
    \includegraphics[width=\linewidth]{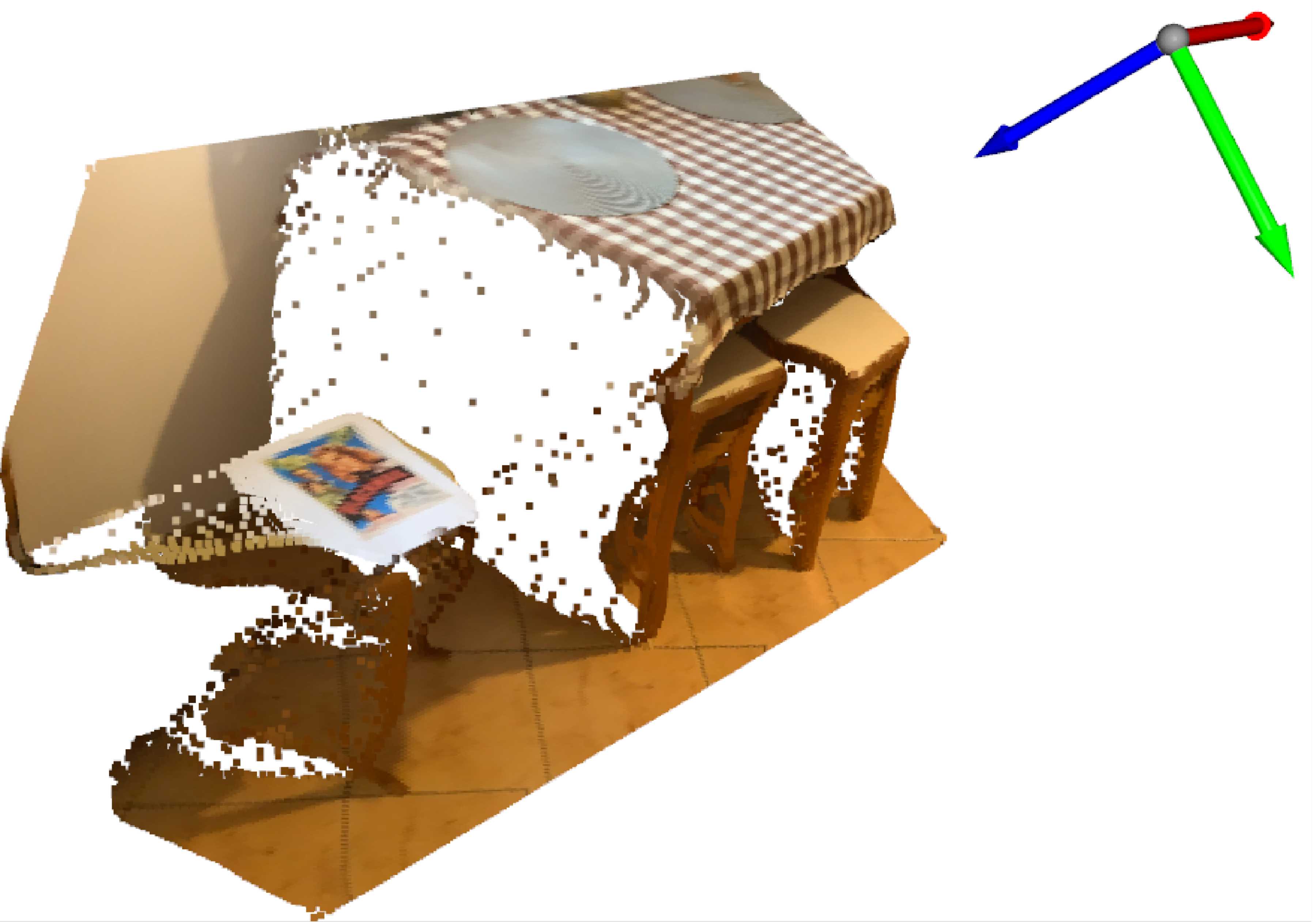}
    \caption*{(a) Before gravity alignment}
    \label{fig:gravity_before}
  \end{minipage}
  \hfill
  \begin{minipage}[b]{0.48\textwidth}
    \centering
    \includegraphics[width=\linewidth]{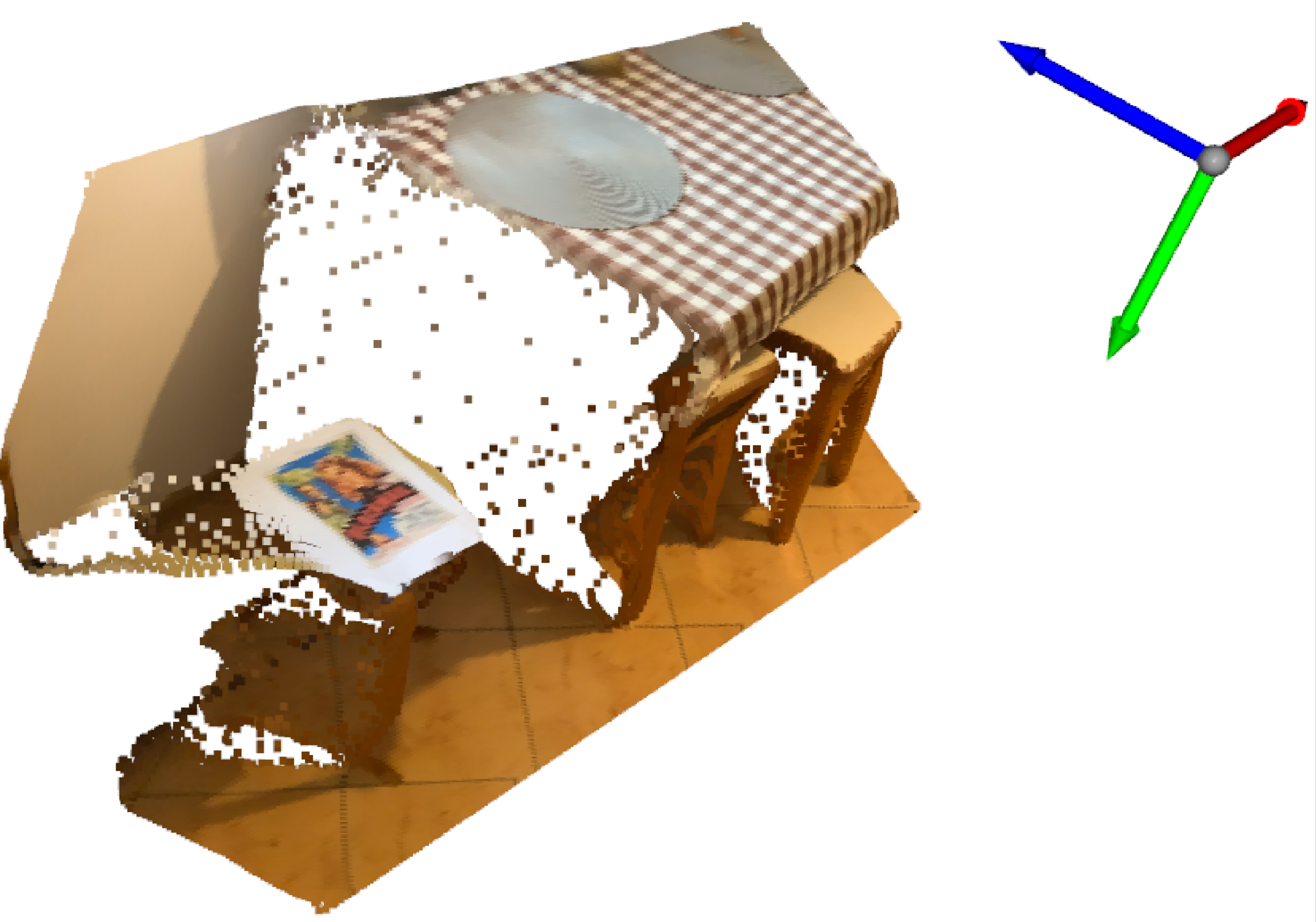}
    \caption*{(b) After gravity alignment}
    \label{fig:gravity_after}
  \end{minipage}
  \caption{Visualization of gravity alignment. (a) The scene before alignment. (b) The scene after alignment, with the Y-axis oriented along the gravity vector.}
  \label{fig:gravity_alignment_effect}
\end{figure}

\textbf{Step 2: Gravity Alignment and Top-Down Projection.}
To enable top-down (orthographic) scene projection, we apply gravity alignment matrices from the CA-1M to transform point clouds and object {3D bounding box} into a consistent frame where gravity uniformly points downward. 
This normalization allows for projection onto a plane orthogonal to the gravity vector, revealing object layouts and spatial relations more clearly. 
Fig.~\ref{fig:gravity_alignment_effect} shows the point cloud and coordinate axes before and after alignment.


\textbf{Step 3: Programmatic Platform Association.}
After gravity alignment, we associate objects with supporting platforms based on their spatial relationships.
For relationships such as \textbf{``front/behind/left/right''}, we identify the platform supporting a target object by checking the top surfaces of candidate platforms. 
A platform qualifies if its top surface is within $0.05$ meters of the object’s bottom surface and overlaps at least $70$\% of the object’s bottom area in top-down view.
For the \textbf{``below''} relationship, when an object may be suspended in mid-air, we identify the supporting platform as the one located beneath it, whose top surface is nearest to the object's bottom, and whose top-down intersection with the object exceeds 70\% of the object's area.
For the \textbf{``above''} relationship, the object's own top surface is treated as the reference platform, eliminating the need for additional platform search.
For the \textbf{``between''} relationship involving two objects, each object's supporting platform is determined independently using the same procedure as in the \textbf{``front/back/left/right''} cases. 
The space between them is considered valid only if both share the same supporting platform, ensuring spatial reasoning occurs within a unified physical context.

\textbf{Step 4: Identifying Other Objects on the Platform.}
\label{par:identifying_other_objects}
After identifying the target object and its supporting platform, we locate other relevant objects co-occurring on the same platform. 
For \textbf{``front/behind/left/right/between''}, candidate objects are selected based on the following criteria:

\begin{enumerate}
    \item Their bottom must not be significantly higher than the top of the target object.
    \item Their top must remain above the platform surface.
    \item Their XZ-plane footprint must intersect with that of the target object.
    \item Their volume must not exceed $4.236$ times that of the target object. 
\end{enumerate}



\begin{figure}[h!]
  \centering
  \begin{minipage}[b]{0.32\textwidth}
    \centering
    \includegraphics[width=\linewidth]{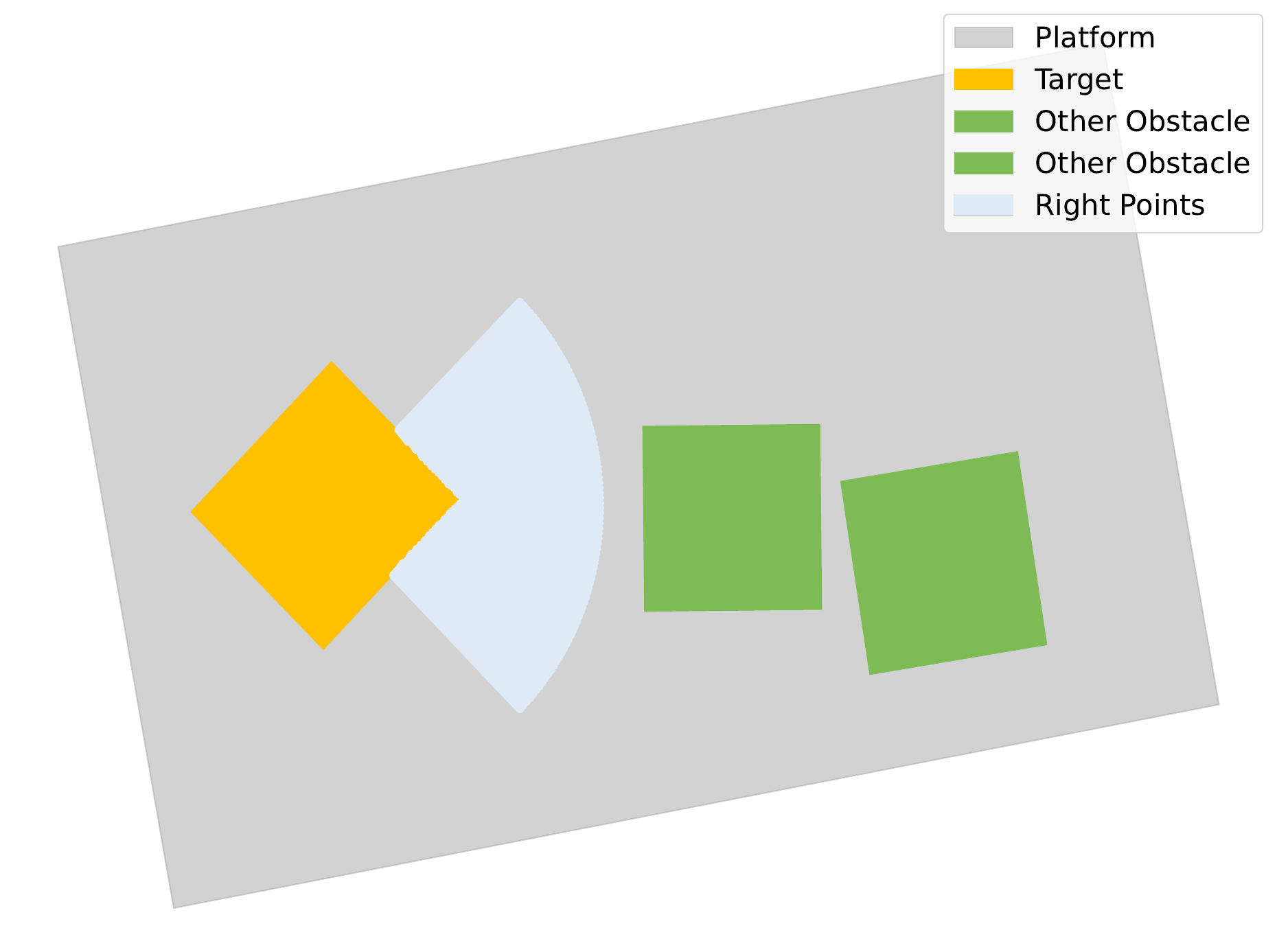}
    \caption*{(a) Top-view occupancy map with right-side search area}
    \label{fig:right_topview}
  \end{minipage}
  \hfill
  \begin{minipage}[b]{0.32\textwidth}
    \centering
    \includegraphics[width=\linewidth]{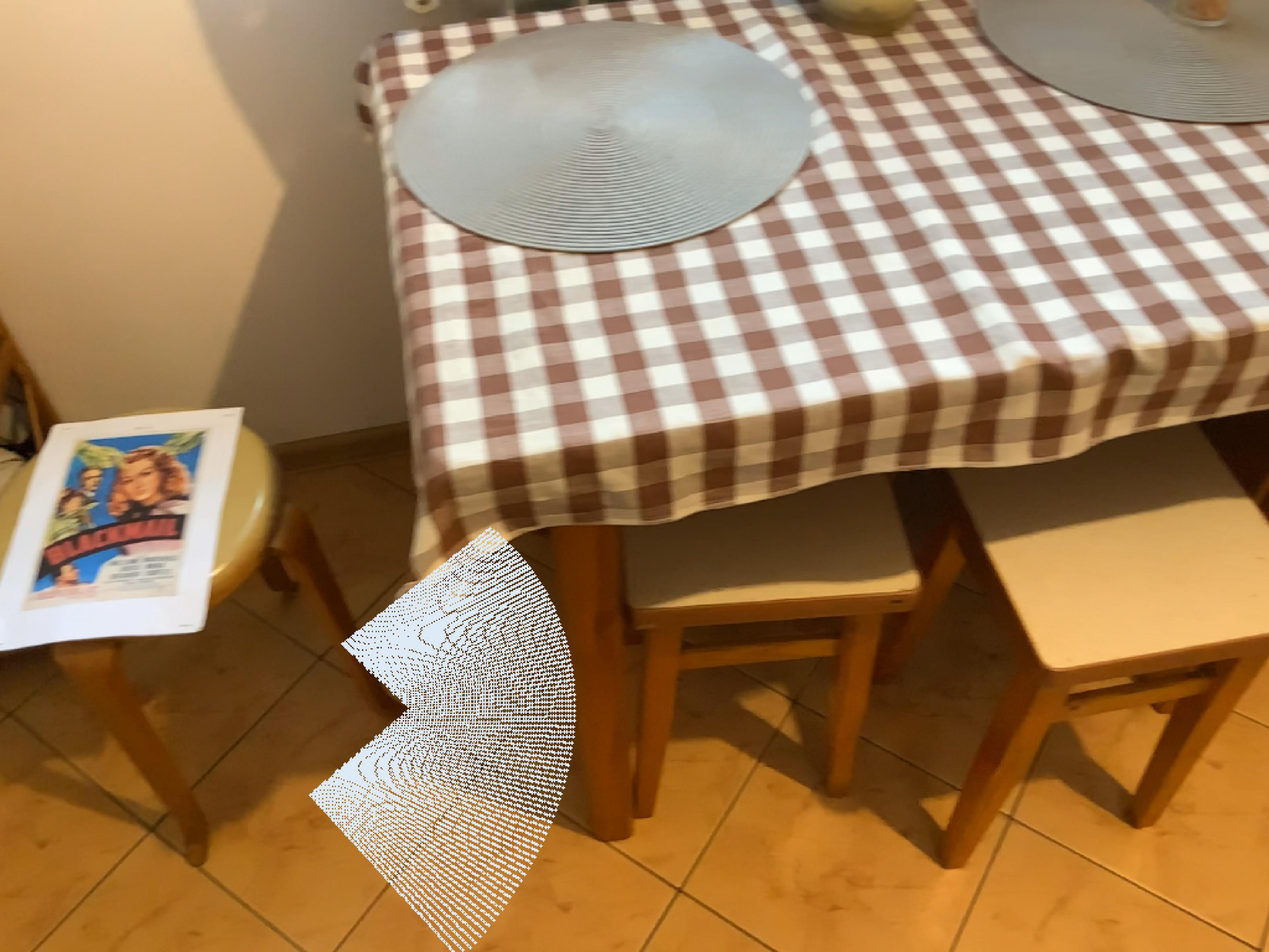}
    \caption*{(b) Sampled points projected onto the 2D image}
    \label{fig:right_projected_points}
  \end{minipage}
  \hfill
  \begin{minipage}[b]{0.32\textwidth}
    \centering
    \includegraphics[width=\linewidth]{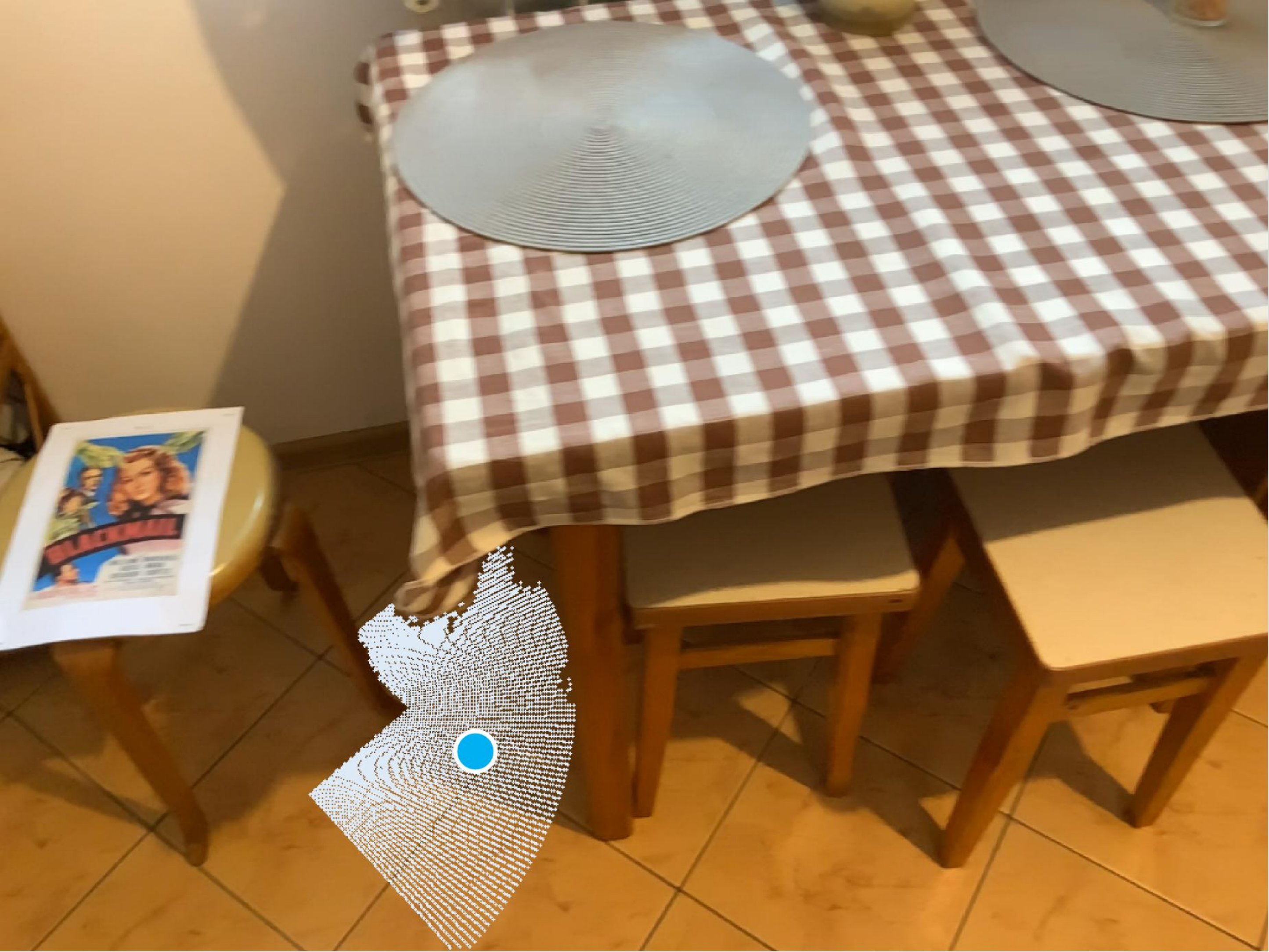}
    \caption*{(c) Visible points with the final placement center}
    \label{fig:right_visible_filtered}
  \end{minipage}
  \caption{
    Visualization of \textbf{right}-side free space identification.
    (a) Top-view occupancy map with the target's right-side search area.
    (b) Projection of sampled candidate points into the image plane.
    (c) Non-visible points are removed; the final placement center is marked with a blue circle.
  }
  \label{fig:right_free_space_pipeline}
\end{figure}

\begin{figure}[h!]
  \centering
  \begin{minipage}[b]{0.32\textwidth}
    \centering
    \includegraphics[width=\linewidth]{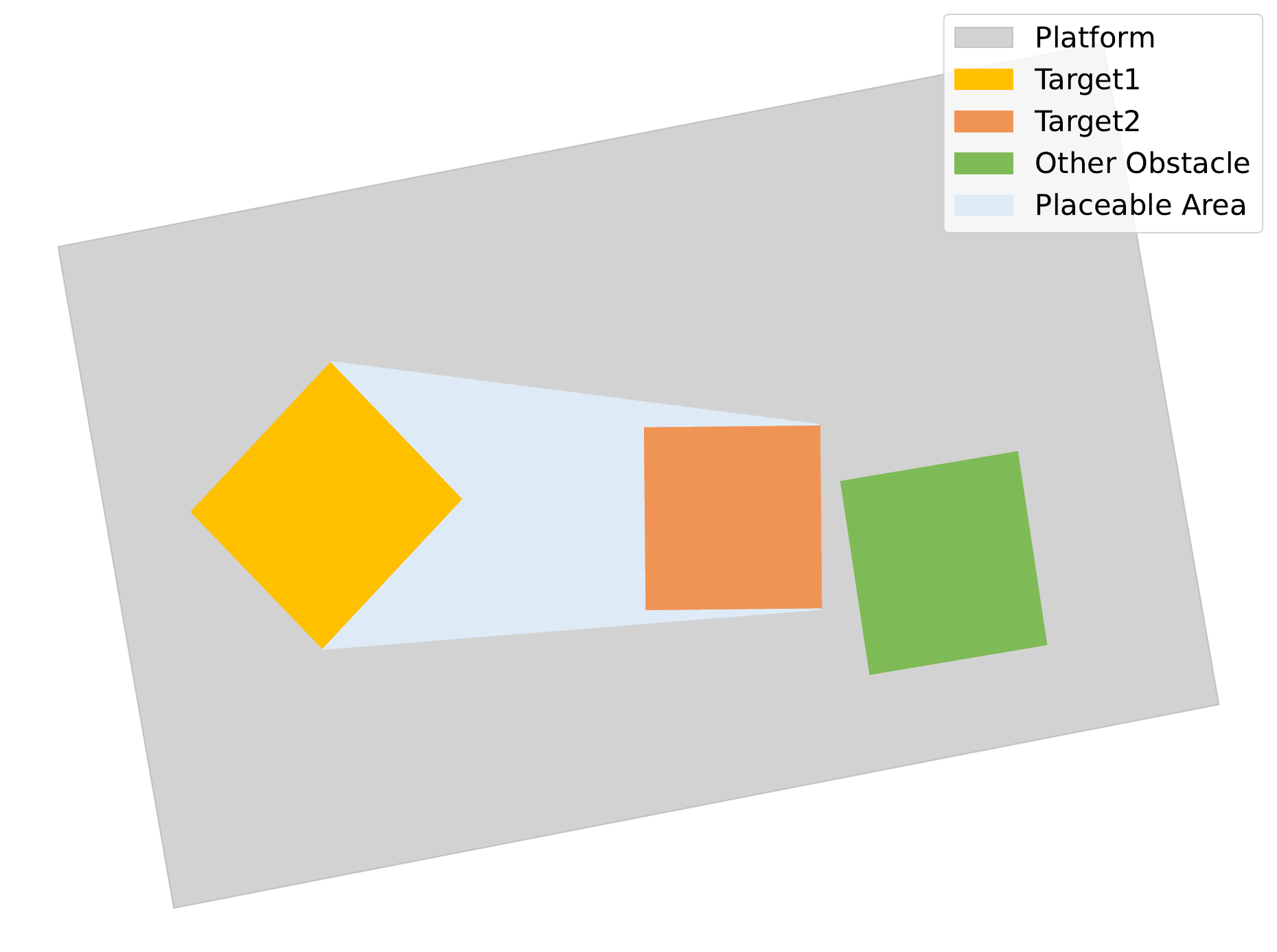}
    \caption*{(a) Top-view occupancy map with between search area}
    \label{fig:between_topview}
  \end{minipage}
  \hfill
  \begin{minipage}[b]{0.32\textwidth}
    \centering
    \includegraphics[width=\linewidth]{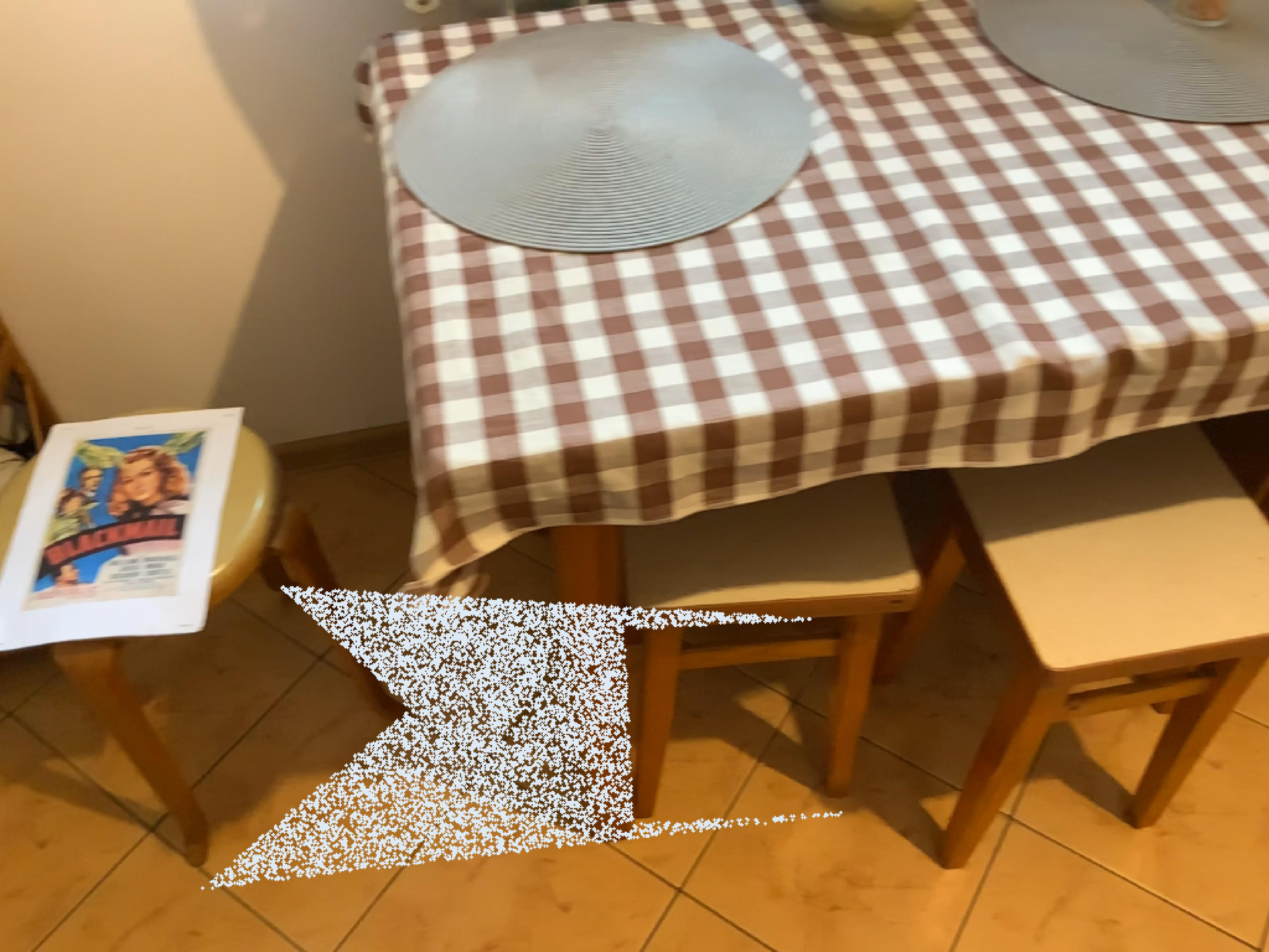}
    \caption*{(b) Sampled points projected into the 2D image}
    \label{fig:between_projected_points}
  \end{minipage}
  \hfill
  \begin{minipage}[b]{0.32\textwidth}
    \centering
    \includegraphics[width=\linewidth]{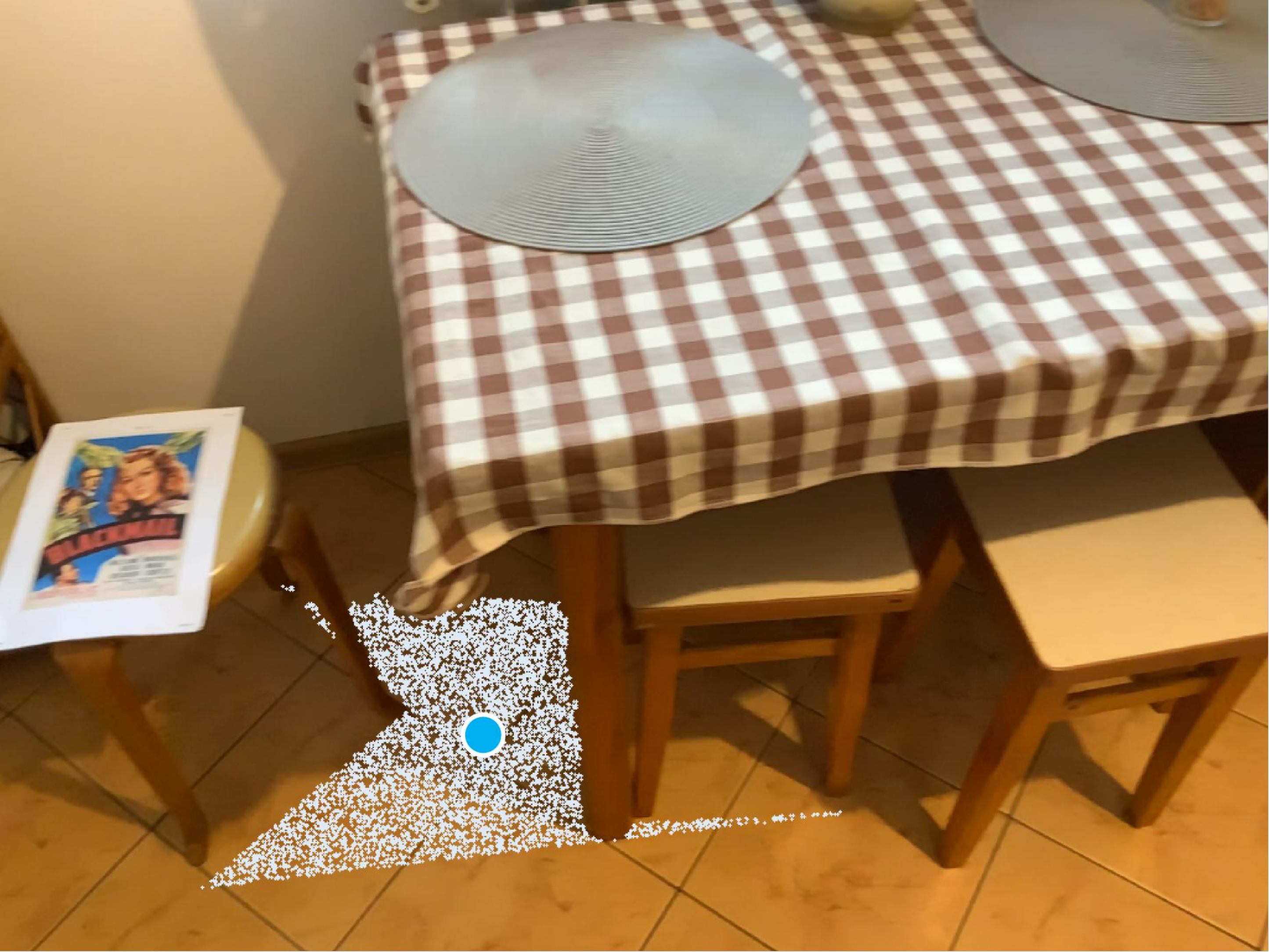}
    \caption*{(c) Visible points with the final placement center}
    \label{fig:between_visible_filtered}
  \end{minipage}
  \caption{
    Visualization of \textbf{between}-object free space identification.
    (a) Top-view occupancy map showing the area between two target objects on the same platform.
    (b) Projection of sampled points into the image plane.
    (c) Non-visible points are filtered; the final placement center is indicated.
  }
  \label{fig:between_free_space_pipeline}
\end{figure}

To mitigate the over-filtering of solid objects during visible point selection via depth matching, we introduce a volumetric constraint (thresholded at approximately $4.236$, \ie, $(1/0.618)^3$, the cube of the reciprocal of the golden ratio). 
This prevents large, potentially hollow objects—such as tables with substantial under-space—from being misclassified as fully occluding based solely on their bounding box volume, which would mislead placement reasoning on the primary platform.
Instead, the filter targets small to medium-sized objects, where the bounding box volume more accurately reflects actual occupancy. 
For such objects, even if hollow, the limited under-space is typically negligible for placement purposes.
This behavior is shown in the top-view occupancy maps in Fig.~\ref{fig:right_topview} and Fig.~\ref{fig:between_topview}, where a large table is excluded due to exceeding the volume threshold. 
Despite being present in the scene, its hollow geometry allows for usable space beneath, justifying its omission.




For the \textbf{``below''} relation, an object on a platform is considered below the target object (when the target might be suspended or on a higher tier) if:
\begin{enumerate}
    \item Its footprint (XZ-plane projection) intersects with that of the target;
    \item Its bottom is no higher than the top of the target;
    \item  Its top is not below the top surface of its supporting platform.
\end{enumerate}

For the \textbf{``above''} relation, an object is considered above the platform of target object if:
\begin{enumerate}
    \item Its bottom is within 20\,cm above the platform's top surface;
    \item Its top is not below the platform's top surface;
    \item Its footprint (projection on the XZ plane) overlaps with that of the platform.
\end{enumerate}



\begin{figure}[h!]
  \centering
  \begin{minipage}[b]{0.32\textwidth}
    \centering
    \includegraphics[width=\linewidth]{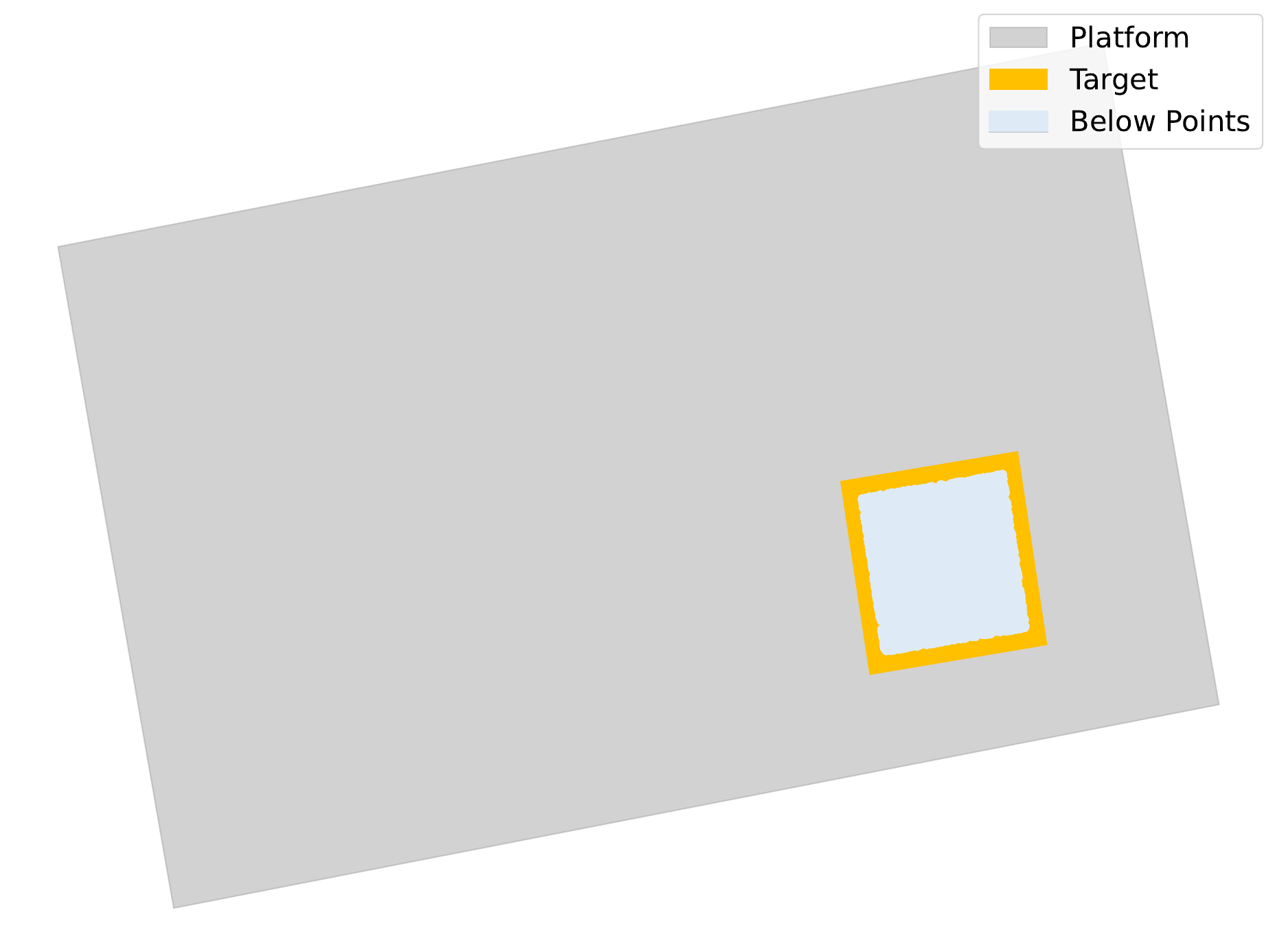}
    \caption*{(a) Top-view occupancy map of the object's bottom surface}
    \label{fig:below_topview}
  \end{minipage}
  \hfill
  \begin{minipage}[b]{0.32\textwidth}
    \centering
    \includegraphics[width=\linewidth]{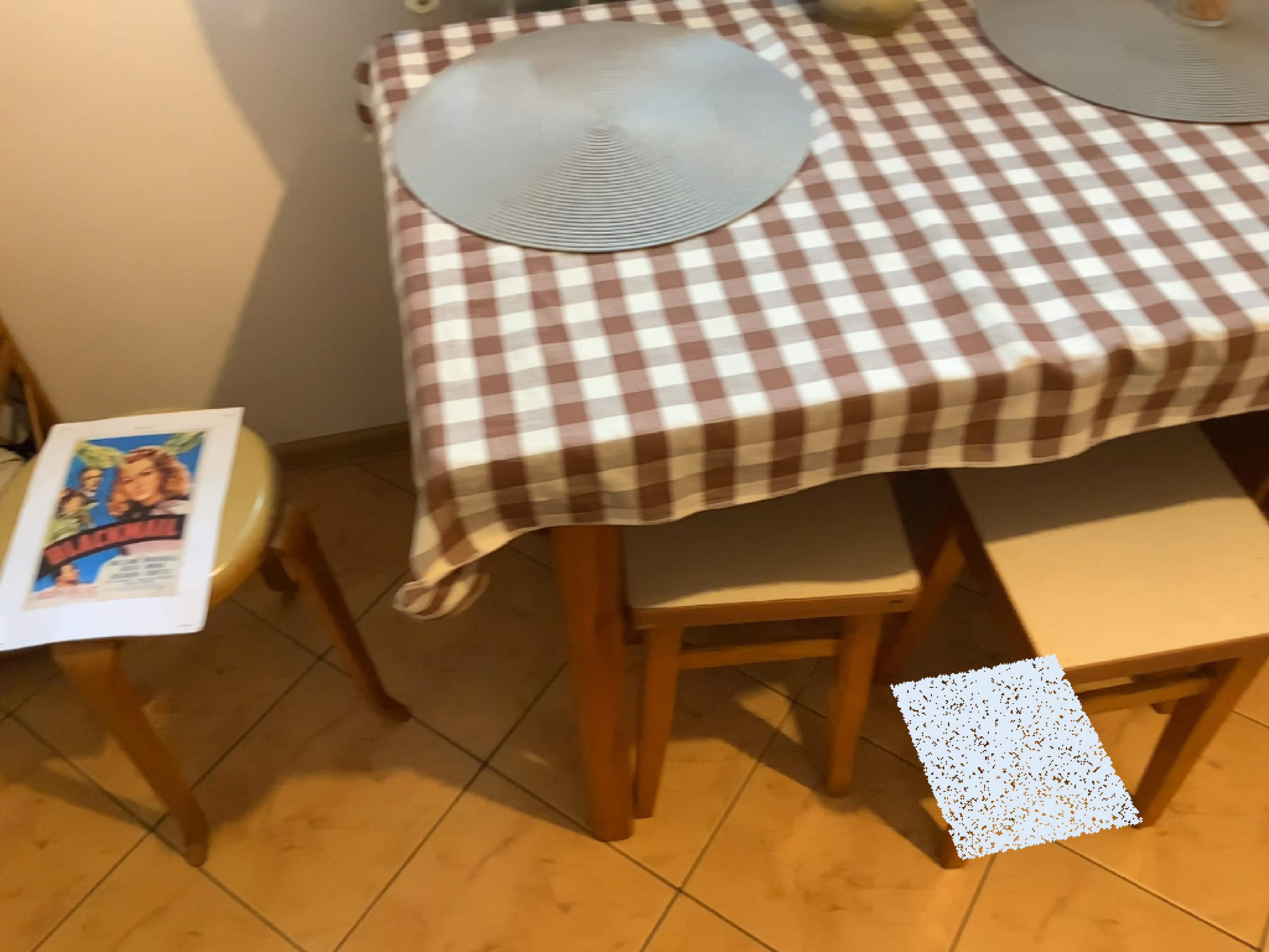}
    \caption*{(b) Sampled points projected into the 2D image}
    \label{fig:below_projected_points}
  \end{minipage}
  \hfill
  \begin{minipage}[b]{0.32\textwidth}
    \centering
    \includegraphics[width=\linewidth]{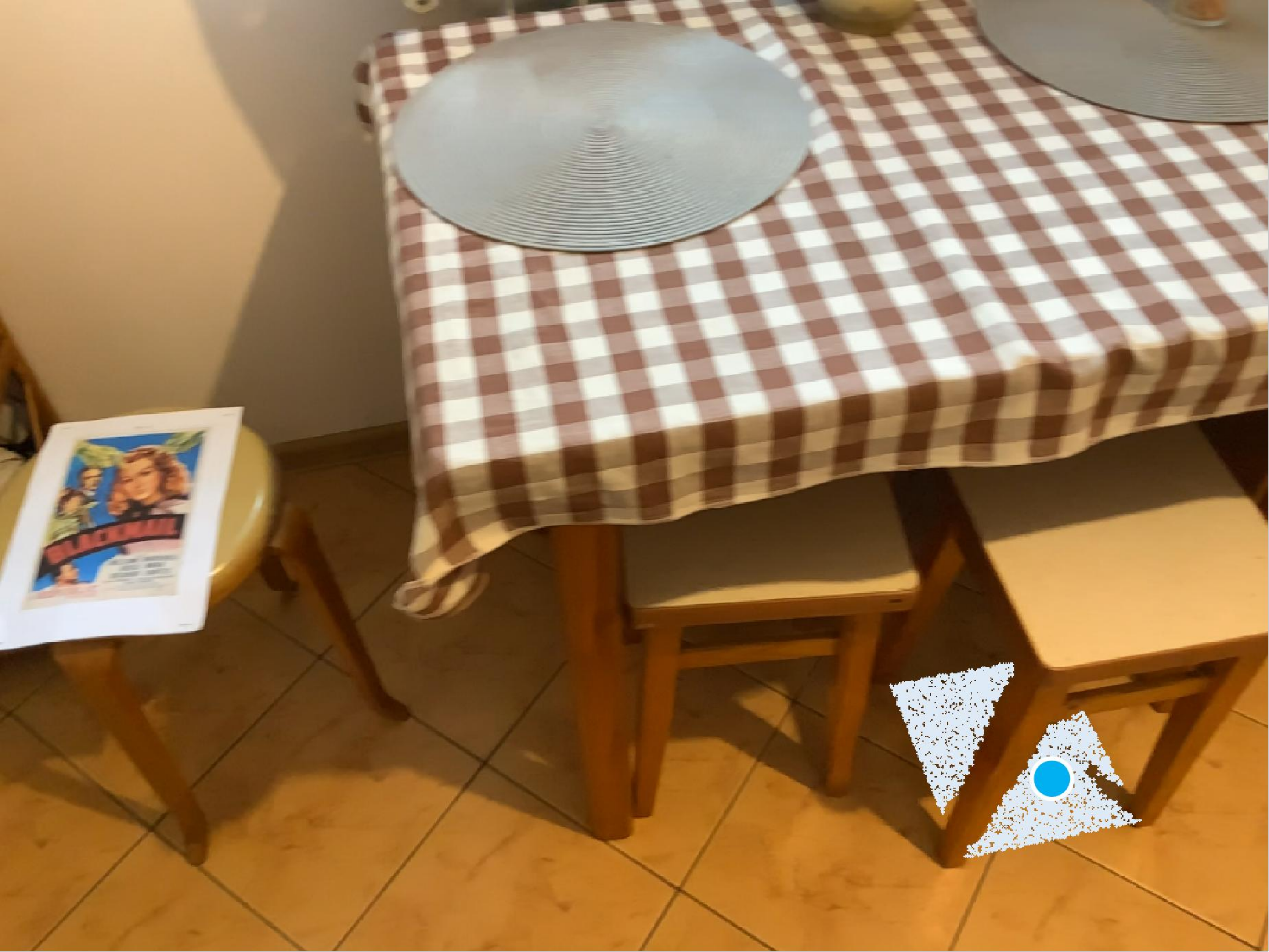}
    \caption*{(c) Visible points with the final placement center}
    \label{fig:below_visible_filtered}
  \end{minipage}
  \caption{
    Visualization of \textbf{bottom}-side free space identification.
    (a) Top-view occupancy map of the target's bottom surface on the platform.
    (b) Projection of candidate points into the image plane.
    (c) Non-visible points are filtered; the final placement center is indicated.
  }
  \label{fig:below_free_space_pipeline}
\end{figure}

\textbf{Step 5: Sampling Unoccupied Points in the Top-View Occupancy Map.}
After identifying the target object, its supporting platform, and adjacent objects, we determine the surrounding free space. This includes regions in front, behind, left, right, above, or below the target.
To locate free space in the horizontal directions (\textbf{``front/behind/left/right''}), we define a $90^\circ$ sector centered on the target and oriented in the respective direction. The sector radius is set to the maximum of either the diagonal of the object’s footprint or a fixed $20$ cm, ensuring adequate coverage.
For vertical directions (\textbf{``above/below''}), we project the object's top or bottom surface onto the supporting platform. To mitigate overestimation from coarse 3D bounding boxes, we shrink the projection to 80\% of its original size (centered), reducing overlap with nearby objects and better approximating usable space.
To identify free space \textbf{``between''} two target objects, we define the search region as the planar area enclosed by the projections of both objects onto their shared supporting surface. This region is evaluated for occupancy by other objects.
Across all spatial contexts (\textbf{``above/below/between''}), we enforce a minimum free area constraint: the unoccupied region must exceed $0.036m^2$ (half an A4 sheet) in a top-down view. This threshold filters out trivial cases and ensures the queried space can accommodate small objects such as a book or cup.
Free space is computed by analyzing object footprints in the gravity-aligned top-down view. 
In Fig.~\ref{fig:right_topview}, \ref{fig:below_topview}, \ref{fig:between_topview}, blue-shaded regions indicate the final sampling areas after accounting for bounding box scaling and occlusions. 
These visualizations highlight viable unoccupied regions for subsequent placement analysis.

\textbf{Step 6: Projection and Visibility Filtering.}
Given candidate points sampled in the top-down view (XZ-plane) as free space or target locations, we project them into the original 2D camera image to assess visibility. Each point is assigned the $y$ coordinate of the platform's top surface, forming full 3D coordinates.
Using camera intrinsics, extrinsics, and gravity alignment, we project these 3D points onto the 2D image plane, as shown in Fig.~\ref{fig:right_projected_points}, \ref{fig:below_projected_points}, \ref{fig:between_projected_points}. 
To determine visibility, we compare the $z$-coordinate of each 3D point with the corresponding depth value in the aligned depth image. Points are discarded as occluded if this difference exceeds $2.5$\,cm.
For the \textbf{front}, \textbf{behind}, \textbf{left}, and \textbf{right} directions, we sample $9,000$ points per direction and retain the direction if at least $2,000$ points remain visible.
For the \textbf{above}, \textbf{below}, and \textbf{between} relations, we sample $10,000$ points and require a minimum of $6,000$ visible points.
We compute the mean position of the remaining visible points as the representative target location. 
If its depth deviates from the depth image by more than $2.5$\,cm, we instead choose the nearest point within this threshold. 
Fig.~\ref{fig:right_visible_filtered}, \ref{fig:below_visible_filtered}, \ref{fig:between_visible_filtered} illustrate this process, highlighting visible points and the final selected target (blue circle).

\subsubsection{Generating Diverse QA via precise 3D annotations}
\label{subsubsec:qa_generation_3d}
Building on the 2D QA generation pipeline (Appx.~\ref{subsubsec:qa_generation_2d}), we utilize a template-based approach augmented with QwQ-32B to generate diverse QA pairs for 3D scenes.
\highlight{Unlike 2D images, 3D datasets offer rich and precise spatial annotations—such as depth maps, camera poses, and per-object 3D bounding boxes—which enable the construction of more complex QA with spatial referring with reasoning.}
Additionally, QA generation from 3D data accounts for above/below relations in both world and camera reference frames: the former reflects gravity, while the latter corresponds to vertical image orientation.
Leveraging these annotations, we design both qualitative and quantitative QA templates grounded in the following spatial concept categories:

\begin{enumerate}
  \item \textbf{Relative Position Relations:}  
  Encompasses spatial relations such as left/right, above/below, front/behind, inside/outside, touching/separated, and near/far. These queries require accurate 3D positioning and spatial layout to infer inter-object relationships in physical space.

  \item \textbf{Orientation and Rotation Reasoning:}  
  Involves reasoning over face/back direction, orientation (horizontal/vertical), and relative angles, using 3D object or camera poses (\eg, orientation vectors, rotation matrices) to infer facing direction or viewpoint shifts.

  \item \textbf{Geometric Attribute Comparisons:}  
  Covers attributes like size (big/small), height (tall/short), and width (wide/thin). These comparisons rely on true 3D dimensions, mitigating distortions from 2D projections.

  \item \textbf{Quantitative Spatial Reasoning:}  
  Involves computing depth, distance, relative angles, and spatial betweenness using precise 3D coordinates and metric reasoning.

  \item \textbf{Free Space Reasoning:}  
  Identifies free space above, below, or between objects. As illustrated in Fig.~\ref{fig:right_free_space_pipeline}, \ref{fig:below_free_space_pipeline}, and \ref{fig:between_free_space_pipeline}, blue-shaded regions represent unoccupied areas computed from object footprints and platform segmentation. To mitigate overestimation from large bounding boxes, we apply a shrink factor (\eg, 80\%) to the projected surfaces for above/below queries.

  \item \textbf{Location and Placement Prediction for Spatial Referring:}  
    Involves predicting precise 2D coordinates from language descriptions, \eg,  
    ``\textit{Point to the second chair from the left}'' — identifying a target object, or  
    ``\textit{Indicate a free spot to the right of the white box on the second shelf}'' — selecting a valid placement location.  
    These tasks require accurate 2D-3D projection and fine-grained spatial understanding, forming a vital bridge between visual perception and physical interaction and execution.
\end{enumerate}

Building on the structured templates, we design a diverse suite of 3D QA covering spatial reasoning, geometric comparison, viewpoint inference, environmental understanding, and coordinate-level localization. 
Leveraging QwQ-32B's powerful capabilities, our pipeline also generates complex reasoning QA pairs that are both structurally diverse and semantically rich.

\textbf{3D training data visualization.}
For specific examples of 3D training data and their visualizations, please refer to Appx.~\ref{suppsec: more demonstrations}, which contains detailed sample presentations.

\subsection{Synthetic Data Generation in the simulator}
\label{suppsubsec: Synthetic Data Generation}

We want to arm our model with multi-step referring capabilities with spatial reasoning. 
While 2D and 3D data enable single-step spatial understanding, they are less scalable for multi-step spatial referring with reasoning.
Therefore, we generate synthetic data in the simulator.

\subsubsection{Indoor Scene Generation}

\textbf{Initial Scene Generation.}
We utilize Infinigen~\cite{raistrick2024infinigen} to generate a large corpus of indoor scenes.
To be specific, we configure the generation process to exclude small objects by setting \texttt{compose\_indoors.solve\_steps\_small=0} to avoid pre-existing clutter on target surfaces.
This allows us to reserve space for the subsequent placement of our curated 3D assets. 
This initial step yields over 3k unique indoor scenes.

\textbf{Scene Filtering.} 
The generated scenes underwent a rigorous filtering process to ensure their suitability for our downstream tasks. 
The primary filtering criteria included:

\begin{itemize}
    \item \textbf{Adequate Tabletop Area:} Selected scenes contain at least one sufficiently large, continuous tabletop surface (\eg, desk, table, counter) suitable for object placement. Scenes with absent or impractically small surfaces are excluded.
    \item \textbf{Acceptable Lighting Conditions:} Scenes with extreme lighting issues (\eg, darkness, oversaturation, unnatural hues) are discarded to ensure a viable baseline for subsequent lighting adjustments.
    \item \textbf{Scene Realism and Coherence:} Scenes with severe geometric inconsistencies or implausible layouts are removed to maintain physical plausibility.
    \item \textbf{Camera Accessibility:} Scenes are required to support feasible camera placement with clear views of the target surfaces. Highly cluttered or confined environments are deprioritized.
\end{itemize}


\textbf{Scene Adjustments.} 
To enhance diversity and control experimental variables, filtered scenes underwent automated modifications:
\begin{itemize}
    \item \textbf{Lighting Randomization:} Light source intensities (\eg, ceiling lights, lamps) are uniformly scaled within $[0.6I, 1.4I]$, where $I$ denotes the original intensity.
    \item \textbf{Camera Pose Adjustment:} For each tabletop, camera viewpoints are defined with pitch angles randomly sampled from $[-60^\circ, -30^\circ]$ relative to the tabletop plane, oriented toward the region center.
    \item \textbf{Camera Height and Distance Variation:} Camera height is uniformly sampled from $0.3$--$0.8\,m$ above the tabletop. Distance to the target area is adjusted to maintain full visibility, conditioned on surface size and field of view.
\end{itemize}

\begin{figure}[h!]
  \centering
    \includegraphics[width=\textwidth]{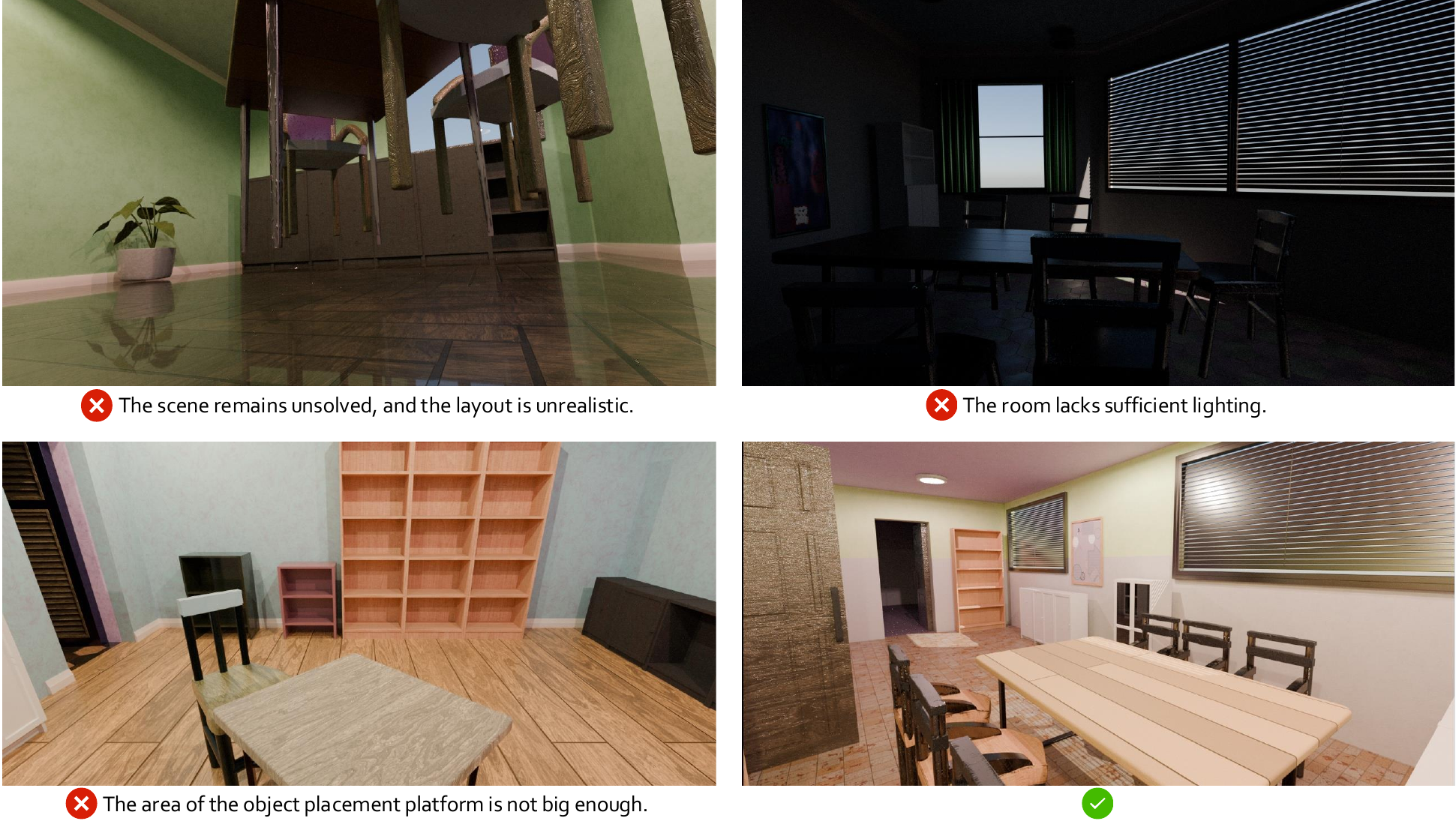}

    \caption{Cases of Scene Filtering}
  \label{fig:scene_filter}
\end{figure}

Some typical cases of scene filtering are shown in Fig.\ref{fig:scene_filter}.

\subsubsection{3D Asset Selection and Preparation}

Our 3D assets are sourced from the Objaverse~\cite{deitke2023objaverse} LVIS dataset and undergo a two-stage filtering process to ensure quality and relevance.

\textbf{Stage 1: Category Filtering.}
We select objects based on LVIS annotations, following categories:
\begin{itemize}
    \item Are typically placeable on surfaces.
    \item Have a maximum dimension under 1 meter, suitable for tabletop scenarios.
\end{itemize}

\textbf{Stage 2: Attribute-based Filtering.}
Next, we apply fine-grained filtering using attributes from the OrienText300K~\cite{qi2025sofar} dataset. Retained assets satisfy the following criteria:

\begin{itemize}
    \item \textbf{Axis Alignment:} Key features (\eg, edges, handles) align with canonical camera axes.
    \item \textbf{Single Object:} Represents a standalone object, not a scene or object collection.
    \item \textbf{Color Diversity:} Contains colors beyond white or gray.
    \item \textbf{No Ground Plane:} Excludes auxiliary visualization ground planes.
    \item \textbf{High Quality:} Clean, well-constructed geometry without artifacts.
    \item \textbf{Distinguishable Views:} Canonical views (front, back, top, bottom, left, right) exhibit meaningful visual or semantic differences.
    \item \textbf{Reasonable Object:} Represents a common, identifiable object, not an abstract shape or an unidentifiable entity.
\end{itemize}

This process yields a curated set of over 9k high-quality 3D assets.

\textbf{Stage 3: Manual Filtering.}
Due to the suboptimal quality of annotations, we perform manual verification based on the seven defined rules. 
Since object size estimation relies on LVIS-labeled categories with a ±30\% tolerance, we additionally discard instances whose irregular geometry (\eg, entangled cables of a wired mouse) distorts the bounding box and hinders reliable scaling. 
After filtering, we retain over 3k high-quality 3D assets compliant with our criteria. 

\begin{figure}[h!]
  \centering
    \includegraphics[width=\textwidth]{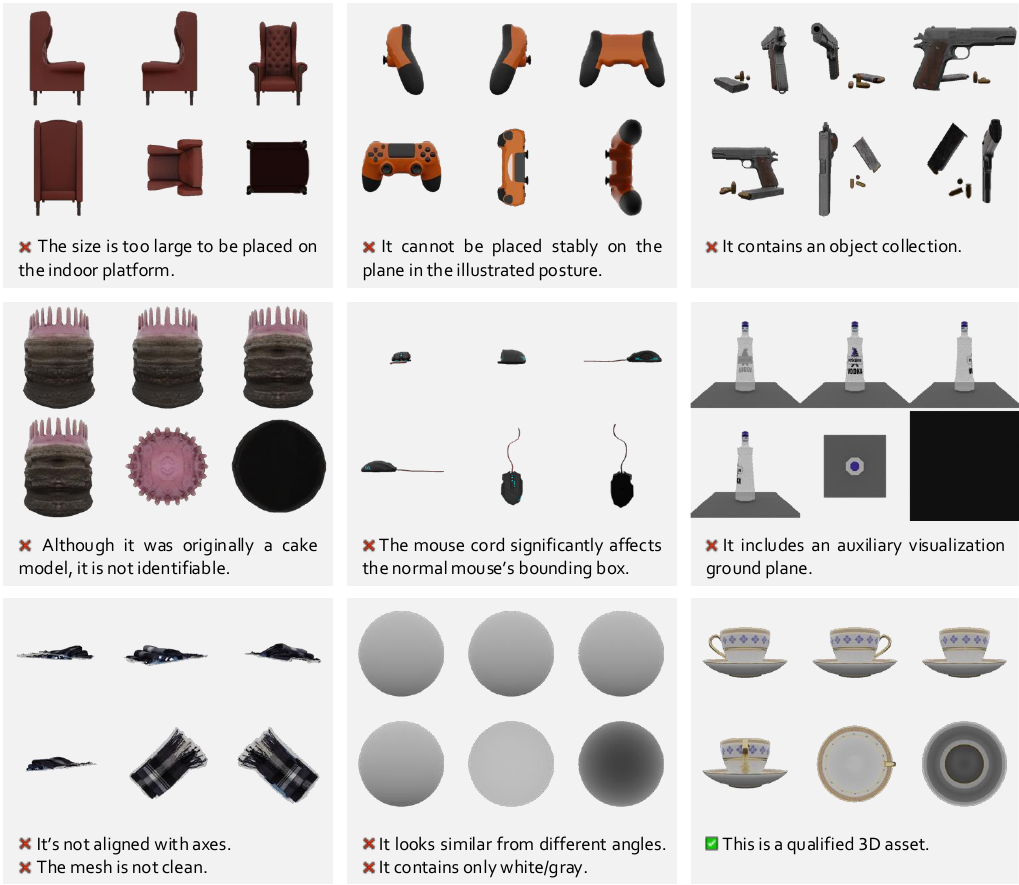}
    \caption{Cases of 3D assets Filtering}
  \label{fig:obj_filter}
\end{figure}

Fig.\ref{fig:obj_filter} shows the screening results for some representative 3D assets, which encompass all the criteria mentioned above.

\subsubsection{3D Asset Annotation with LLM}

To generate diverse annotations, we leverage OrienText300K’s orientation and caption data, processed via GPT-4o with tailored prompts to extract structured textual attributes for each sample.

\textbf{Generated Attributes:}
\begin{itemize}
    \item \textbf{Orientation Descriptions:} Prepositional phrases indicating the object's canonical front, salient parts, or intrinsic orientation (\eg, ``on the front of ...'', ``on the handle side of ...''), suitable for insertion into sentence templates.
    \item \textbf{Color Labels:} A single-word descriptor of the object's dominant color. If multiple colors are prominent, the attribute is marked as ``none'' (\eg, ``blue'', ``none'').
    \item \textbf{Object Labels:} Concise noun phrases specifying the object category (\eg, ``coffee mug'', ``computer mouse''), usable as subjects or objects in templates.
    \item \textbf{Category Consistency:} A boolean flag indicating alignment between the object’s visual category and its textual description.
\end{itemize}

\textbf{LLM Prompt:}
The exact prompt used to obtain these annotations is provided in Listing~\ref{lst:prompt_generating_brief_description}.




\begin{lstlisting}[basicstyle=\ttfamily\footnotesize, backgroundcolor=\color{myblue!50}, caption={Prompt of generating brief description, color and orientation preposition.}, captionpos=t, breaklines=false, label={lst:prompt_generating_brief_description}]
Your objective is to generate four distinct labels derived from the
provided 3D asset information. Each asset is characterized by the
following attributes:

Category: {category}
Detailed description: {description}
Direction hint: {direction_hint}

Based on the information above, you are required to perform
the following four tasks:

Task 1: Consistency Verification
Evaluate whether the asset's specified category aligns semantically
with its detailed description. Output "True" if consistent,
and "False" otherwise.

Task 2: Object Description Phrase Generation
ormulate a concise and unambiguous object description phrase. This
phrase must encapsulate the primary characteristics of the object,
not contain any articles (\eg, "a", "an", "the"), and be
grammatically suitable for use as either the subject or object
within a template sentence.

Task 3: Simplified Directional Phrase Generation
Generate a concise and clear simplified directional description phrase
based on the provided direction_hint. This phrase must be capable of
functioning as a prepositional phrase to describe the relative
position of other objects within a template sentence. In the generated
phrase, the current object should be represented by "OBJECT".

Task 4: Color Extraction
Extract the dominant color of the asset based on the detailed
description, adhering to the following criteria:
1. The extracted color must be a single English word in lowercase.
2. If the detailed description does not explicitly state a color,
return the string "none".

Please ensure your output strictly adheres to the following JSON
format. The output must be a valid JSON object without any
supplementary explanations, comments, or introductory/closing
remarks.
{
  "consistent": (True/False),
  "simple_desc": "simplified object description phrase",
  "simple_dir": "simplified directional description phrase",
  "color": "the main color"
}
\end{lstlisting}

\begin{figure}[h!]
  \centering
  \includegraphics[width=\linewidth]{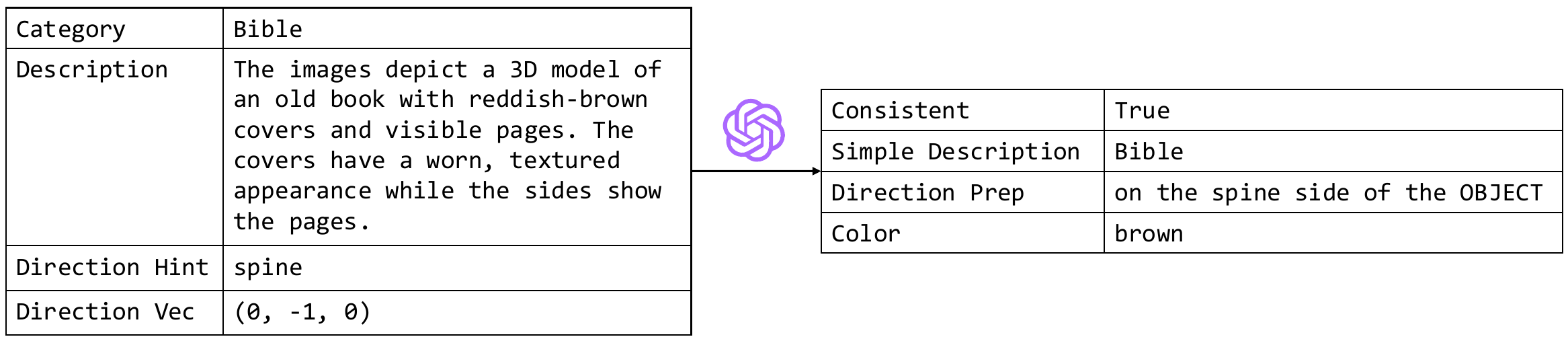} 
  \caption{A Sample of generating object phrases, colors, and orientation descriptions from 3D asset information and orientation hints.}
  \label{fig:obj_anno_sample}
\end{figure}

\subsubsection{Scene Population and Data Rendering}

The curated 3D assets are then programmatically placed into the filtered and adjusted indoor scenes.
\textbf{Asset Placement Strategy:}
\begin{itemize}
    \item Assets are placed predominantly on identified tabletop surfaces.
    \item To enhance the model's understanding of object orientation and inter-object relationships, placement strategies included:
        \begin{itemize}
            \item Increasing the proportion of objects with orientation vectors within the X-Y plane (\eg, laptops, teddy bears, mugs) in many scenes.
            \item Increasing the co-occurrence of objects from the same category that have significantly different features.  (\eg, a ceramic cup with a handle and a cola cup).
            \item Ensuring plausible physical arrangements (\eg, no excessive interpenetration, objects placed upright).
        \end{itemize}
    \item The number of assets per scene varied from 3 to 9.
\end{itemize}

\textbf{Rendering Details:}
\begin{itemize}
    \item \textbf{Renderer:} High-fidelity, physically-based images are rendered using the Blender Cycles.
    \item \textbf{Image Resolution:} Output images are rendered at $960 \times 540$ pixels.
    \item \textbf{Render Quality:} The \texttt{configure\_render\_cycles.num\_samples} parameter is set to 2048 to achieve high-quality renders while maintaining reasonably controlled noise.
\end{itemize}

\subsubsection{Question-Answer Pair Generation}

Based on the rendered scenes and their corresponding ground truth information (scene graphs, object properties, masks), we generate question-answering (QA) pairs. This process involves two steps:

\textbf{Step 1: Generating Unique Object Referring Expressions.}
For each object in a scene, we formulate unique identifiers or referring expressions based on a combination of its attributes and relationships:
\begin{itemize}
    \item \textbf{Feature Category:} The semantic class of the object (\eg, ``the mug'', ``the laptop'').
    \item \textbf{Color:} The primary color of the object (\eg, ``the red mug'').
    \item \textbf{Left-Right Rank:} Ordinal position from left to right (\eg, ``third bottle from left to right'').
    \item \textbf{Front-Back Rank:} Ordinal position from front to back (\eg, ``farthest LEGO minifigure'').
    \item \textbf{Distance Rank from Anchor:} Ordinal position based on distance from a salient anchor object (\eg, ``the closest plate to the blue mug'').
    \item \textbf{Height Rank:} Ordinal position based on the height (\eg, ``the second tallest teddy bear'').
\end{itemize}

These components are combined to create unambiguous references (\eg, ``the second red object from the left''). When filling objects into template structures in the subsequent process, we can randomly select from all referring expressions.

\textbf{Step 2: Applying QA Templates.}
Due to the heavy reliance on spatial relationships between objects, we initially developed a template structure to formalize these relationships: 
\begin{itemize}
    \item \textbf{Position:} Location of an object relative to another (\eg, ``on the left of the green bottle'').
    \item \textbf{Orientation:} Questions involving an object's intrinsic orientation (\eg, ``on the handle side of the red mug'').
    \item \textbf{Distance Queries:} The precise distance (\eg, ``0.2 meters to the left of the plate'').
    \item \textbf{Betweenness:} Identifying an object located between two other objects (\eg, ``between the stapler and the telephone'').
    \item \textbf{Specific Surface Locations:} Locating objects relative to parts of a surface (\eg, ``at the far left corner of the table'').
\end{itemize}

\textbf{QA Template Types:}
\begin{itemize}
    \item \textbf{Locate from Description:} Given a unique referring expression for an object, ask for its location (\eg, ``\textit{Give me the position of ...}'').
    \item \textbf{Identify from Relations:} Provide several spatial relationships an object satisfies and ask to identify the object (\eg, ``\textit{Please specify an object on the desktop that satisfies the following spatial constraints: ...}'').
    \item \textbf{Locate Empty Space:} Define a point in an empty area on a surface based on its spatial relationships with surrounding objects, and ask to confirm this empty location (\eg, ``\textit{Please provide a point in the vacant area on the desktop that simultaneously satisfies the following spatial conditions: ...}'').
\end{itemize}

\textbf{Generation of Thought Processes (Reasoning Steps):}
For each QA pair, a structured thought process or chain of reasoning is also generated. This involved selecting the pertinent pieces of information from the complete scene graph and ground truth data that are necessary to arrive at the answer. This recorded reasoning follows a predefined format.

\section{Implementation Details and Samples of {\bname}}
\label{suppsec: implementation_of_benchmark}
The {\bname} benchmark evaluates spatial referring with reasoning in complex 3D indoor scenes through two tasks: \textbf{Location Prediction} and \textbf{Placement Prediction}, each comprising $100$ samples.
Each sample includes a manually selected image, a referring caption, and precise mask annotations.
Moreover, to evaluate the effectiveness of the RFT training strategy, we further select $77$ samples from these $200$ samples and define it as the \textbf{Unseen} set, which contains novel spatial relation combinations absent from {\dname} to test the model's generalization.


\textbf{Location Task:}  
Given an indoor scene and a unique referring expression, the model predicts a 2D point indicating the target object. Expressions may reference color, shape, spatial order (\eg, ``the second chair from the left''), or spatial anchors.

\textbf{Placement Task:}  
Given a caption specifying a free space (\eg, ``the vacant area to the right of the white box on the second shelf''), the model predicts a 2D point within that region. Queries often involve complex spatial relations, multiple anchors, hierarchical references, or implied placements.

\textbf{Unseen Set:}  
This set comprises $77$ samples from the Location/Placement task, specifically designed to evaluate model generalization after SFT/RFT training on {\dname}, as it includes novel spatial relation combinations not present in {\dname}.
%

Notably, \highlight{we introduce \textit{reasoning steps} (\textit{step}) for each benchmark sample, quantifying the number of anchor objects and their associated spatial relations that effectively narrow the search space.}
Specifically, each \textit{step} corresponds to either an explicitly mentioned anchor object or a directional phrase linked to an anchor that greatly reduces ambiguity (\eg, ``on the left of’’, ``above’’, ``in front of’’, ``behind’’, ``between’’).
We exclude the ``viewer’’ as an anchor and disregard the spatial relation ``on’’, since it typically refers to an implied surface of an identified anchor, offering minimal disambiguation.
Intrinsic attributes of the target (\eg, color, shape, size, or image-relative position such as ``the orange box’’ {or ``on the right of the image''}) also do not count towards \textit{step}.

\highlight{A higher \textit{step} value indicates increased reasoning complexity, requiring stronger spatial understanding and reasoning about the environments.
Empirically, we find that beyond $5$ \textit{steps}, additional qualifiers yield diminishing returns in narrowing the search space. 
Thus, we cap the \textit{step} value at $5$. 
Instructions with \textit{step} $\geq$ 3 already exhibit substantial spatial complexity.}
Detailed statistics on \texttt{step} distributions and instruction lengths are provided in the Tab.~\ref{tab:benchmark_stats}.
To further show the diversity and reasoning complexity of {\bname}, we present representative examples from both the Location and Placement tasks. 
Fig.~\ref{fig:location_example1}, \ref{fig:location_example2}, \ref{fig:location_example3}, and \ref{fig:location_example4} show Location queries with varying reasoning step counts, where {\mname} accurately localizes the target object (marked by a blue dot). 
Similarly, Fig.~\ref{fig:placement_example1}, \ref{fig:placement_example2},  \ref{fig:placement_example3}, and \ref{fig:placement_example4} show Placement queries involving the identification of free space based on spatial relations.
These examples highlight the step-wise complexity of the queries and the effectiveness of {\mname} in addressing challenging spatial referring tasks.

\begin{table}[ht]
\small
\centering
\caption{Statistics of the {\bname} across Location/Placement tasks and unseen sets.}
\label{tab:benchmark_stats}
\begin{tabular}{lcccc}
\toprule
\textbf{{\bname}} & \textbf{Step} & \textbf{Samples} & \textbf{Avg. Prompt Length} \\
\midrule
\multirow{3}{*}{\textbf{Location}} 
& Step 1 & 30 & 11.13 \\
& Step 2 & 38 & 11.97 \\
& Step 3 & 32 & 15.28 \\
\cmidrule{2-4}
& \multicolumn{2}{r}{\textbf{Avg. (All)}} & 12.78 \\
\midrule
\multirow{4}{*}{\textbf{Placement}} 
& Step 2 & 43 & 15.47 \\
& Step 3 & 28 & 16.07 \\
& Step 4 & 22 & 22.68 \\
& Step 5 & 7  & 22.71 \\
\cmidrule{2-4}
& \multicolumn{2}{r}{\textbf{Avg. (All)}} & 17.68 \\
\midrule
\multirow{4}{*}{\textbf{Unseen}} 
& Step 2 & 29 & 17.41 \\
& Step 3 & 26 & 17.46 \\
& Step 4 & 17 & 24.71 \\
& Step 5 & 5  & 23.8 \\
\cmidrule{2-4}
& \multicolumn{2}{r}{\textbf{Avg. (All)}} & 19.45 \\
\bottomrule
\end{tabular}
\end{table}

To more comprehensive evaluate the spatial referring task, we expand the original RefSpatial-Bench in terms of difficulty and diversity, producing the manually annotated RefSpatial-Expand-Bench. It includes more indoor cases (\eg, shops and factories), and also introduces outdoor scenes not present in RefSpatial-Bench (\eg, streets, parking lots, and parks). Statistics of this extension are provided in Tab.~\ref{tab:expand_stats} and Tab.~\ref{tab:expand_step_stats}. The detailed evaluation results of  RoboRefer on this expanded benchmark are showed in Tab.~\ref{tab:accuracy_comparison}.
\begin{table}[h]
\centering
\small
\caption{Statistics of the RefSpatial-Expand-Bench}
\label{tab:expand_stats}
\begin{tabular}{lccc} 
\toprule
\textbf{Task Type} & \textbf{Indoor} & \textbf{Outdoor} & \textbf{Total} \\
\midrule
Location           & 115             & 126              & 241            \\
Placement          & 120             & 80               & 200            \\
\midrule
\textbf{Total}     & \textbf{235}    & \textbf{206}     & \textbf{441}   \\
\bottomrule
\end{tabular}
\end{table}


\begin{table}[ht]
\small
\centering
\caption{Statistics of the RefSpatial-Expand-Bench by step and task.}
\label{tab:expand_step_stats}
\begin{tabular}{lccc}
\toprule
\textbf{Task Type} & \textbf{Step} & \textbf{Samples} & \textbf{Avg. Prompt Length} \\
\midrule
\multirow{3}{*}{\textbf{Location}}
& Step 1 & 54  & 10.61 \\
& Step 2 & 129 & 12.56 \\
& Step 3 & 58  & 16.10 \\ 
\cmidrule(lr){2-4}
& \textbf{Avg. (All)} & 241 & 12.98 \\
\midrule
\multirow{5}{*}{\textbf{Placement}}
& Step 1 & 3   & 15.00 \\ 
& Step 2 & 86  & 15.14 \\
& Step 3 & 75  & 16.95 \\
& Step 4 & 29  & 22.24 \\
& Step 5 & 7   & 22.71 \\
\cmidrule(lr){2-4}
& \textbf{Avg. (All)}& 200 & 17.11 \\ 
\bottomrule
\end{tabular}
\end{table}
\begin{table}[ht]
\small
\centering
\caption{Accuracy results of 2B SFT and 8B SFT Models on RefSpatial-Expand-Bench.}
\label{tab:accuracy_comparison}
\begin{tabular}{llcc}
\toprule
\textbf{Task} & \textbf{Category} & \textbf{2B SFT} & \textbf{8B SFT} \\
\midrule
\multirow{6}{*}{\textbf{Location}}
                     & Over all          & 50.21 & 61.00 \\
\cmidrule(lr){2-4}
                     & Indoor            & 49.57 & 58.26 \\
                     & Outdoor           & 50.79 & 63.49 \\
\cmidrule(lr){2-4}
                     & Step 1            & 61.11 & 72.22 \\
                     & Step 2            & 52.71 & 62.02 \\
                     & Step 3            & 34.48 & 48.28 \\
\midrule
\multirow{8}{*}{\textbf{Placement}}
                     & Over all          & 48.50 & 60.00 \\
\cmidrule(lr){2-4}
                     & Indoor            & 50.83 & 60.00 \\
                     & Outdoor           & 45.00 & 60.00 \\
\cmidrule(lr){2-4}
                     & Step 1            & 33.33 & 33.33 \\
                     & Step 2            & 41.86 & 51.16 \\
                     & Step 3            & 54.67 & 70.67 \\
                     & Step 4            & 48.28 & 55.17 \\
                     & Step 5            & 71.43 & 85.71 \\
\bottomrule
\end{tabular}
\end{table}
\section{Implementation Details for {\mname}}
\label{suppsec: implementation details}

\subsection{Architecture}
\label{suppsubsec: architecture}

We adopt NVILA~\cite{liu2024nvila} as base model, including a visual encoder, an LLM, and a multimodal projector.

\textbf{Visual Encoder.} 
We use the same image encoder as \texttt{siglip-so400m-patch14-448}~\cite{zhai2023sigmoid} from NVILA~\cite{liu2024nvila}, supporting $448 \times 448$ resolution for richer visual details.
Rather than simply resizing the image to a fixed resolution and producing the same number of tokens, this image encoder processes inputs at dynamic resolutions, yielding more visual tokens from higher-resolution images via finer patch division.
This enables fine-grained vision-language understanding, crucial for tasks like point prediction that require detailed perception beyond VQA.
We further incorporate a dedicated depth encoder, structurally mirroring the image encoder and initialized with its weights.
It encodes relative depth maps as special images, providing spatial cues to enhance 3D understanding.

\textbf{Large Language Model.} 
We adopt the Qwen2 LLM backbone from NVILA~\cite{liu2024nvila}, which has been fully fine-tuned with extensive data during supervised training. This endows the model with rich visual knowledge, facilitating downstream 3D spatial understanding and reasoning tasks.

\textbf{Multi-Modal Projector.} 
To align multi-modal representations (\eg, image to language, depth to language), we use linear connectors, the same as NVILA~\cite{liu2024nvila}, which is better than Q-Former, to allow the LLM to focus on visual understanding and improve generalization. 
Separate connectors for image and depth embeddings ensure modality-specific processing and prevent cross-modal interference.

\subsection{Training Data}
\label{suppsubsec: training data}
Here we highlight the training data used at each stage, including the number of samples per dataset and the overall total.
See Tab.~\ref{supptab: dataset} for details.

\textbf{SFT stage.}
Specifically, in the first step of the SFT stage, \ie, depth alignment, we train a depth projector to align depth and language space using the {\dname} (RGB-D) dataset with $2.5$M samples.
To increase training efficiency, we slice multi-turn conversations (up to $15$ turns per sample), yielding $3.4$M samples post-processing to train our model.
In the second step, \ie, spatial understanding enhancement via full-parameter fine-tuning—we use both {\dname} (RGB) and {\dname} (RGB-D) datasets, yielding $6.8$M samples after slicing.
To further improve instruction-following and referring capabilities, we incorporate auxiliary datasets: $965$k samples from instruction-tuned data (LLaVA-1.5~\cite{liu2024improved}, LRV~\cite{liu2024mitigating}), $321$k from referring datasets (RefCOCO/+/g~\cite{yu2016modeling}), $176$k from SAT~\cite{ray2024sat} benchmark training sets, and $127$k from EmbSpatial~\cite{du2024embspatial} benchmark training sets.
These additions help bridge distribution gaps between {\dname} and benchmark-style queries. 
After slicing, the total number of samples used in this stage reaches $8.5$M post-slicing.

\textbf{RFT Stage.}
In the RFT stage, we train the model using {\dname} data annotated with detailed reasoning processes, including key intermediate steps and final answers.
To ensure both training efficiency and effective learning, we use moderately difficult samples (typically involving three reasoning steps), resulting in a $100$k-sample dataset.

\begin{table}[t]
\caption{
Details about the training datasets used in the SFT and RFT stages.
D.A. and S.U.E denote the Depth Alignment and Spatial Understanding Enhancement step in the SFT stage, respectively.}
\footnotesize
\centering
\setlength{\tabcolsep}{1pt}
\begin{tabular}{l|l|l}
\toprule                          

Stage & Categories & Datasets \\
\midrule
SFT (D.A) & Spatial & {\dname} (RGB-D) \\
\midrule
\multirow{3}{*}{SFT (S.U.E)} & Spatial & {\dname} (RGB), {\dname} (RGB-D), SAT~\cite{ray2024sat}, EmbSpatial~\cite{du2024embspatial} \\
 & General & COCO~\cite{lin2014microsoft}, GQA~\cite{hudson2019gqa}, OCR-VQA~\cite{mishra2019ocr}, TextVQA~\cite{singh2019towards}, VG~\cite{krishna2017visual}, LRV~\cite{liu2024mitigating} \\
 & REC & RefCOCO/+/g~\cite{yu2016modeling} \\
\midrule
RFT & Spatial & {\dname} (RGB-D) w/ Reasoning Processing \\


\bottomrule[1pt]
\end{tabular}
\label{supptab: dataset}
\end{table}

\subsection{SFT Training Details}
\label{suppsubsec: SFT training details}

We formulate the SFT training stage as follows: given a dataset $\mathcal{D}$ consisting of samples in the form of triplets ($\mathcal{O}$, $\mathcal{Q}$, $\mathcal{A}$), where $\mathcal{O}$ is a sensor image (either RGB or RGB-D), $\mathcal{Q}$ is a textual question, and $\mathcal{A}$ is the corresponding answer.
The answer $\mathcal{A}$ may be a direct response (\eg, a point coordinate) or include intermediate reasoning steps (\eg, perceptual results followed by the final answer). 
The training objective is to maximize the likelihood of generating the answer given the input pair ($\mathcal{Q}$, $\mathcal{A}$):

{
\begin{equation}
\mathcal{L}_{\mathrm{SFT}}=-\mathbb{E}_{(\mathcal{O}, \mathcal{Q}, \mathcal{A})\sim\mathcal{D}}\sum_{t=1}^T\log\pi_\theta(y_t\mid \mathcal{O}, \mathcal{Q}, y_{<t}),
\tag{1}
\end{equation}
}

where $\pi_\theta$ is the model’s token distribution. 
The output model $\pi_\mathrm{SFT}$ serves as the initialization for the next RFT stage, ensuring a robust foundation for reinforcement learning.

To be specific, our SFT consists of two steps. 
In the first step, depth alignment, only the depth projector is updated by using the {\dname}~(RGB-D). We employ a maximum learning rate of 1e-4, a weight decay of 0, and a warm-up ratio of 0.03. 
The 2B variant is trained with a batch size of 7 per GPU, and the 8B variant with 3, both for one epoch.
In the second step of spatial understanding enhancement, we fine-tune all model parameters using the datasets described in Sec.~\ref{suppsubsec: training data}.
Training is conducted for one epoch with a maximum learning rate of 5e-5. We use a batch size of 6 per GPU for the 2B model and 2 for the 8B model. 
Other hyperparameters follow those in the first step.
For more details, please refer to NVILA~\cite{liu2024nvila} settings during alignment and SFT.

\subsection{RFT Training Details}
\label{suppsubsec: RFT training details}

During the RFT stage, we refine $\pi_\mathrm{SFT}$ via GRPO~\cite{shao2024deepseekmath}, a reinforcement learning method designed for efficiency and scalability. 
Unlike PPO~\cite{schulman2017proximal}, which relies on a costly value network, GRPO estimates relative advantages by comparing intra-group rewards, reducing computation, and simplifying optimization. 
This makes it well-suited for reasoning-intensive spatial referring tasks.
In detail, we modify R1-V~\cite{zhang2025r1} to support our 3D-aware architecture.
Training is conducted for two epochs with a batch size of 1 per GPU and 8 outputs in GRPO.
For details about hyperparameters, see R1-V~\cite{zhang2025r1}.

\subsubsection{Sampling Action Groups}
Given an input state \( s = (\mathcal{O}, \mathcal{Q}) \), where \( \mathcal{O} \) denotes the visual encoding of the RGB or RGB-D observation and \( \mathcal{Q} \) the textual encoding of the question, GRPO samples a set of actions \( \{a_1, a_2, \dots, a_N\} \) from the current policy \( \pi_\theta \), initialized from \( \pi_{\mathrm{SFT}} \).
The sampling process is:

{
\begin{equation}
a_i\sim\pi_\theta(a\mid \mathcal{O}, \mathcal{Q}),\quad\mathrm{for~}i=1,2,\ldots,N
\tag{2}
\end{equation}
}

This strategy ensures diverse responses, promoting exploration and preventing premature convergence.

\subsubsection{Reward Design and Policy Update}
\label{suppsubsubsection: reward design and policy update}
Each sampled action $a_i$ is assigned a reward $ R (a_i) $ based on verifiable criteria, yielding a reward set $ r_1, r_2, \ldots, r_N$. 
For spatial referring tasks, $R(a_i)$ integrates two outcome-based and our proposed two process-based components.
The outcome-based reward functions are defined as follows: 

\textbf{Outcome Format Reward $R_{OF}$.}
This component ensures structured and interpretable outputs by requiring the model to a predefined format: reasoning within ``$\texttt{<think>}\ldots\texttt{</think>}$'' and the final answer in ``$\texttt{<answer>}\ldots\texttt{</answer>}$''. A reward is assigned 1 for strict compliance, 0 otherwise.

\textbf{Point L1 Reward $R_{P}$.}
This component evaluates the accuracy of the model’s final point prediction by comparing it with the ground truth from the annotations of {\dname}.
Following the criterion inspired by Seg-zero~\cite{liu2025seg}, a stricter reward of 1 is assigned if the L1 distance between the predicted and ground-truth points is within 50 pixels; otherwise, the reward is 0.

Notably, most process-based rewards depend on a Process Reward Model (PRM), typically a fine-tuned LLM or VLM tasked with providing feedback. 
However, applying such an approach in our setting presents two main challenges.
\textbf{(1)} LLMs cannot process images, making it impossible to determine whether predicted coordinates match the target object.
\textbf{(2)} Although VLMs integrate visual and textual information, prior work~\cite{majumdar2024openeqa} has shown they may lack precise visual understanding when dealing with textual coordinates.
Since the correct assessment of predicted coordinates is paramount for reward assignment, additional or specialized methods are needed to ensure reliable feedback.

\highlight{To address this issue, we propose a rule-based process reward for spatial referring that obviates the need for a Process Reward Model.}
Our approach directly evaluates key intermediate perceptual steps using the ground-truth step-wise annotations provided in {\dname}.
This contrasts with most existing methods on process-based rewards~\cite{khalifa2025process, liu2025can}, which emphasize strictly sequential reasoning and rely on a PRM for evaluation.
\highlight{In contrast, our method employs \textit{metric-sensitive} rule-based process reward functions to assess intermediate perceptual results in an \textit{order-invariant} manner.}
Our key insight lies in two aspects:
\highlight{(1) Metric-sensitivity}: Different spatial attributes require distinct metrics due to inherent differences in their representations (\eg, points for positions, vectors for orientations).
\highlight{(2) Order-invariance}: The reasoning process in spatial referring is not strictly sequential; for instance, identifying the position of the keyboard or the mouse first does not affect the final interpretation of ``the free area between the keyboard and the mouse''.

In detail, we have two process-based reward functions:

\textbf{Process Format Reward $R_{PF}$.}
Similar to the Outcome Format Reward strategy, this component enforces a structured and interpretable reasoning process, thereby facilitating accurate reward computation. In particular, the model is required to produce outputs in the following format:
\[
\texttt{[Perception Type] [Target Object]: [Value]}
\tag{3}
\]

where ``\texttt{Perception Type}'' must be one of three categories: ``\texttt{Position}'', ``\texttt{Orientation}'', or ``\texttt{Size}''.
The ``\texttt{Target Object}'' corresponds to a uniquely identifiable entity (\eg, ``\texttt{the second largest cup}'' or ``\texttt{the second large cup from large to small}''). 
The “\texttt{Value}” depends on the selected ``\texttt{Perception Type}'':

\begin{itemize}
  \item For ``\texttt{Position}'', the value should be a normalized 2D coordinate of the form \texttt{[(x, y)]}, where both \(x\) and \(y\) lie in the interval \([0, 1]\), rounded to three decimal places.  
  \item For ``\texttt{Orientation}'', the value is a 3D unit vector \texttt{(x, y, z)} representing the object's semantic orientation in the camera coordinate system.
  \item For ``\texttt{Size}'', the value represents a scalar measured in meters.
\end{itemize}

Below are examples to illustrate the expected format:

\begin{itemize}
  \item \texttt{[Position] [the second largest cup]}: [(0.245, 0.147)]
  \item \texttt{[Orientation] [the handle of the second largest cup]}: (1.000, 0.000, 0.000)
  \item \texttt{[Size] [the second largest cup]}: 0.12
\end{itemize}

\textbf{Accuracy Reward $R_{Acc}$.}
The reward is computed only for steps annotated as key steps in {\dname}. 
In detail, we use regex matching to determine whether the ``\texttt{Target Object}'' in the current process format appears in the key-step annotations. 
If not, the step receives no reward.
Since the model has already undergone a cold-start phase in SFT, it can interpret instructions and identify relevant target objects. 
Thus, a failed match implies that the model cannot accurately refer to the object linguistically, and no reward is assigned.
For each perception type, we apply a specific metric to compute the reward:
(1) ``\texttt{Position}'': If the L1 distance between the predicted point and the ground truth is below $50$ pixels, the reward is $1$; otherwise, $0$.
(2) ``\texttt{Orientation}'': If the cosine similarity between the predicted and ground-truth vectors exceeds $0.8$, the reward is $1$; otherwise, $0$.
(3) ``\texttt{Size}'': If the predicted value falls within $\pm15$\% of the ground truth, the reward is $1$; otherwise, $0$.

We prioritize the correctness of the final outcome over intermediate steps. 
To prevent reward accumulation from multi-step processes, we scale the process reward by $0.25$. The final reward function is defined as:
\[
r_i = R_{OF}(a_i) + R_{P}(a_i) + \alpha R_{PF}(a_i) + \alpha R_{Acc}(a_i)
\tag{4}
\]

where $\alpha$ is set to $0.25$.
By normalizing the rewards within the sampled group, we obtain the set of relative advantages \(\{A_1, A_2,\ldots, A_N\}\) defined as

\[
A_i = \frac{r_i - \text{mean}(\{r_j\})}{\text{std}(\{r_j\})},
\tag{5}
\]

which measures how each reward compares to the mean in units of standard deviation. We then update the policy based on these advantages, reinforcing actions with higher relative advantages while reducing the likelihood of those deemed less effective.
To ensure stable reinforcement learning, the update is further constrained by minimizing the KL divergence between the updated policy and its reference counterpart, thereby promoting incremental and controlled policy adjustments.

\section{Experimental Setting and Details}

\subsection{Experiments Compute Resources}
\label{suppsubsec: compute resources}

We conduct experiments on an A100 GPU cluster, with each node equipped with 8 GPUs.

\textbf{2D Web Data Coarse Filtering.}
We perform the initial coarse filtering of 1.7M OpenImages using SigLIP2. 
The process runs on 1 node and takes 8.5 hours and yields 933k high-quality samples.

\textbf{2D Web Data Fine-grained Filtering.}
We further filter 933K samples using Qwen 2.5-VL 7B to ensure high visual quality and spatially relevant QA pairs. 
The process is conducted on 1 node (two models per GPU) over 2.5 days, yielding approximately 845k high-quality samples.

\textbf{Pseudo-3D Scene Graphs Construction.}
We construct pseudo-3D scene graphs for 845k samples using 3 nodes, requiring 10 hours for depth estimation and another 10 hours for camera parameter extraction, segmentation masks, and point cloud generation. 
Additionally, we generate object-level captions for all instances using 4 nodes over 18 hours by using Qwen 2.5-VL.

\textbf{Reasoning QA Generation from 2D data source.}
To enrich factual statements with contextual scenarios, we employ QwQ-32B to construct reasoning QA, utilizing 4 nodes over 3.75 days.

\textbf{3D Data Filtering and Scene Graphs Construction.}
Given the limited amount of 3D data (100k) and the availability of precise annotations, only 2D bounding box bidirectional matching is required. This process is completed in 3 hours using 1 node.

\textbf{Reasoning QA Generation from 3D data source.}
To enrich factual statements with contextual scenarios, we employ QwQ-32B to construct reasoning QA, utilizing 4 nodes over 1.5 days.

\textbf{Synthetic Data Generation in Simulator.}
We use 4 RTX 4090 GPUs to generate data for one week.

\textbf{Depth Alignment in SFT.}
The process is conducted on 10 nodes over 12 hours for 2B variants and 8 nodes over 40 hours for 8B variants. Both variants training use ZeRO-3.

\textbf{Spatial Understanding Enhancement in SFT.}
The process is conducted on 10 nodes over 2 days for 2B variants and 10 nodes over nearly 1 week for 8B variants. Both variants training use ZeRO-3.

\textbf{Spatial Referring in RFT.}
The process is conducted on 1 node over 3 days for 2B variants.
However, our model is over twice as slow as other Qwen 2/2.5-VL-based methods~\cite{zhang2025r1, shen2025vlm}, mainly because they process only a single RGB image during training and can leverage vLLM for group inference acceleration. In contrast, our method requires RGB-D inputs and modifies the original NVILA architecture, making it incompatible with vLLM or SGLang acceleration.

\subsection{Spatial Understanding Benchmarks}
\label{suppsubsec: spatial understanding benchmarks}

We evaluate several public \textit{single-step spatial understanding} benchmarks, including CV-Bench~\cite{tong2024cambrian} (2D Spatial Relation, 3D Depth Order, 3D Distance), the BLINK~\cite{fu2024blink} validation set (Spatial Relation, Relative Depth), RoboSpatial~\cite{song2024robospatial} (configuration), SAT~\cite{ray2024sat}, and EmbSpatial~\cite{du2024embspatial}, following their official evaluation protocols. 
We exclude non-spatial tasks from our evaluation, such as 2D Counting in CV-Bench and Art Style or IQ Test in BLINK.
Since all these benchmarks are multiple-choice tasks, we report accuracy as the evaluation metric.

We compare $3$ categories of models: 
(1) proprietary VLMs, such as Gemini-2.5-Pro~\cite{team2023gemini}, which show strong spatial perception, as shown in the Gemini-Robotics~\cite{team2025gemini} paper; 
(2) open-source VLMs trained on general VQA datasets; 
and (3) spatially specialized models trained on spatially relevant datasets, offering basic spatial understanding.

\subsection{Spatial Referring Benchmarks}
\label{suppsubsec: spatial referring benchmarks}

We evaluate three recent robotic referring benchmarks—RoboRefIt~\cite{lu2023vl}, Where2Place~\cite{yuan2024robopoint}, and RoboSpatial~\cite{song2024robospatial}, all limited to $2$ reasoning steps. 
Specifically, RoboRefIt concentrates on object location referring by leveraging object attributes and spatial relations with anchor objects. 
Where2Place further explores how to place objects relative to anchor objects in the camera’s coordinate frame.
RoboSpatial builds on this idea, considering the same form of placement relative to anchor objects, but many samples in benchmarks consider the object-centric coordinate frame.
We also evaluate more complex multi-step spatial referring on {\bname}, a challenging benchmark based on real-world cluttered scenes, as introduced in Appx.~\ref{suppsec: implementation_of_benchmark}. 
For all these benchmarks, we use the same evaluation protocol: we first compute the proportion of predicted points that fall within the ground truth mask for each sample, and then average the results across all samples to obtain the success rate.

We compare two main categories of models: (1) Proprietary models (\ie, Gemini-2.5-Pro) with strong spatial referring capabilities, and (2) spatially specialized models trained on spatially relevant datasets, exhibiting basic spatial referring abilities.

\subsection{Simulation Evaluation}
\label{suppsubsec: simulation evaluation}

We use the same evaluation protocol of Open6DOR V2 introduced in SoFar~\cite{qi2025sofar}, following the official repository\footnote{\href{https://github.com/Zhangwenyao1/Open6DOR_V2_Execution}{https://github.com/Zhangwenyao1/Open6DOR\_V2\_Execution}}.
Specifically, we only test the position track as this work focuses on location and placement via spatial referring rather than executing 6DOF manipulation tasks.
Notably, we find that our model achieves nearly 100\% success in the perception stage (\ie, determining location and placement), with failures primarily attributed to motion planning errors such as IK failures or collision-prone trajectories.
We show more demonstrations of simulation evaluation in Appx.~\ref{suppsec: more demonstrations}.

\subsection{Real-world Evaluation}
\label{suppsubsec: real-world evaluation}

\subsubsection{UR5 Manipulation}

We show two demos for UR5 \footnote{\href{https://www.universal-robots.com/products/ur5e/}{https://www.universal-robots.com/products/ur5e/}} Manipulation: human disturbance and voice interruption.
In the human disturbance case, {\mname} runs at 2.5Hz. 
Significant shifts in predicted 2D coordinates trigger motion interruption and re-planning.
In the voice interruption case, incoming speech commands are continuously monitored. 
Upon detection, the current task is halted. We use the Whisper~\cite{radford2023robust} ASR model to transcribe speech, which {\mname} processes into new 2D coordinates for task redirection.


For grasping, the 2D coordinates are fed into SAM2~\cite{ravi2024sam} to generate a segmentation mask, which filters the target object's point cloud from the scene captured by a third-person Intel RealSense L515~\footnote{\href{https://www.intelrealsense.com/lidar-camera-l515/}{https://www.intelrealsense.com/lidar-camera-l515/}} depth camera. 
The extracted point cloud is input to AnyGrasp~\cite{fang2023anygrasp} to predict a grasp pose in the camera coordinate frame. 
Using an eye-to-hand calibration method, the grasp pose is transformed into the UR5 robot's base frame for execution.

For placement, {\mname} predicts the 2D placement point, which is converted to 3D coordinates using the camera's intrinsic parameters and depth data. 
The 3D point is then transformed into the robot’s coordinate system to guide the placement action.

\begin{figure}
    \centering
    \includegraphics[width=0.9\linewidth]{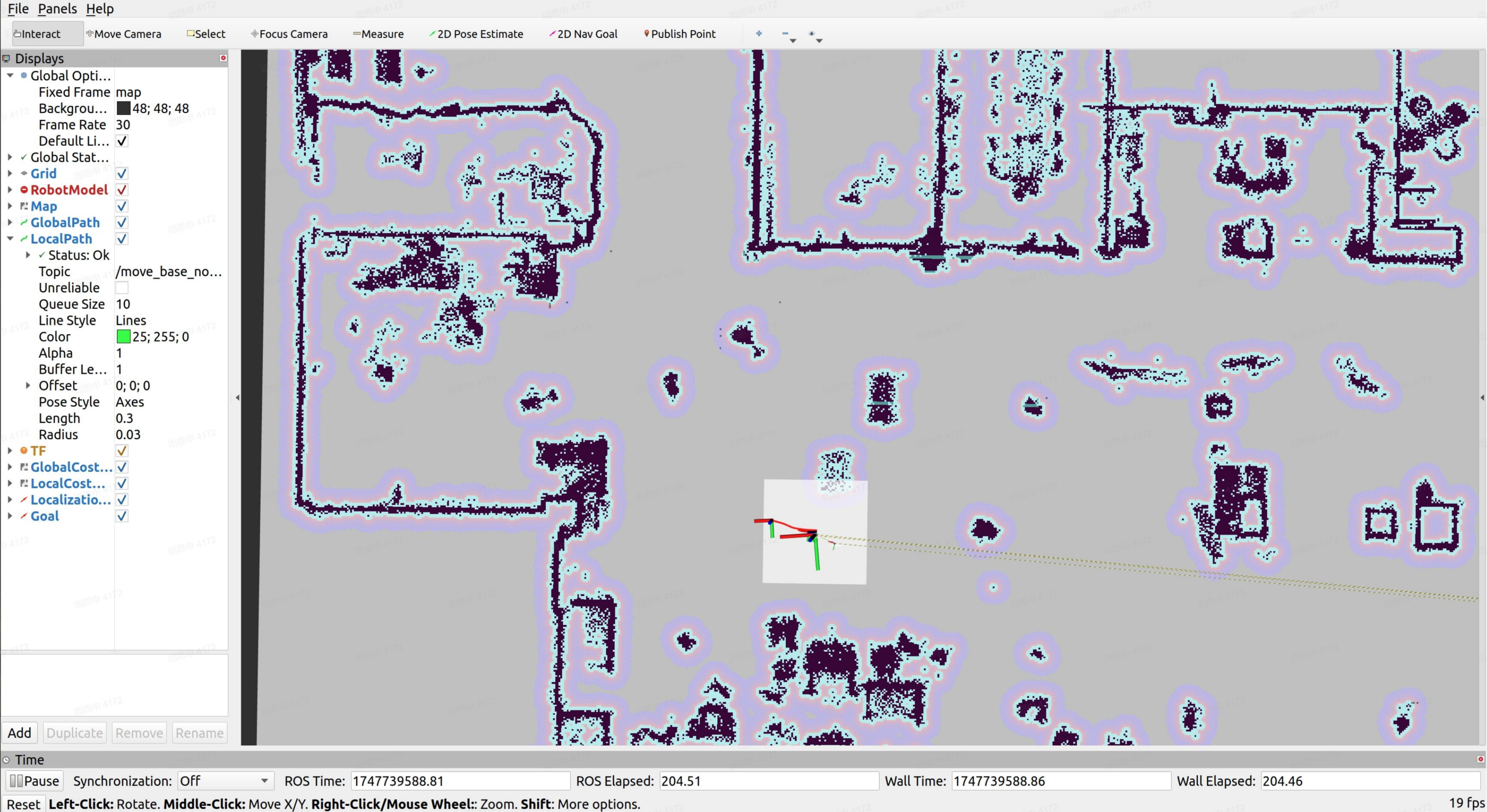}
    \caption{Map Visualization (RViz).}
    \label{fig:enter-label}
\end{figure}
\begin{figure}
    \centering
    \includegraphics[width=0.9\linewidth]{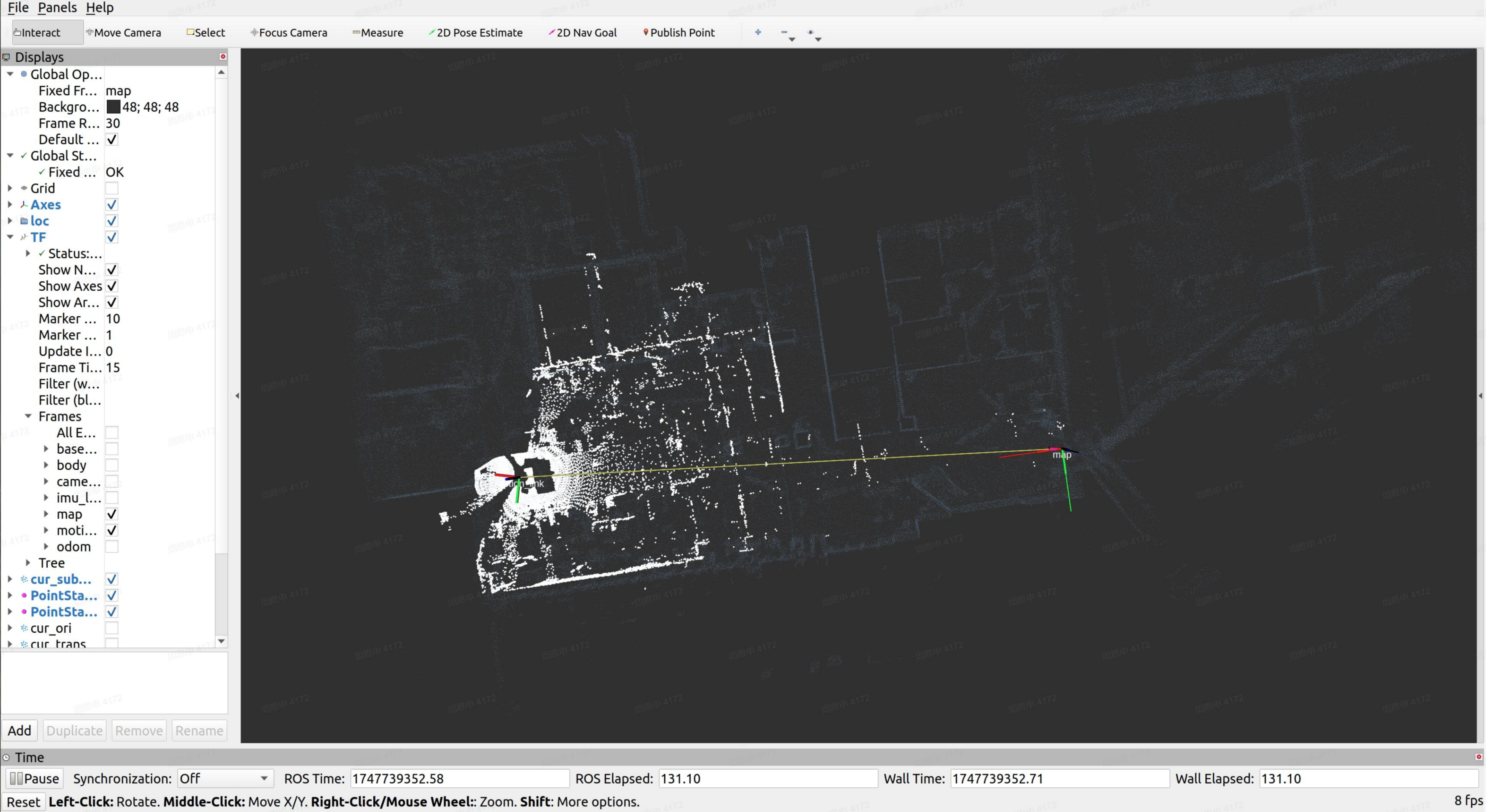}
    \caption{SLAM Navigation (RViz).}
    \label{fig:enter-label 2}
\end{figure}

\subsubsection{G1 Humanoid Mobile Manipulation}

For grasping, we employ a head-mounted Intel RealSense D435\footnote{\href{https://www.intelrealsense.com/depth-camera-d435/}{https://www.intelrealsense.com/depth-camera-d435/}} on the Unitree G1 humanoid to capture RGB-D images, which are processed by {\mname} to extract 2D target coordinates. 
These coordinates guide SAM2~\cite{ravi2024sam} to generate a segmentation mask, which filters the third-person D435 point cloud to isolate the target object. 
The filtered point cloud is then passed to AnyGrasp~\cite{fang2023anygrasp} to predict a grasp pose in the third-person frame, which is transformed to the robot’s base frame using known camera-to-robot calibration.

For navigation, the chest-mounted L515 camera continuously captures images used by {\mname} to detect nearby landmarks (\eg, a table near the robot). 
The resulting 2D locations, combined with depth and intrinsics, are projected into 3D world coordinates and integrated into a global map via FAST\_LIO\_LOCALIZATION\_HUMANOID\footnote{\href{https://github.com/deepglint/FAST_LIO_LOCALIZATION_HUMANOID/tree/main}{https://github.com/deepglint/FAST\_LIO\_LOCALIZATION\_HUMANOID/tree/main}} for SLAM-based navigation, as shown in Fig~\ref{fig:enter-label}, \ref{fig:enter-label 2}.

For placement, the head-mounted D435 captures images processed by {\mname} to localize the target placement region. The corresponding 3D coordinates, computed from depth and intrinsics, are transformed into the robot’s base frame for accurate placement execution.

\begin{table}[t]
\caption{Ablation on the combination of RGB/RGB-D from {\dname} for SFT (\ie, spatial understanding enhancement).
Top-1/Top-2 accuracies are represented using \textbf{bold text}, and \underline{underlines}.
}
\centering
\small
\begin{tabular}{ll|ccc|cc}
\toprule
\multirow{2}{*}{Method}  & \multirow{2}{*}{Input}             & \multicolumn{3}{c}{CV-Bench~\cite{tong2024cambrian}}       & \multicolumn{2}{c}{$\text{BLINK}_{val}$~\cite{fu2024blink}}  \\
                        &                                    & 2D-Relation & 3D-Depth & 3D-Distance & 2D-Relation & 3D-Depth \\
\midrule
\multicolumn{7}{c}{\cellcolor{mygreen}\textit{Only RGB-D from {\dname}}} \\
\midrule
{\mname}-2B-SFT     & RGB    & 87.69 & 86.83 & 82.50 & 79.02 & 81.45 \\
\midrule 
\multicolumn{7}{c}{\cellcolor{myblue}\textit{Combination of RGB and RGB-D from {\dname}}} \\
\midrule
{\mname}-2B-SFT     & RGB    & \underline{96.15} & \underline{95.83} & \underline{90.67} & \underline{83.92} & \underline{88.71} \\
{\mname}-2B-SFT     & RGB-D  & \textbf{96.31} & \textbf{97.17} & \textbf{90.83} & \textbf{87.41} & \textbf{91.13}  \\
\bottomrule[1pt]
\end{tabular}
\label{supptab: understanding}
\end{table}

\subsection{More Ablation Studies}

We conduct additional ablation studies to identify which design choices enhance the performance.

\highlight{{\dname} RGB and RGB-D combination for SFT training encourages the image encoder to learn spatial understanding beyond depth cues.}
In Tab.~\ref{supptab: understanding}, incorporating both RGB and RGB-D data from {\dname} in the second stage of SFT training effectively enhances the image encoder’s spatial understanding. 
In contrast, training solely with RGB-D may lead to over-reliance on the depth encoder, limiting the image encoder’s ability to learn spatial cues from RGB images alone.

\section{More Demonstrations}
\label{suppsec: more demonstrations}

\begin{figure}[ht]
\centering
\includegraphics[width=1\linewidth]{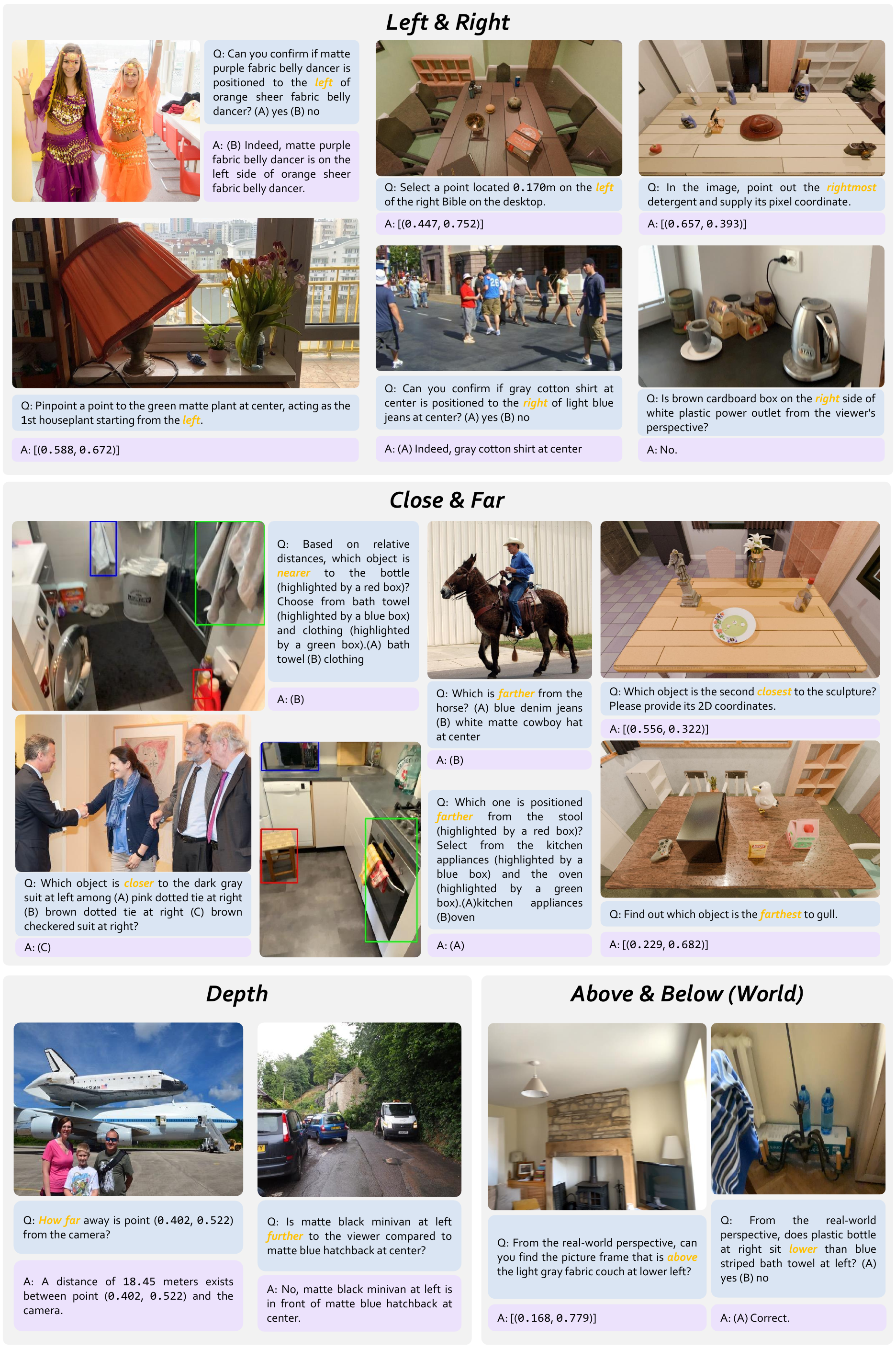}
\caption{Visualization of {\dname}.}
\label{fig:left_right_close_far_depth_above_below}
\end{figure}

\begin{figure}[ht]
\centering
\includegraphics[width=1\linewidth]{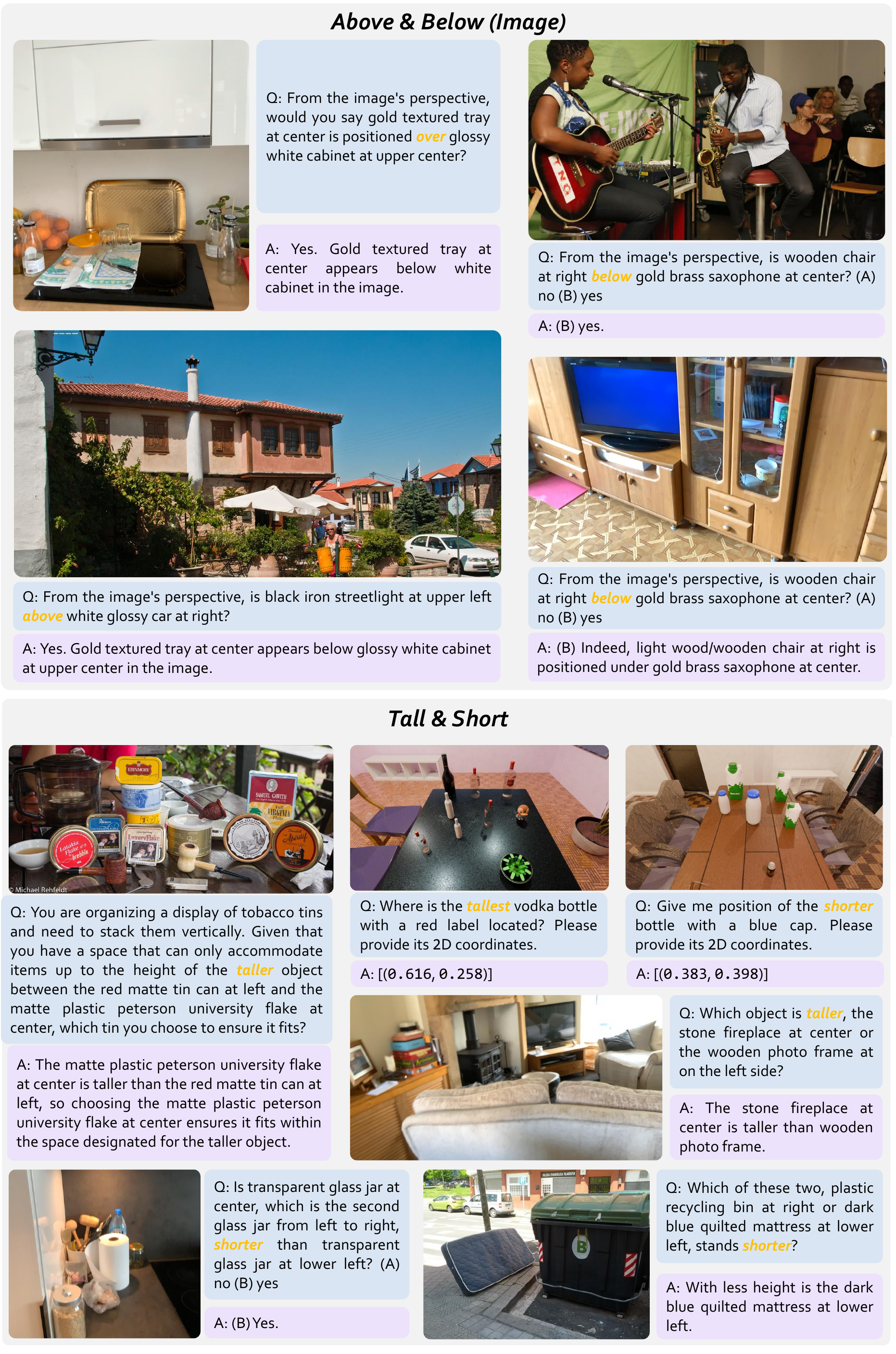}
\caption{Visualization of {\dname}.}
\label{fig:below_above_tall_short}
\end{figure}

\begin{figure}[ht]
\centering
\includegraphics[width=1\linewidth]{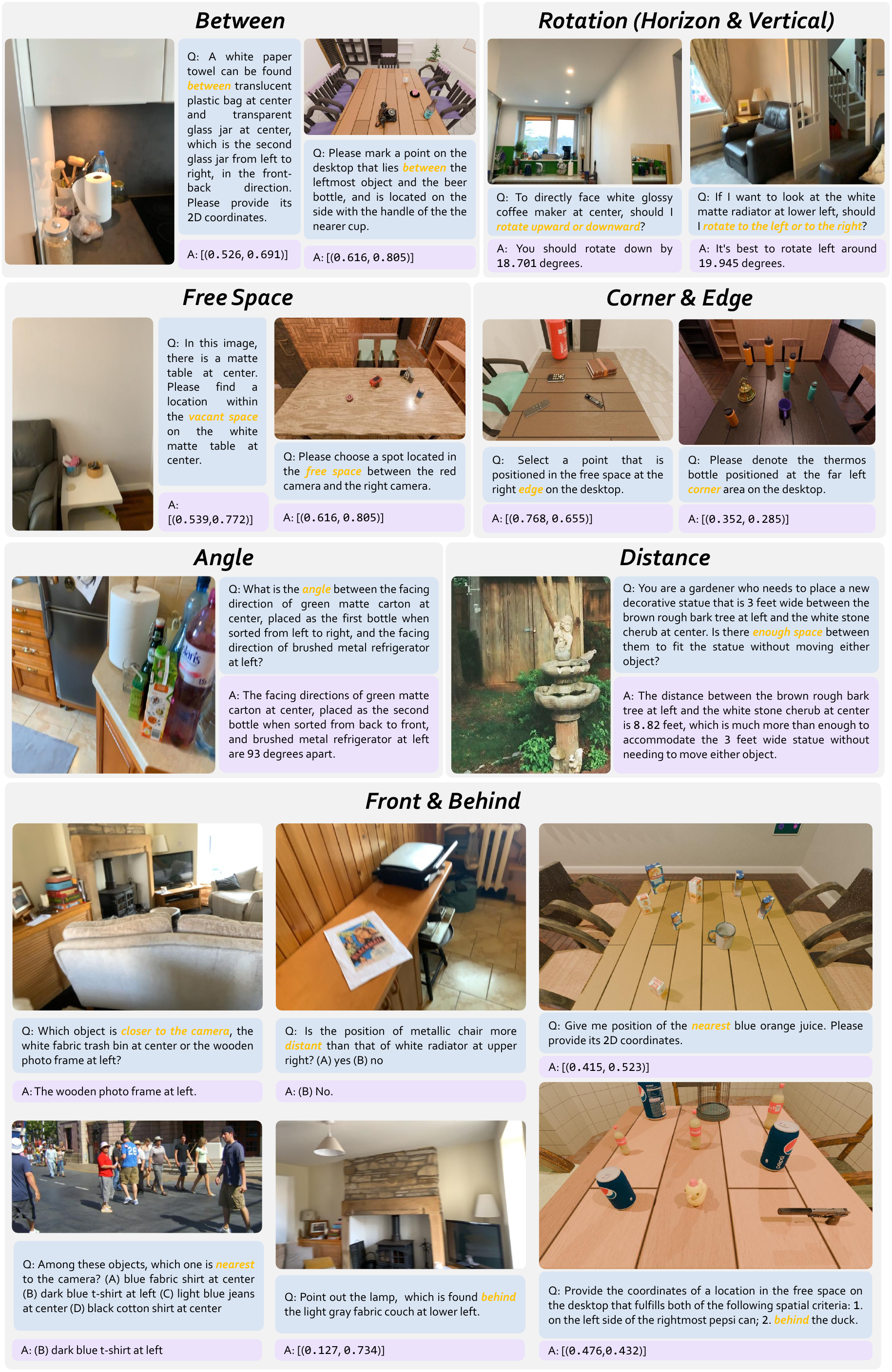}
\caption{Visualization of {\dname}.}
\label{fig:between_rotation_free_corner_edge_angle_distance_front_behind}
\end{figure}

\begin{figure}[ht]
\centering
\includegraphics[width=1\linewidth]{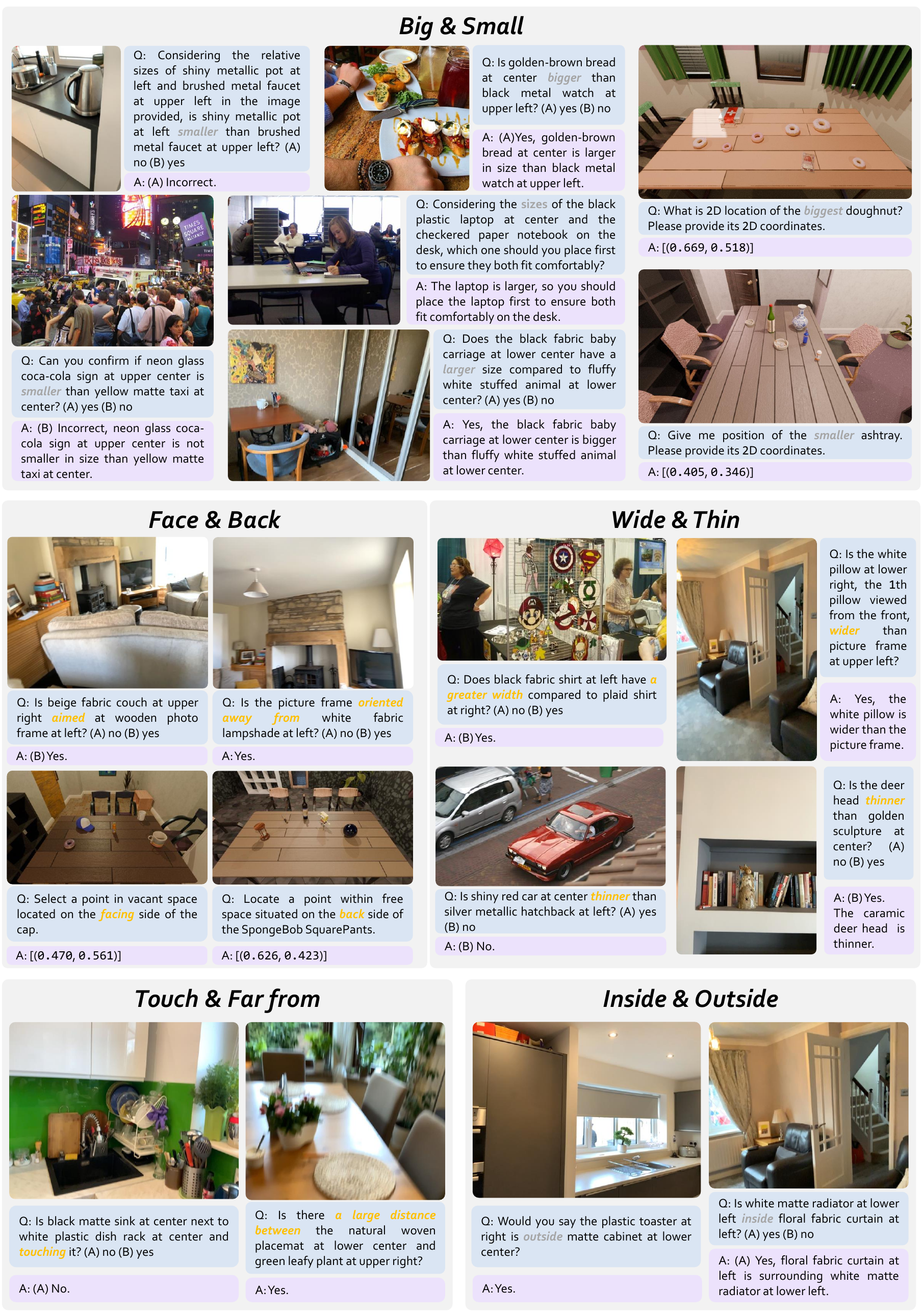}
\caption{Visualization of {\dname}.}
\label{fig:big_small_face_wide_touch_inside}
\end{figure}

\textbf{Visualization of {\dname}.}
We present dataset examples in Fig~\ref{fig:left_right_close_far_depth_above_below},  \ref{fig:below_above_tall_short}, \ref{fig:between_rotation_free_corner_edge_angle_distance_front_behind}, \ref{fig:big_small_face_wide_touch_inside}, which cover 31 distinct types of spatial relationships.

\begin{figure}[ht]
\centering
\includegraphics[width=1\linewidth]{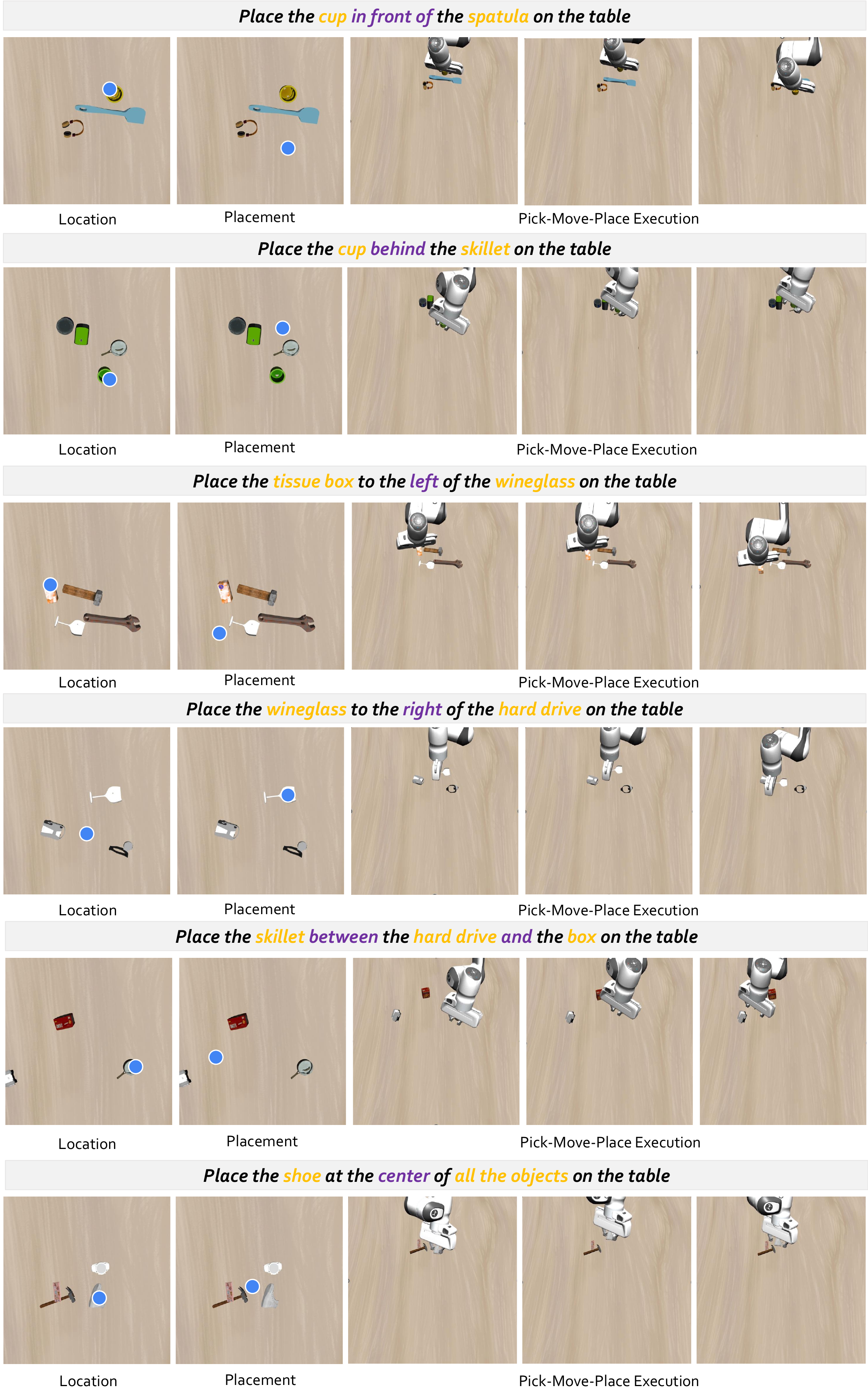}
\vspace{-7mm}
\caption{Visualization of {\mname}'s prediction (blue point) on Open6DOR V2~\cite{qi2025sofar} benchmark.}
\label{fig:simulator_result}
\end{figure}

\textbf{Visualization of Simulation Evaluation.}
We present example rollouts of {\mname} in Fig.~\ref{fig:simulator_result}.

\begin{figure}[ht]
\centering
\includegraphics[width=1\linewidth]{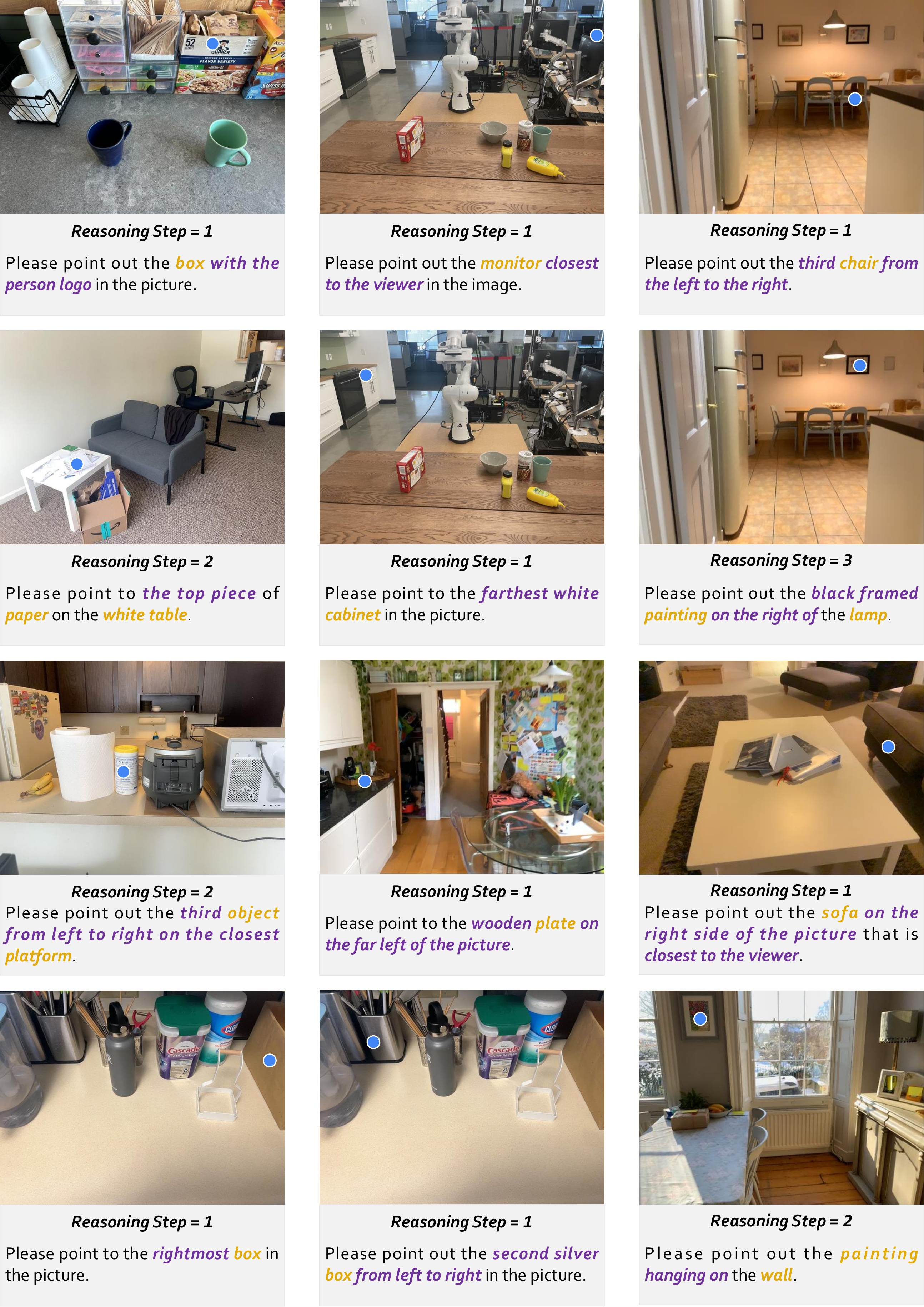}
\caption{Some Location Examples. The model is asked to identify the object referred to by a prompt. The blue point shows the {\mname}'s prediction (all correct).}
\label{fig:location_example1}
\end{figure}

\begin{figure}[ht]
\centering
\includegraphics[width=1\linewidth]{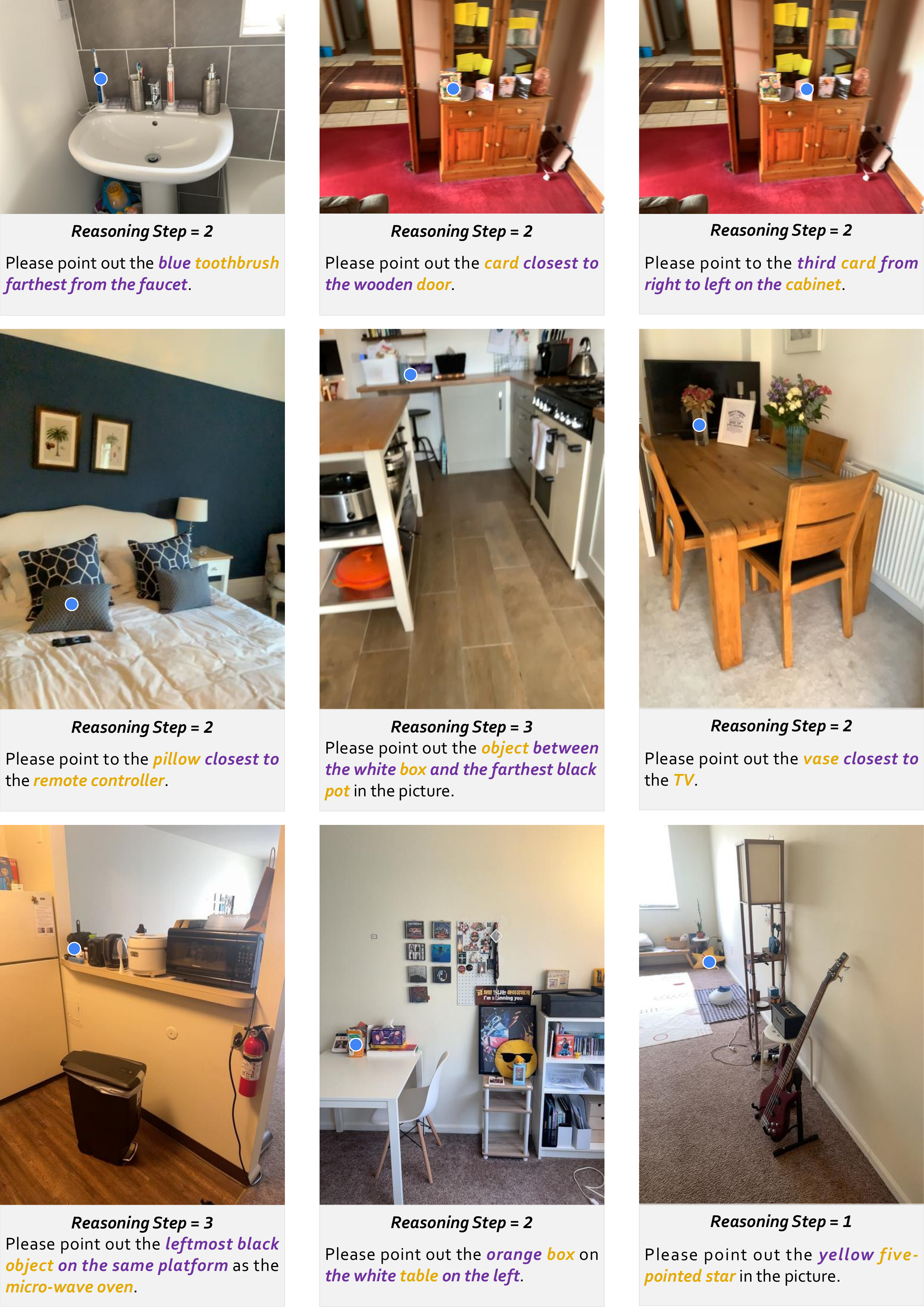}
\caption{Some Location Examples. The model is asked to identify the object referred to by a prompt. The blue point shows the {\mname}'s prediction (all correct).}
\label{fig:location_example2}
\end{figure}

\begin{figure}[ht]
\centering
\includegraphics[width=1\linewidth]{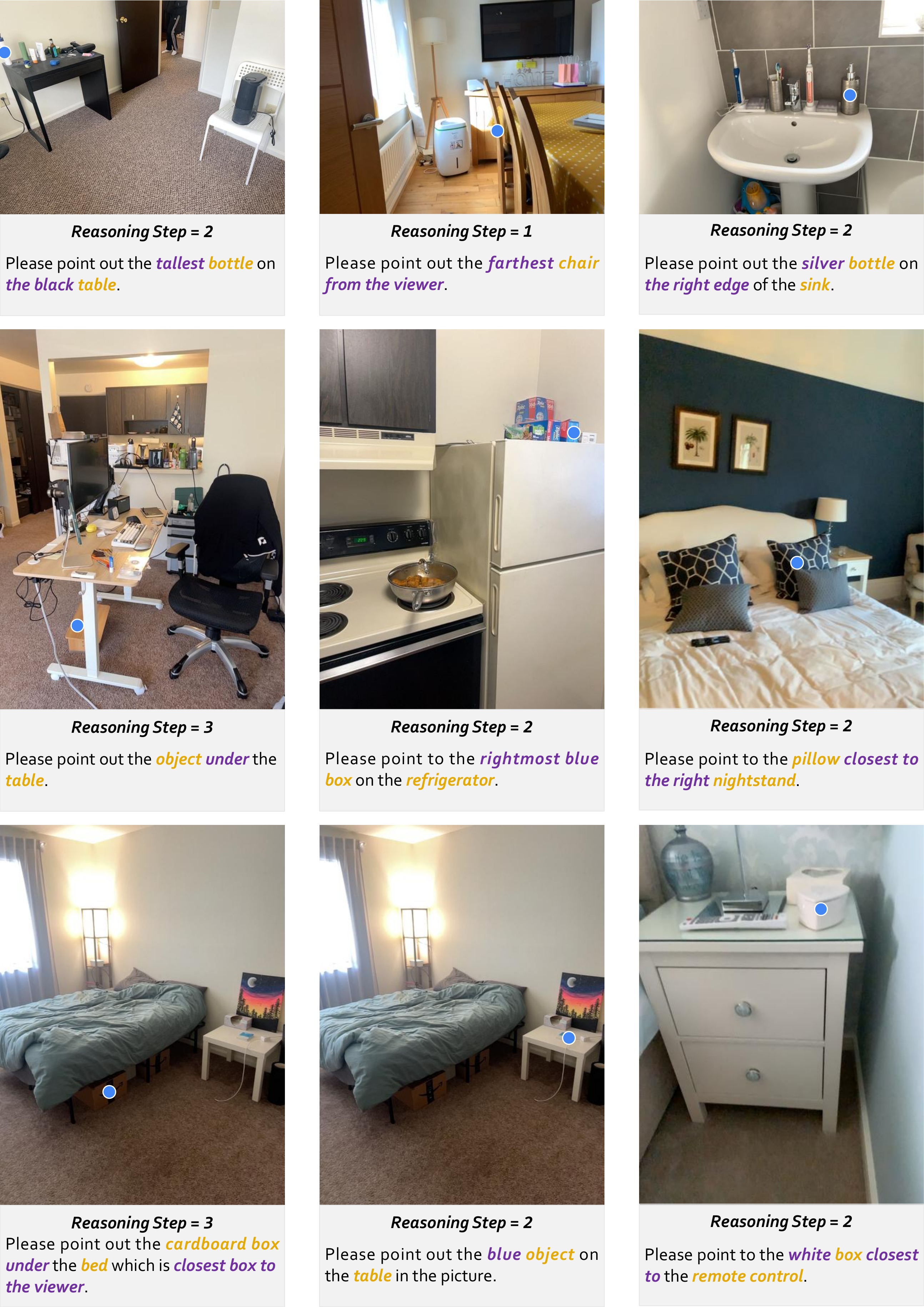}
\caption{Some Location Examples. The model is asked to identify the object referred to by a prompt. The blue point shows the {\mname}'s prediction (all correct).}
\label{fig:location_example3}
\end{figure}

\begin{figure}[ht]
\centering
\includegraphics[width=1\linewidth]{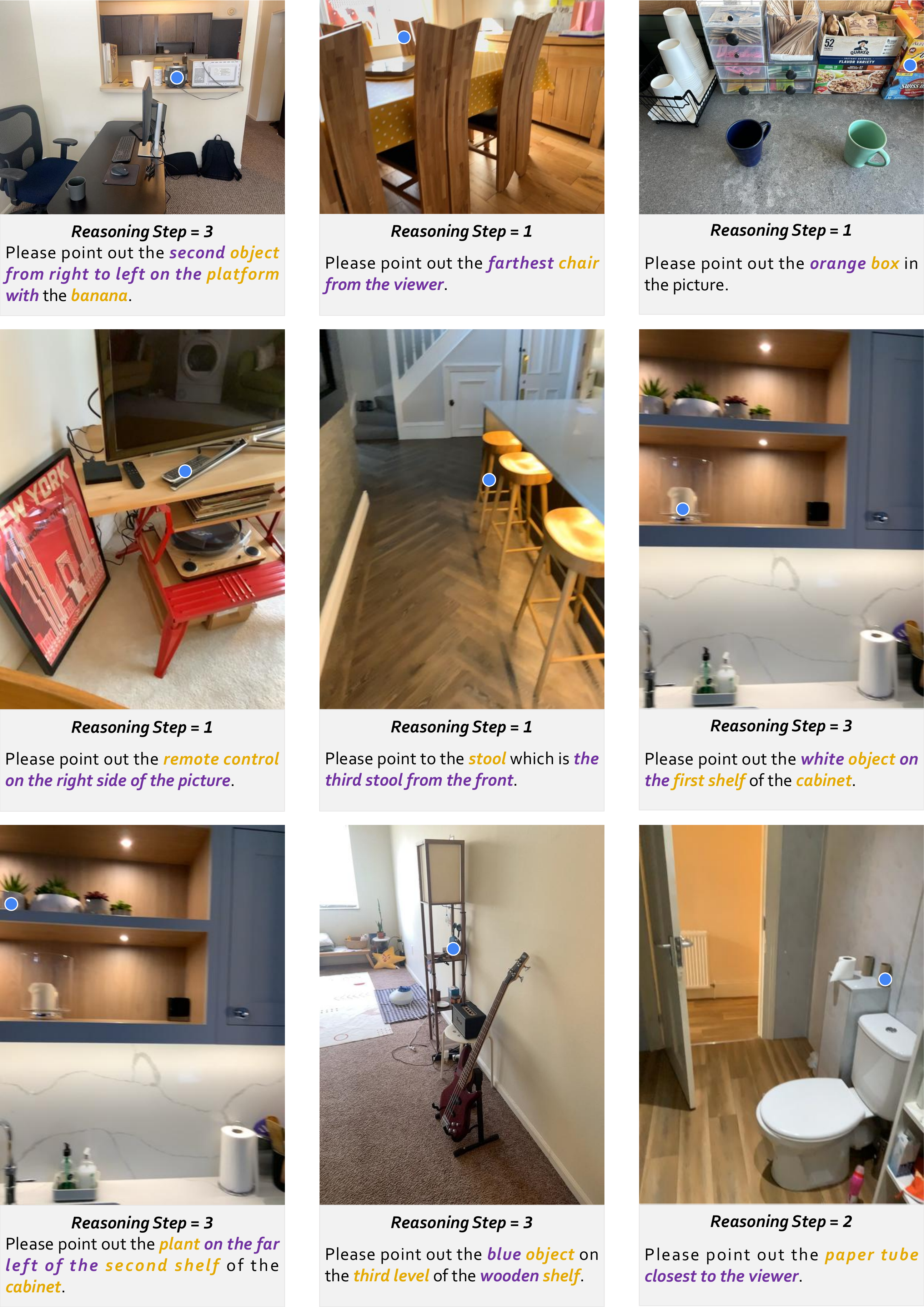}
\caption{Some Location Examples. The model is asked to identify the object referred to by a prompt. The blue point shows the {\mname}'s prediction (all correct).}
\label{fig:location_example4}
\end{figure}

\begin{figure}[ht]
\centering
\includegraphics[width=1\linewidth]{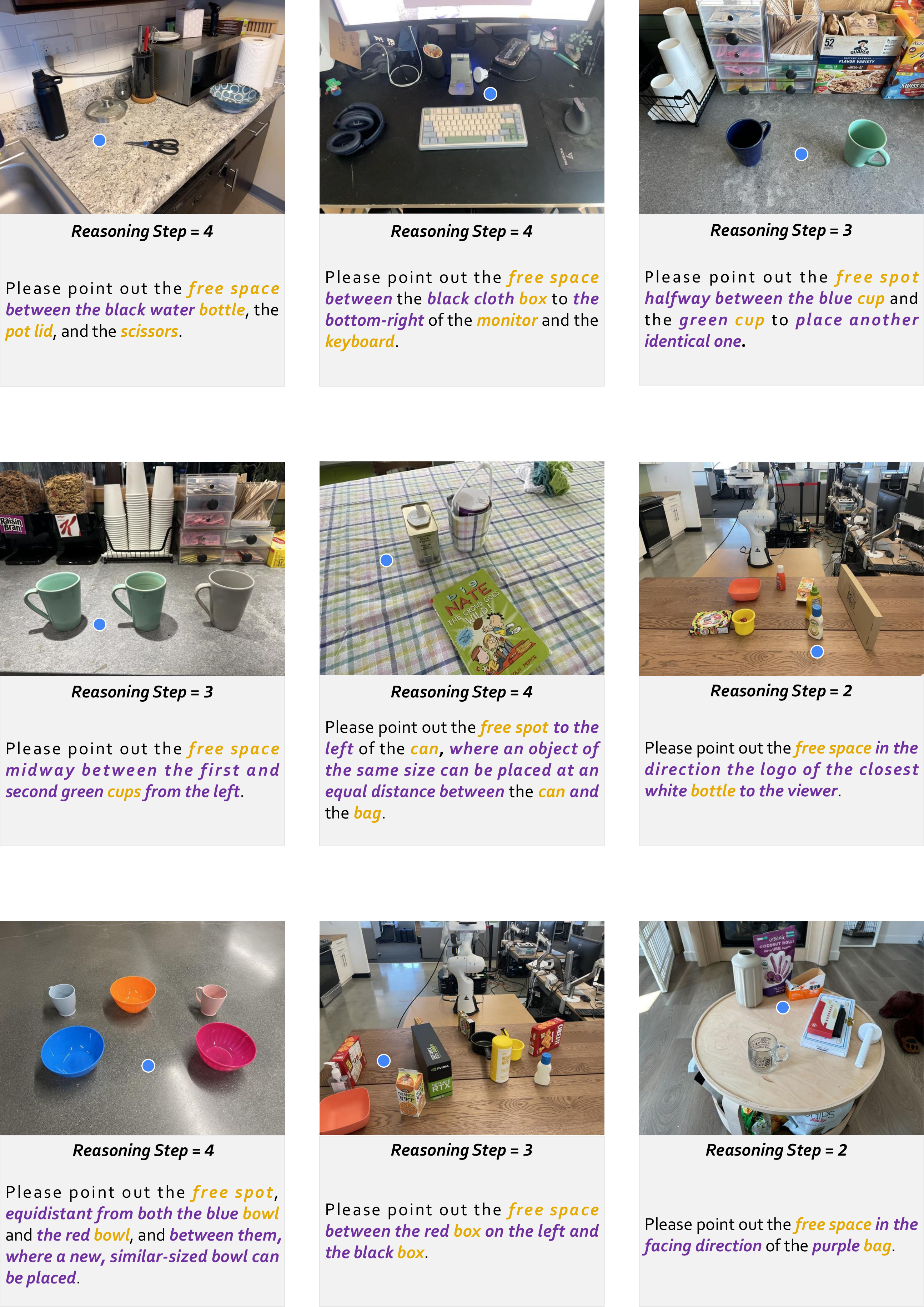}
\caption{Some Placement Examples. The model is asked to identify a valid free space based on spatial reference. The blue point shows the {\mname}'s prediction (all correct).}
\label{fig:placement_example1}
\end{figure}

\begin{figure}[ht]
\centering
\includegraphics[width=1\linewidth]{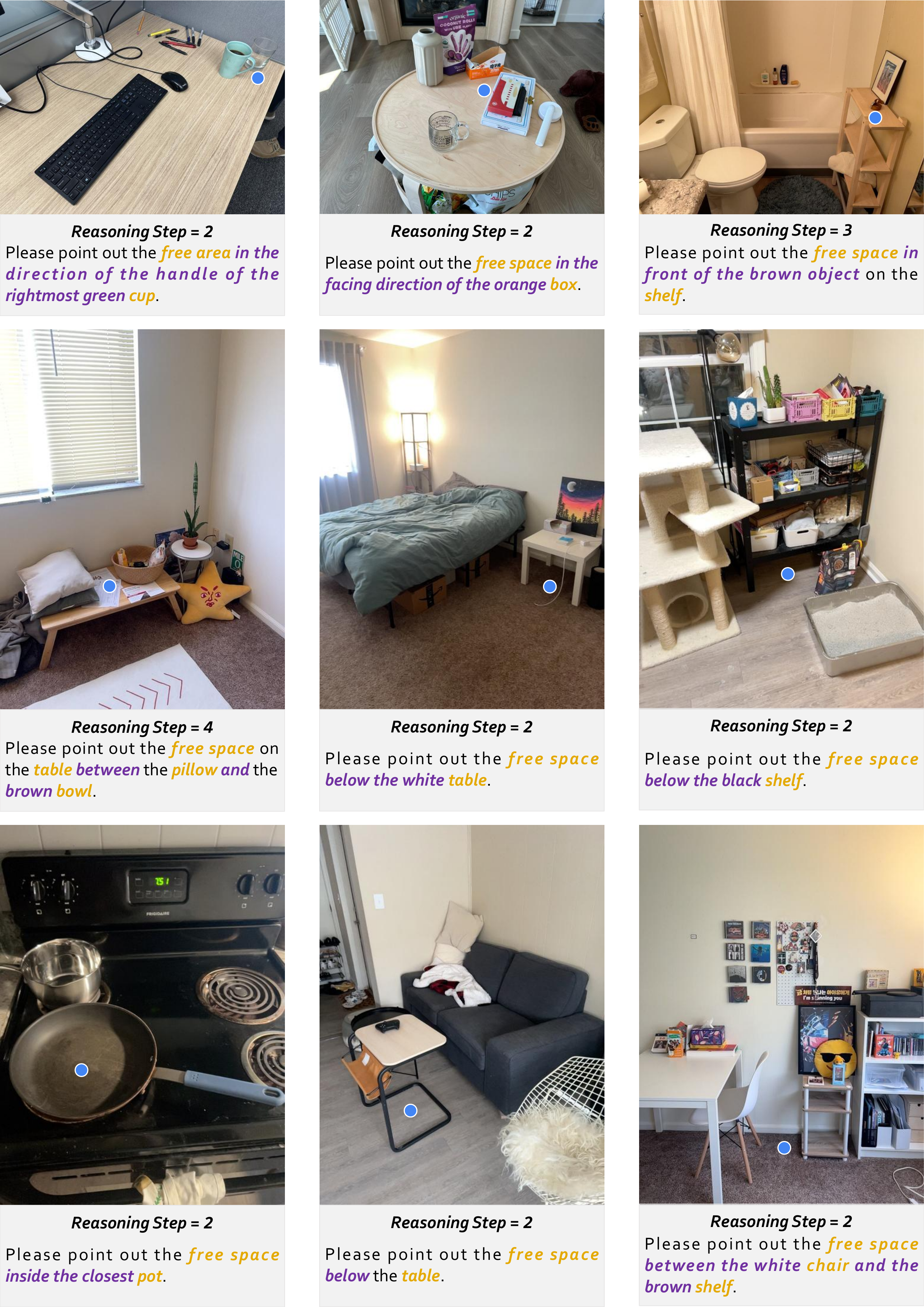}
\caption{Some Placement Examples. The model is asked to identify a valid free space based on spatial reference. The blue point shows the {\mname}'s prediction (all correct).}
\label{fig:placement_example2}
\end{figure}

\begin{figure}[ht]
\centering
\includegraphics[width=1\linewidth]{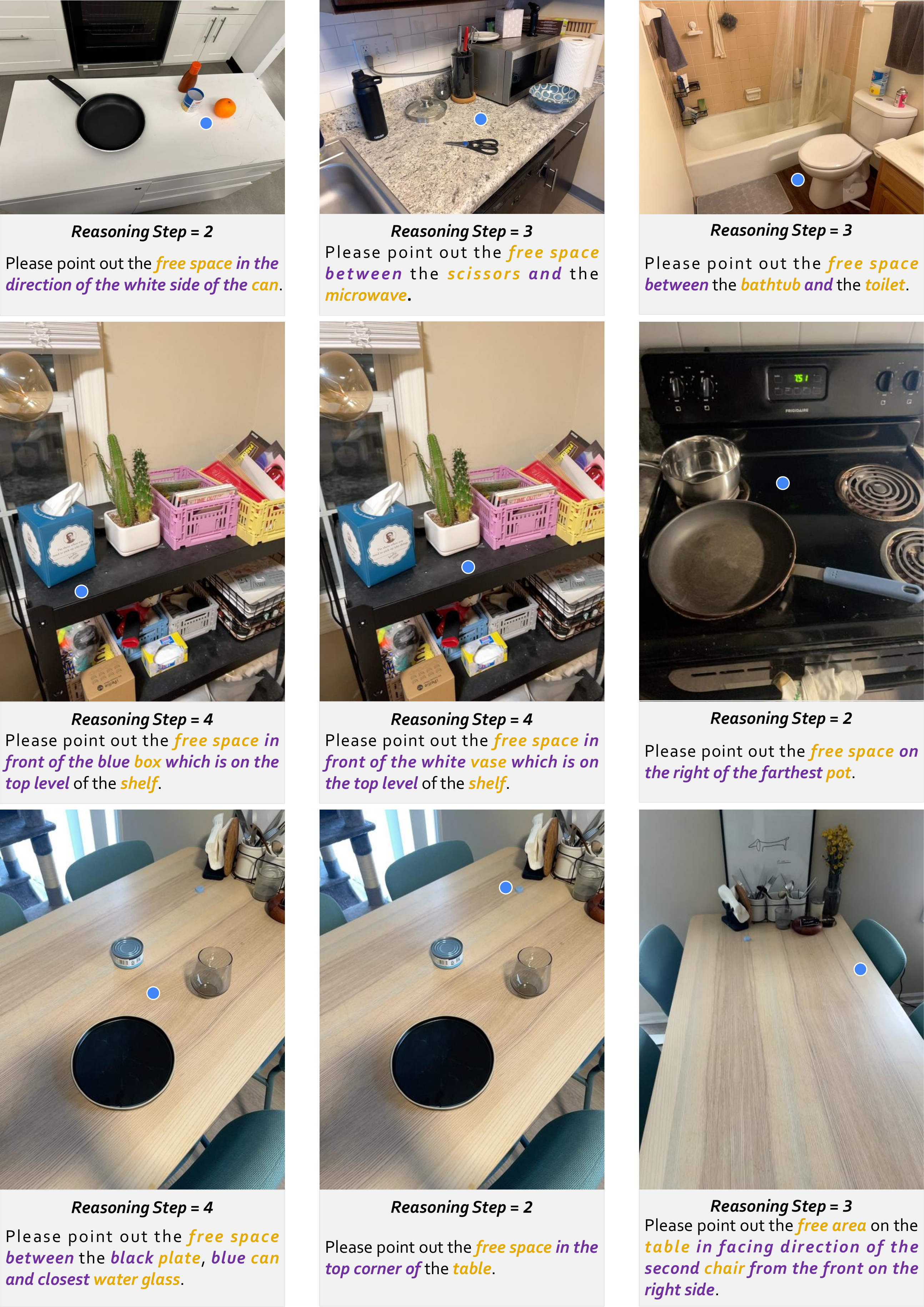}
\caption{Some Placement Examples. The model is asked to identify a valid free space based on spatial reference. The blue point shows the {\mname}'s prediction (all correct).}
\label{fig:placement_example3}
\end{figure}

\begin{figure}[ht]
\centering
\includegraphics[width=1\linewidth]{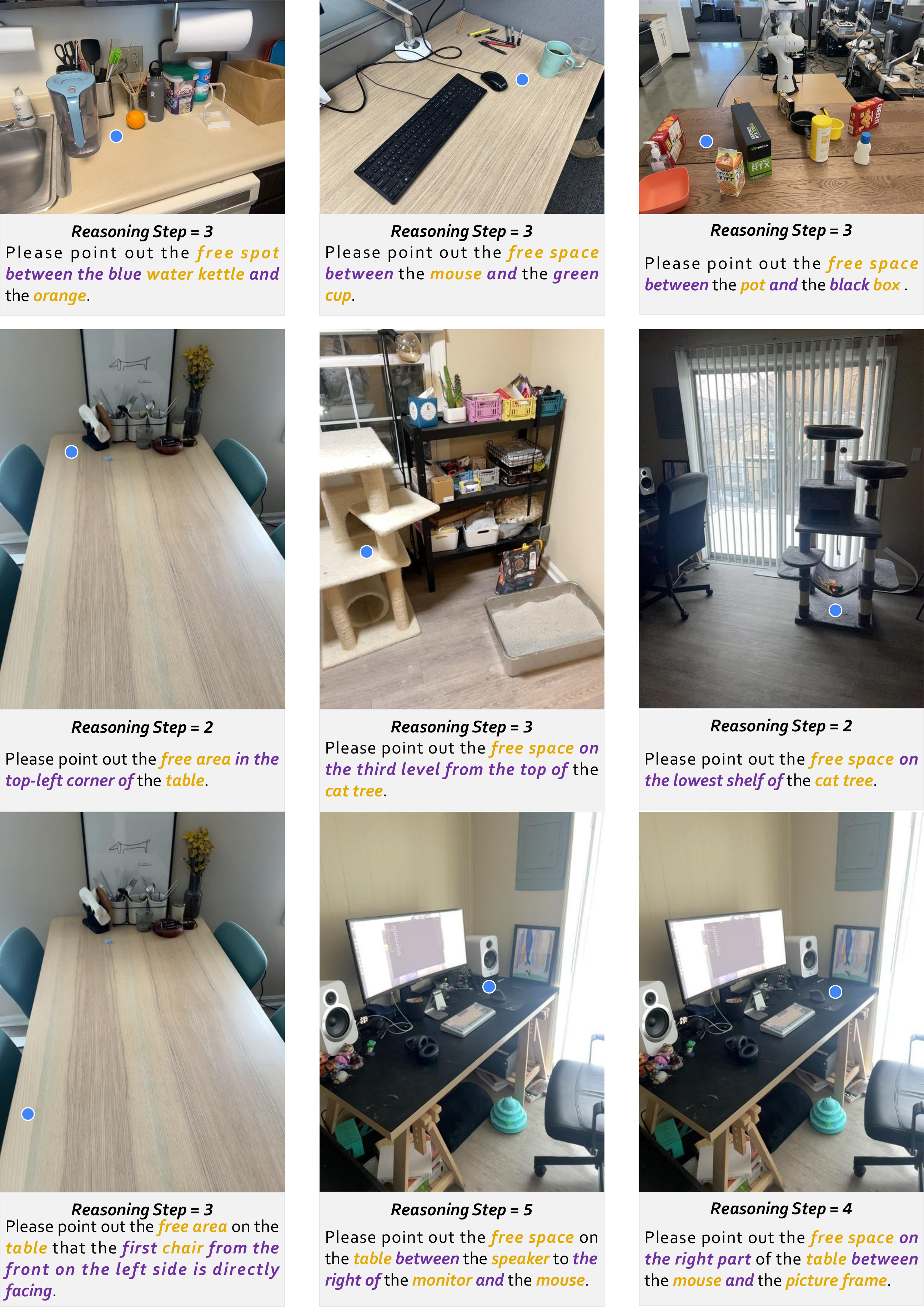}
\caption{Some Placement Examples. The model is asked to identify a valid free space based on spatial reference. The blue point shows the {\mname}'s prediction (all correct).}
\label{fig:placement_example4}
\end{figure}

\textbf{Visualization of {\bname}.}
We present examples of location in Fig.~\ref{fig:location_example1}, \ref{fig:location_example2}, \ref{fig:location_example3}, \ref{fig:location_example4} and placement in Fig.~\ref{fig:placement_example1}, \ref{fig:placement_example2}, \ref{fig:placement_example3}, \ref{fig:placement_example4} with {\mname} predictions.

\textbf{Visualization of Simulator.}
We present example rollouts with {\mname} predictions in Fig.~\ref{fig:simulator_result}.

\begin{figure}[ht]
\centering
\includegraphics[width=1\linewidth]{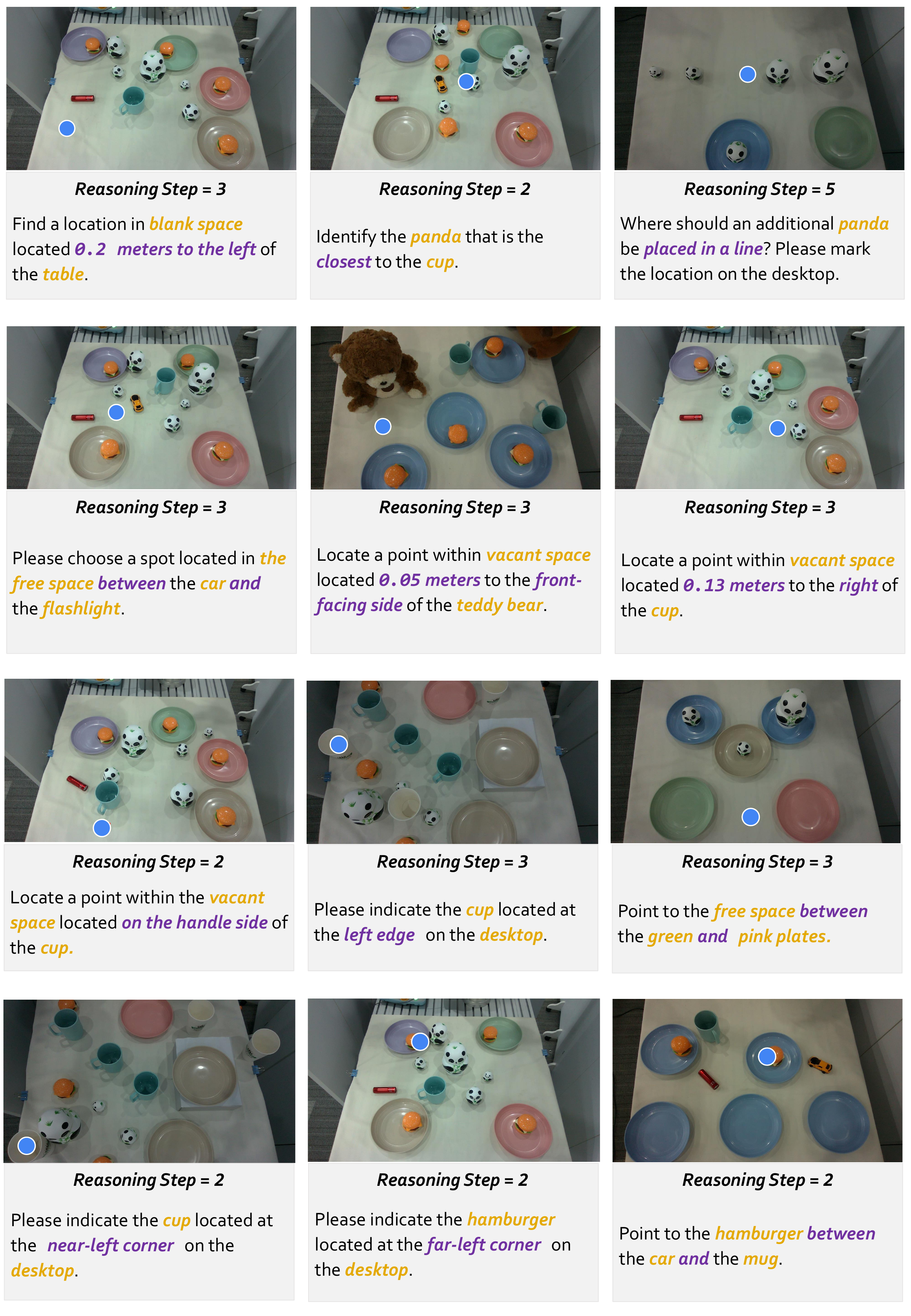}
\caption{Visualization of {\mname}'s prediction (blue point) in the real-world evaluation.}
\label{fig:realworld_experiments}
\end{figure}

\begin{figure}[ht]
\centering
\includegraphics[width=1\linewidth]{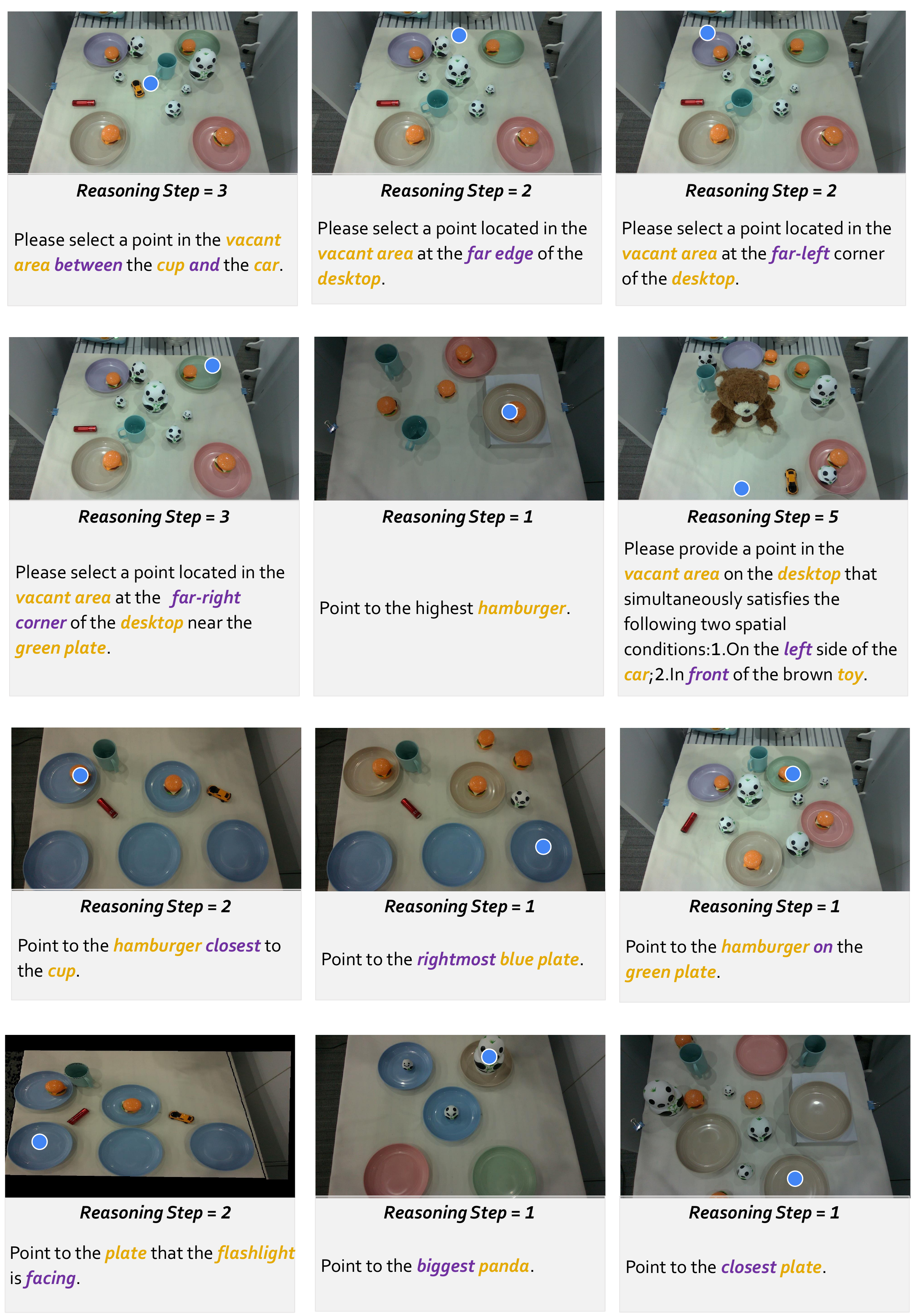}
\caption{Visualization of {\mname}'s prediction (blue point) in the real-world evaluation.}
\label{fig:realworld_experiments_2}
\end{figure}

\begin{figure}[ht]
\centering
\includegraphics[width=1\linewidth]{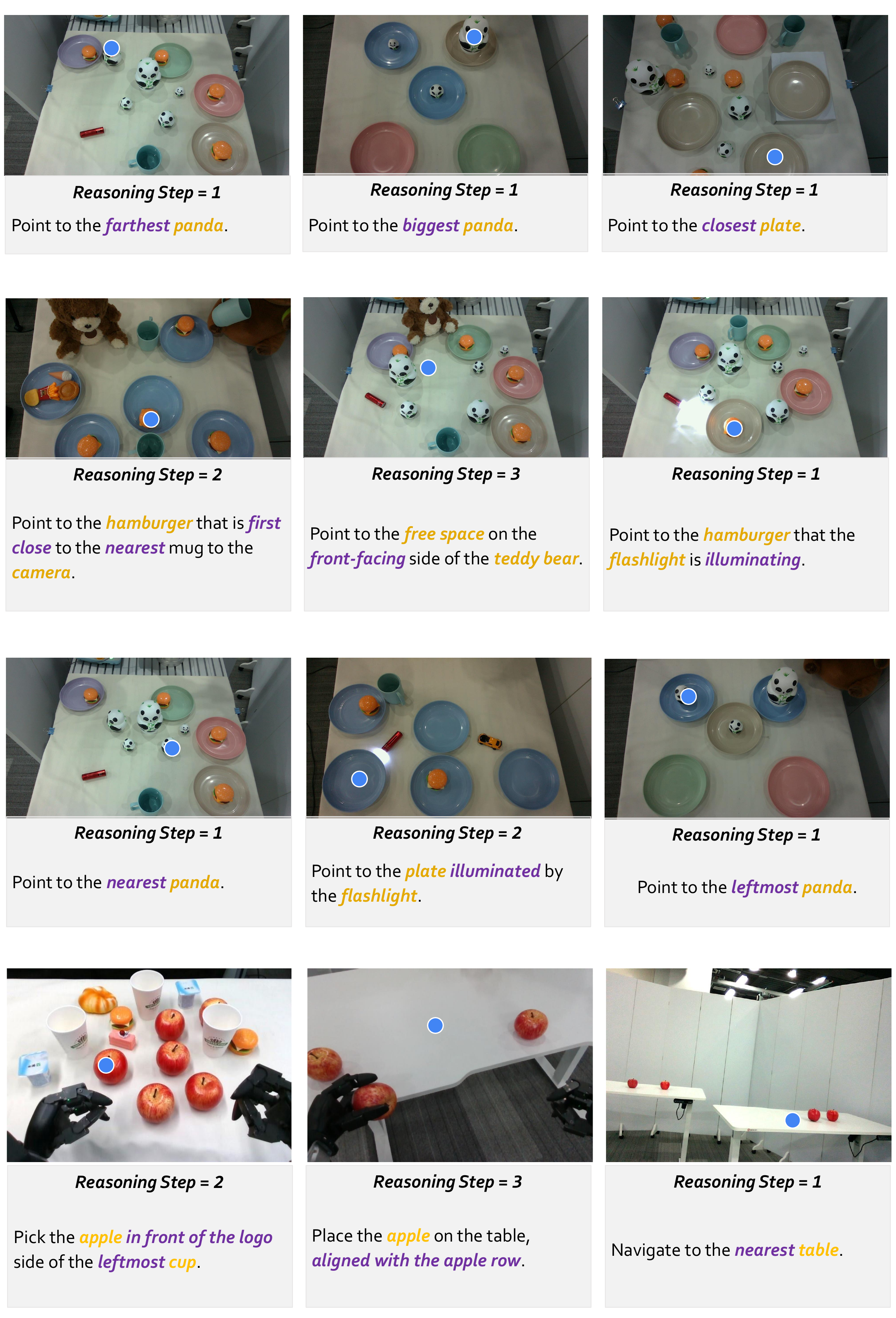}
\caption{Visualization of {\mname}'s prediction (blue point) in the real-world evaluation.}
\label{fig:realworld_experiments_3}
\end{figure}

\textbf{Visualization of Real-world Evaluation.}
We present examples in Fig.~\ref{fig:realworld_experiments}, \ref{fig:realworld_experiments_2}, \ref{fig:realworld_experiments_3}.

\section{More Discussion on Limitations and Future Work}
\label{suppsec: limitation}

Despite achieving promising results, our model still has limitations. 
In particular, it relies on precise textual descriptions to pinpoint specific object locations and placement targets, including accurate references to anchor objects. 
However, in practical real-world robotics scenarios, human instructions are often ambiguous. 
While such utterances may still imply a unique location, resolving them typically requires sophisticated visual-linguistic reasoning and a process of elimination grounded in human prior knowledge, capabilities that challenge our current model.

\begin{figure}[ht]
\centering
\includegraphics[width=0.6\linewidth]{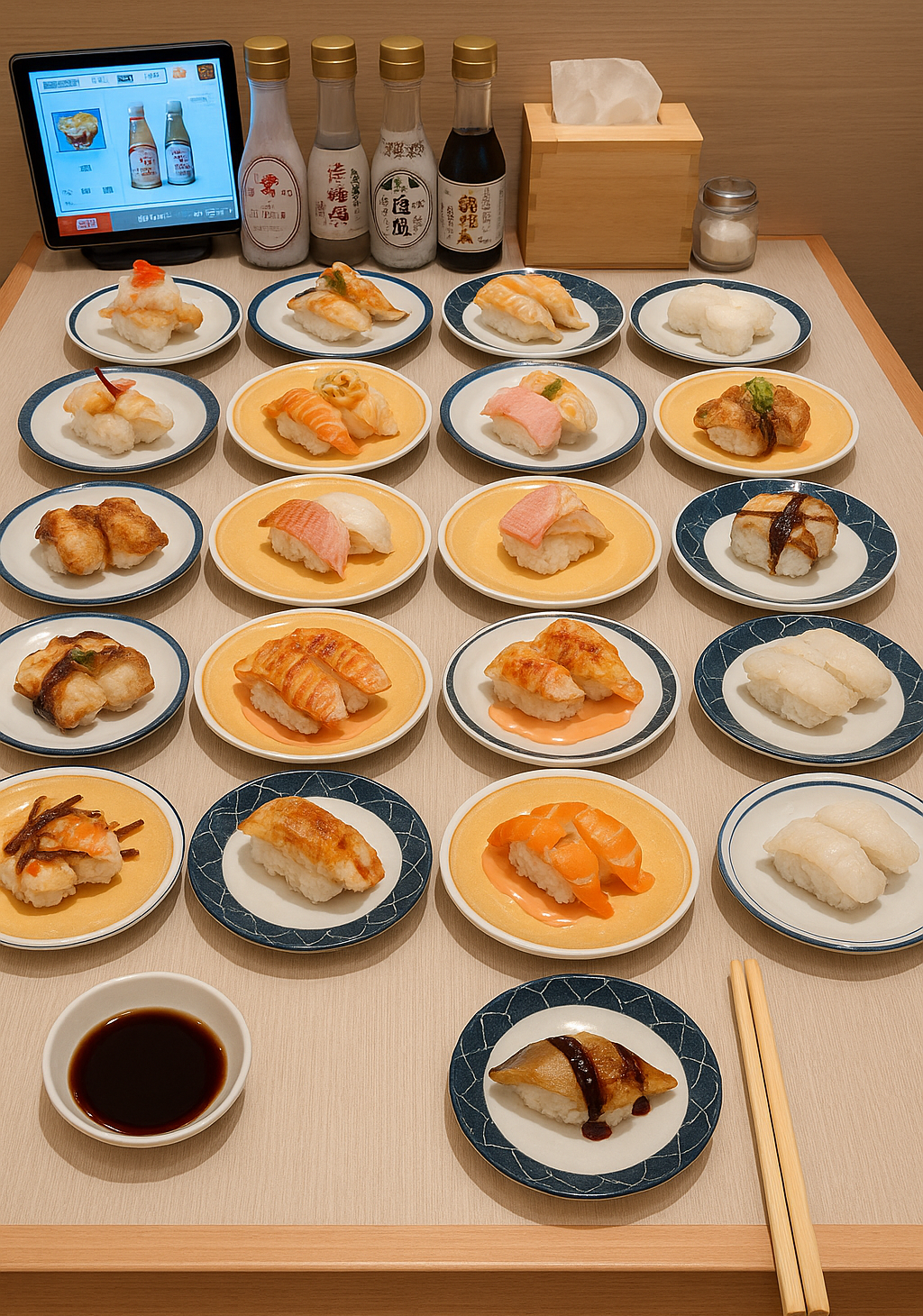}
\caption{Discussion about the limitation of human prior knowledge and intent understanding.}
\label{suppfig:teaser}
\end{figure}

We show two representative but interesting examples that highlight the need for human intent understanding and visual-linguistic reasoning:
\textbf{(1) Probabilistic Preference.} As shown in Fig.~\ref{suppfig:teaser}, humans may refer to a sushi plate as \textit{``pick the one facing the drink''}. In the depicted scene, there are four drink bottles, yet only the two in the middle align with the second sushi plate from the left in the farthest row; the leftmost drink aligns with the first plate, and the rightmost with the third. Despite this ambiguity, people often judge the second plate to be the intended reference due to its higher likelihood of being aligned with two out of the four drinks, reflecting a probabilistic bias in interpretation.
\textbf{(2) Spatial Compatibility.} As shown in Fig.~\ref{suppfig:teaser}, a user might instruct, \textit{``Place another sushi between the plate and the soy sauce dish''}. Although multiple plate–soy sauce pairs exist, only the pair closest to the observer affords sufficient physical space to place another sushi plate. Thus, implicit spatial feasibility guides the correct interpretation, even without explicit constraints.

Our model struggles with these cases because {\dname} lacks multi-step referring data that embeds human priors and intent understanding.
As {\dname} is procedurally generated, incorporating such characteristics at scale remains challenging.
%
%
Future work may explore procedural synthesis of intent-aware data or improve model performance via co-training with datasets supporting intent comprehension, such as PixMo-Points~\cite{deitke2024molmo}.

\section{Broader Impacts}

{\mname} can serve as a versatile visual assistant with advanced spatial understanding and reasoning. 
Due to its integration with large language models (LLMs), it inherits both potential benefits and risks, similar to other VLMs, such as output hallucinations, biases from base models, and heightened energy consumption associated with model upscaling. 
Beyond these considerations, {\mname} can also function as a high-level planner with spatial referring abilities, guiding robots in tasks like manipulation and navigation. 
While such capabilities substantially enhance robotic control, they also pose safety challenges when combined with existing control policies.

Despite these concerns, releasing {\mname} to the broader research community would be highly advantageous. 
Open access would foster continued development and refinement of spatial referring with reasoning, ultimately benefiting diverse robotics platforms (\eg, robotic arms, humanoids) performing various tasks (\eg, manipulation, navigation).

\section{Licenses}

\textbf{(1)} 2D web image data: OpenImages~\cite{kuznetsova2020open} is released under \texttt{Apache License 2.0}.

\textbf{(2)} 3D embodied video data: CA-1M~\cite{lazarow2024cubify} is released under \texttt{CC-by-NC-ND}.

\textbf{(3)} Procedural scene generation: Infinigen~\cite{raistrick2024infinigen} is released under \texttt{BSD 3-Clause License}.

\textbf{(4)} 3D digital assets: objaverse-xl~\cite{deitke2023objaverse} is released under \texttt{Apache License 2.0}.


